\documentclass[journal]{IEEEtran}

\usepackage{cite}
\usepackage{amsmath,amssymb,amsfonts}
\usepackage{algorithmic}
\usepackage{graphicx}
\usepackage{float}
\usepackage{textcomp}
\usepackage{xcolor}
\usepackage{subfigure}
\usepackage{caption}
\usepackage{algorithm}
\usepackage{amsmath}
\usepackage{url}
\usepackage{makecell}
\usepackage{hyperref}
\usepackage{verbatim}
\usepackage{fancyhdr}

\begin{document}
%

\pagestyle{fancy}
\fancyhead[C]{This paper appears in IEEE Journal on Selected Areas in Communications (JSAC), 2023.}
\title{Asynchronous Hybrid Reinforcement Learning for Latency and Reliability Optimization in the Metaverse over Wireless Communications}


%
%

\author{Wenhan Yu,
        Terence Jie Chua, 
        Jun Zhao\thanks{The authors are all with the School of Computer Science and Engineering, Nanyang Technological University, Singapore.  Emails: \newline wenhan002@e.ntu.edu.sg, terencej001@e.ntu.edu.sg, JunZhao@ntu.edu.sg \\ Corresponding author: Jun Zhao
        } 
}

\maketitle
\thispagestyle{fancy}


\begin{abstract}
Technology advancements in wireless communications and high-performance Extended Reality (XR) have empowered the developments of the Metaverse. The demand for the Metaverse applications and hence, real-time digital twinning of real-world scenes is increasing. Nevertheless, the replication of 2D physical world images into 3D virtual objects is computationally intensive and requires computation offloading. The disparity in transmitted object dimension (2D as opposed to 3D) leads to asymmetric data sizes in uplink (UL) and downlink (DL). To ensure the reliability and low latency of the system, we consider an asynchronous joint \mbox{UL-DL} scenario where in the UL stage, the smaller data size of the physical world images captured by multiple extended reality users (XUs) will be uploaded to the Metaverse Console (MC) to be construed and rendered. In the DL stage, the larger-size 3D virtual objects need to be transmitted back to the XUs. We design a novel multi-agent reinforcement learning algorithm structure, namely Asynchronous Actors Hybrid Critic (AAHC), to optimize the decisions pertaining to computation offloading and channel assignment in the UL stage and optimize the DL transmission power in the DL stage. Extensive experiments demonstrate that compared to proposed baselines, AAHC obtains better solutions with satisfactory training time.
\end{abstract}

\begin{IEEEkeywords}
Reinforcement learning, resource allocation, latency, reliability, wireless communications, Metaverse.
\end{IEEEkeywords}

%
\IEEEpeerreviewmaketitle


\section{Introduction}
\subsection{Background}
The Metaverse has brought about a revolution in the Internet space, and is an important element of the \textit{Web} 3.0~\cite{goldman,grayscale}. The introduction of the Metaverse opened doors to not only interactive socialization through Extended Reality (XR), but also through its highly connected virtual world and ecosystem. XR is one of the highlights of the Metaverse~\cite{goldman}. With the extensive research and development in XR technologies~\cite{grayscale}, the once-so-expensive equipment has become affordable for both professionals and ordinary people, entering our daily life. The increasing accessibility of XR equipment drives the penetration of XR technologies, and it is postulated to be a key feature in multi-player games, work-place meetings, and potentially, simulations for scientific research and engineering. The crux of XR is to integrate virtual world scenes with physical world environments. This integrated virtual-physical experience not only promotes greater interactivity for the users with virtual scenes, but also paints an illustrative experience for the users. 

\textbf{XR Computation Offloading. }The \textit{real-time virtual entity transformation} with XR technologies is undoubtedly a highlight application of the Metaverse. We can think of these virtual entities as a merger of our physical world with the virtual world~\cite{lee2021all}. Each physical object in the real world is 3-dimensional (3D), and it has to retain its 3D form when translated into the virtual world. Hence, scenes of real-world objects have to be converted into 3D virtual objects. This replication of physical world objects into the virtual space is also known as \textit{digital twinning}, which is a key underlying component of XR~\cite{lee2021all}. In recent years, the real-time translation and rendering of videos and images to 3D objects have been thoroughly studied~\cite{han2019image}, which enables the bridge between our physical world and the Metaverse. An example is the physical world translation technique, NeuralRecon, proposed by Sun \textit{{et al.}}~\cite{realtime3D1}, which obtains remarkable results in real-time 3D construction from videos. Meng \textit{et al.}~\cite{meng2022sampling} addressed the synchronization between physical objects and the digital models by optimizing the sampling rate and the prediction horizon. Nevertheless, despite rapid development in XR technologies, even existing state-of-the-art user devices~\cite{meta} do not have sufficient computing power to render an XR scenario. An alternative, yet feasible method to powering XR is to consider mobile edge computation offloading (MECO)~\cite{lin2019computation}, which needs us to thoroughly study this system over wireless communications.


\textbf{xURLLC. }In the case of XR computation offloading, the real-world scenes can first be captured by the user device and uploaded to an edge console. The console responsible for translating physical world scenes into virtual scenes (3D) will handle the translation and rendering task and subsequently send the rendered XR scenes back to the user device. The 2D to 3D translation and rendering can lead to much data increment in the downlink (DL) stage. For example, the DL data size can be tens of times the uplink (UL) data size~\cite{monstermash}. Therefore, it could be difficult for the DL data to be transmitted back to users in the DL stage with an acceptable delay, which causes transmission failures, and the failures reflect the unreliability of the system. For example, an unexpectedly large UL transmission data would result in orders of magnitude larger DL data size, which may take an astronomical amount of time for DL transmission. To enable truly immersive XR applications, the users’ seamless experience and low latency need to be guaranteed. Therefore, we need to consider the \mbox{UL-DL} transmission as a whole to fulfill the reliability of the system and different Key Performance Indicators (KPIs) in different transmission stages. 

\subsection{Scenario} 
To address the above issue, this paper considers a novel interactive and asymmetric joint \mbox{UL-DL} scenario. Multiple XR users (XUs) are in a localized indoor area, and each user has physical world scenes (data) to be uploaded to the only Metaverse Console (MC) in the building for the generation of a virtual replica (digital twin). The UL and DL stages are taken in turn. In each UL stage, XUs will upload their data (scenes) to the MC with fixed transmission power decided by their devices within a fixed uplink transmission time interval (UTTI) (in 6G, this is expected to be on the order of milliseconds~\cite{6Gmagzine}) to be rendered and transformed into a 3D virtual digital twin. The MC determines which XU will perform computation offloading, and establishes an appropriate MC-XU channel allocation in the UL stage. The XUs that are not allocated to any channel will be idle for this transmission and wait for later transmission intervals to upload their data. In the DL stage, the MC will transfer the generated data to XUs in a limited downlink transmission time interval (DTTI), and the MC arranges power selection for the DL transmission to each XU. The main reason for having the slotted structure with fixed UTTI and limited DTTI is to ensure the reliability of the whole system so that the transmission delay for each iteration (one iteration contains one UL and one DL transmission) does not take an astronomical amount of time, such that it results in a backlog of subsequent data to be transmitted. In the context of current 3D real-time transmissions, only keyframes are used for generation~\cite{realtime3D1} to lessen the computation overhead. In other words, these keyframes are very important and we need to ensure that they will be transmitted to the MC when embedding MECO. Therefore, to guarantee reliability, if the DL delivery of a packet of data is not completed within the limit of the DTTI, the DL transmission is deemed as a failure and the packet will need to be re-transmitted. Therefore, the UL rate should not always be as fast as possible, and we are required to seek a self-adaptive approach for considering the UL-DL stages in tandem.

\subsection{Methodology} 
Due to the asynchronous multi-stage, discrete-continuous mixed action space (decisions on computation offloading, channel assignment, and power allocation) nature of our problem setting, we propose a novel deep Reinforcement Learning (RL) structure, namely Asynchronous Actors Hybrid Critic (AAHC), in quest of a near-optimal solution. We design two interactive RL Agents to optimize the two transmission stages. As our problem is sequential and divided into stages, we decompose the reward into stage-specific rewards and an overarching global reward. We introduce the AAHC model to our system, which is inspired by the renowned \textit{Hybrid Reward Architecture}~\cite{van2017hybrid}, and use the state-of-art DRL algorithm, Proximal Policy Optimization (PPO)~\cite{schulman2017proximal}, as the backbone.

\subsection{Challenges and motivation}
We next explain our motivations regarding our scenarios and methodologies from the following aspects.

\textbf{6G xURLLC in the Metaverse. }
Advancements in 6G wireless communications~\cite{6G} and high-performance XR technology~\cite{XR} have empowered the developments of the Metaverse, but new challenges arise within. Researchers have made great efforts in the 5G URLLC applications~\cite{URLLC}. However, they do not study diverse mission-critical applications involving XR or within the Metaverse. To power mission-critical real-time applications through the seamless digital twinning and projection of complex and detailed real-world entities onto the Metaverse, a reliable and ultra-low latency communication system is required. Furthermore, to cater to an increasingly ``virtual" population in the Metaverse, one has to consider a more reliable, efficient, and comprehensive 6G xURLLC communication system. Therefore, we design a problem setting with several XUs in which we optimize the multi-user XR wireless transmission problem in the Metaverse. Furthermore, running out of battery in the XR device (often battery-limited) can also be somehow understood as unreliable Metaverse experiences, from the perspective of the human user using the XR device. Hence, incorporating energy consumption into our optimization can help reliable Metaverse experiences by human users. Besides, the Metaverse is a gigantic digital world with immense computational devices, which can lead to serious energy waste and pollution. The ``Green Metaverse Networking'' is first proposed by Zhang \textit{et al.}~\cite{dugreen} to address the necessities of energy efficiencies and energy consumption optimizations. Lee \textit{et al.}~\cite{lee2021all} also emphasize that green computing must be highly regarded, as the waste and pollution strongly influence the attitudes of people towards the Metaverse. Based on these, the optimization in our paper also includes {energy consumption}.

\textbf{Joint Optimization in MECO. }Three important variables in improving the MECO efficiency are \textit{decision to computation offloading}, \textit{channel allocation}, and \textit{power selection}. In our proposed scenario, the first two are optimized in the UL stage. The decision on whether to offload computation refers to which XUs will do computation offloading, and the channel allocation refers to the assignment of each XU to a channel for communicating with the Metaverse Console (MC). Appropriate channel allocation is important in creating an efficient computation offloading system. If an XU is allocated to a channel with too many other XUs assigned, there will be great interference, resulting in an overall inefficient data transfer~\cite{andrews2005interference}. The power selection in DL refers to the selection of power for each user, to perform the data DL transmission. An inappropriate allocation of power can increase interference between users, which is detrimental to data transfer efficiency. Hence, it is paramount for us to jointly optimize these three variables in our work.

\textbf{Asymmetric multi-process to Asynchronous MDP. } In existing computation offloading works as discussed in~\cite{lin2019computation,sun2019adaptive}, the data sizes of downloaded scenes are often taken to be similar to the data sizes of scenes to be uploaded. These works mentioned above do not consider the asymmetric data size scenario. In such circumstances, the downloaded, rendered 3D virtual objects can possibly be orders of magnitude larger than the uploaded scenes. This difference in data sizes introduces an issue of sensitivity, where a slight change in the UL transmission size may possibly induce a dramatic change in the DL data size. Also, the UL and DL are connected and influenced by each other. In this scenario, the power allocation in the DL can only be obtained based on the actions in UL. Thus, the orchestration of the UL and DL transmissions has to be considered in tandem.

Furthermore, the consideration of an asymmetric multi-process transmission and the existence of two agents with different states and actions lead to a new challenge: Asynchronous Markov Decision Process (AMDP). In traditional MDP, the transition in each slot can be formulated as $S_1^1\times A_1^1 \rightarrow R_1^1, S_1^2$ (here the superscript and subscript denote the time step and agent index). However, in a two-agent AMDP, the transition has to be represented as such: $S_1^1 \times A_1^1 \times R_1^1 \times S_2^1 \times A_2^1 \times R_2^1 \rightarrow S_1^2, R_g^1$, which means the state of $Agent_2$ is dependent on and influenced by $Agent_1$, and it's not suitable for them to perform actions simultaneously. Correspondingly, in this scenario, $Agent_1$ is responsible for optimizing the DCOs and channel access, and $Agent_2$ should arrange the downlink power allocation based on the DCOs and channel arrangement. And they are influenced by each other through the relationship between UL and DL data sizes. Moreover, we set the global reward $R_g$ as some important metrics like the transmission failure flag can only be obtained after all agents have performed actions in the iteration, and this global reward is related to both agents. Thus, we give it to both agents to offer them a more specific view of the global process for the networks. This complex asynchronous setting serves as the motivation for us to develop an entirely new approach reinforcement learning approach.

\textbf{Multiple Objectives to Hybrid reward RL.}
In our work, we consider a complex problem scenario in which we take into account multiple processes and high-dimensional objectives. Traditional RL approaches struggle with handling such complex problems as they lack the ability to handle each underlying objective on a minuscule scale. Therefore, we proposed and designed a hybrid reward architecture that allows us to achieve optimality in an overarching objective while ensuring near-optimality in the underlying objectives.

In this paper, we propose a novel multi-agent hybrid reinforcement learning approach to solve the joint optimization problem. Our contributions can be summarized as:
\begin{itemize}
    \item \textbf{xURLLC in the Metaverse.} We formulated a novel joint optimization problem in a multi-process xURLLC system on the Metaverse, and proposed a feasible and practical RL-based solution to tackle the problem. The UL and DL are considered in tandem to ensure the reliability and low latency of the system.

    \item \textbf{Joint optimization in multi-process transmissions.} We conducted the joint optimization of decisions on computation offloading, channel assignment, and power allocation in the \mbox{UL-DL} transmissions.
    
    \item \textbf{Asynchronous MDP.} We first studied a novel AMDP wireless communication problem, and devised a new RL structure to solve it.
    
    \item \textbf{Asymmetric Agents.} We adopted two agents with separate local objectives (one discrete and the other continuous) and an overarching global objective.

    \item \textbf{Hybrid Critic PPO.} We proposed a novel approach that uses a hybrid Critic to guide the training and convergence of both agents. Our approach achieves the total delay, re-transmission percentage, and energy cost improvements of $65.48\%, 90.67\%, 56.82\%$ in the most complicated scenario compared to iterative RL~\cite{IPPO}.
\end{itemize}

The rest of this paper is organized as follows. (i) We cover related literature to this piece of work and emphasize the novelty of our work in Section~\ref{related work}. (ii) The system model and problem formulation are then introduced in Section~\ref{section: problem formulation}. (iii) In Section~\ref{environment}, we explain how the RL settings are crafted for our proposed environment. (iv) The details of our algorithm and model implementation are then expounded in Section~\ref{algorithm}. (v) Next, we compare our proposed approach with baseline models and provide in-depth analyses of the results in Section~\ref{experiment}. (vi) Finally, we provide a summary and conclusion to our work in Section~\ref{conclusion}.

\section{Related Work}
\label{related work}
This paper studies joint optimization in an ultra-reliable and low-latency MECO communication system with a reinforcement learning approach. Therefore, the related work contains the aspects of (i) mobile edge URLLC that is relevant to our scenario, (ii) joint optimization with RL as we jointly optimize the decisions on computation offloading, channel assignment, and power allocation, (iii) multi-agent RL as we use multiple RL agents in different transmission stages, and (iv) hybrid reward RL as the objectives in our system are high-dimensional and complicated.

\subsection{Mobile edge ultra-reliable and low latency communication}
Several works have utilized traditional convex optimization tools~\cite{kai2020collaborative}, game theory~\cite{xiao2019vehicular}, reinforcement learning~\cite{alfakih2020task}, distributed learning~\cite{qian} approaches to tackle the channel allocation and power selection problem in MECO studies. However, most of these works consider only single-stage data transmission scenarios: optimizing either the uplink (UL) or DL process. While it may seem that the UL and DL transmission processes are two separate processes, it is important to consider the concurrent optimization of both stages. Besides, many works have studied the applications of 5G URLLC on MECO in enhancing the performance of wireless communication systems applications~\cite{merluzzi2020dynamic,liu2019dynamic}. However, none of them considers the scenarios and KPIs in asynchronous multiple transmission processes in XR or Metaverse. What sets this work apart from those discussed above is that we consider a multi-stage scenario with deep reinforcement learning method, handling both UL and DL transmission which would require two different agents in a realistic scenario.

\subsection{Joint optimization with reinforcement learning}
Recently, there are also some excellent works solving joint optimization problems with RL. Guo \emph{et al.}\cite{JO1} solved the handover control and power allocation joint problem using MAPPO and obtained satisfactory results. It aggregates all users' actions under two stages and follows the traditional Centralized Training Decentralized Execution (CTDE) paradigm. However, the problem they aimed to address is not multi-stage and hence directly uses MAPPO. He \emph{et al.}~\cite{JO2} studied joint power allocation and channel assignment problems with RL in NOMA system. This work models the channel assignment as a reinforcement learning task and conducts power allocation under the current channel assignment. This method optimizes channel assignment and power allocation in turn, but it doesn't solve a time-sequential problem like the ones specified in our work.

\subsection{Multi-agent reinforcement learning}
In our work, we employed multiple agents. However, the widely used MARL algorithms~\cite{lowe2017multi,rashid2018qmix} are not feasible methods to tackle our problem. These algorithms adopt the \textit{centralized training and decentralized execution} (CTDE) approach~\cite{lowe2017multi,foerster2018counterfactual}. Specifically, each agent involved will have its own Critic that considers the actions of all the other agents in a single time step. The individual agents' Actors will have decentralized policies and execute actions in a decentralized fashion. However, the existing adaptations of the conventional MARL algorithm are not suitable methods to solve our proposed problem. Firstly, the agents under the MARL algorithm select actions simultaneously in each time step, and the agents' Critics consider all actions of other agents in each time step. Each agent's observation space does not efficiently take in sequential agent actions, as the observation space of each agent will be sparse in each time step. Secondly, the two agents in our problem are utterly distinct from each other. In the UL stage, the user-console allocation action by the UL agent is discrete, while the output power selection by the DL agent is continuous. The \textbf{asynchronous nature} of our system model and the \textbf{asymmetric roles} of our agents make it inappropriate and impractical to adopt the existing MARL approach, with each agent sharing information with others by allowing the Critic to take in all states and actions of both agents. Thus, we decompose the reward assignment into multiple transmission stages and propose our novel model to address such challenges. We detail and elaborate on our algorithm in Section~\ref{algorithm}.

\subsection{Hybrid reward reinforcement learning}
High-dimensional objective functions are common in communication problems, as we usually need to consider multiple factors such as energy consumption and time delay. When using reinforcement learning to tackle such problems, it is important to design high-dimensional objectives as smaller components of additive rewards. This issue of using RL to solve a high dimensional objective function was first studied in \cite{van2017hybrid}. In their work, they proposed a \textit{Hybrid Reward Architecture} (HRA) for reinforcement learning which aims to decompose high-dimensional objective functions into several simpler objective functions. HRA permits the discovery of reasonably good solutions, easily.

Our proposed Hybrid Critic differs from the HRA~\cite{van2017hybrid} in several aspects. Primarily, the HRA breaks the main objective into simpler objectives, allocating each simpler objective to different agents. Then the state-action value pair for each agent is aggregated to produce a unified value. In our hybrid value network, our reward is decomposed across several Critic branches. Instead of aggregating sub-target values of the single agent as per the HRA, our loss is aggregated across each Critic branches for each part (UL, DL, global) to facilitate the cooperative training of the agents, striving to achieve their own objectives and the global objectives, together. The HRA aims to find an optimal solution to a complex scenario by decomposing the reward into target-specific parts. HRA is designed for the single-agent DRL, which excludes the possibility of us adopting it directly. Our adaption of HRA separates the Critic into three branches, for handling the UL, DL, and global parts, respectively.

Next, our system model will be expounded on in the following section.



\begin{table}[t]
\caption{{}{Notations}}
\begin{center}
\begin{tabular}{|l|p{6cm}|}
\hline
{}{Symbol}&{}{Description} \\\hline
{}{$i,m,t$}&{}{Index of XR users, channels, and iteration/time-step/TTI }\\
{}{$\mathcal{M}$}&{}{Channels set: $\{1,2,\ldots,M\}$}\\
{}{$\mathcal{N}$}&{}{XR users set: $\{1,2,\ldots,N\}$}\\
{}{$\mathcal{T}$}&{}{Iteration/time-step (TTI) set: $\{1,2,\ldots, T\}$}\\
{}{$\boldsymbol{z}^t =[z_n^t|_{n \in \mathcal{N}}]$}&{}{Decisions on computation offloading}\\
{}{$\boldsymbol{\kappa}^t =[\kappa_n^t|_{n\in \mathcal{N}}]$}&{}{Channel access management}\\
{}{${d'}_n^t$}&{}{Downlink transmission delay of user $n$ at iteration $t$}\\
{}{$r_n^t$}&{}{Uplink transmission rate of user $n$ at iteration $t$}\\
{}{${r'}_n^t$}&{}{Downlink transmission rate of user $n$ at iteration $t$}\\
{}{$D_n^t$}&{}{Data from uplink of user $n$ at iteration $t$}\\
{}{${D'}_n^t$}&{}{Data need to be transmitted in downlink of user $n$ at iteration $t$}\\
{}{$B_n^t$}&{}{Left data size in the buffer of user $n$ at iteration $t$}\\
{}{$E^t$}&{}{Energy consumption in downlink at $t$.}\\
{}{$W_m$}&{}{Bandwidth of channel $m$}\\
{}{$\tau_u$}&{}{Fixed uplink transmission time interval (UTTI)}\\
{}{$\tau_d$}&{}{Downlink transmission time interval (DTTI) limit}\\
{}{$p_n$}&{}{Uplink power of user $n$}\\
{}{$\boldsymbol{p'}^t,{p'}_n^t$}&{}{Allocated downlink power from MC to XUs}\\
{}{$h^t_{i,m}$}&{}{Channel gain between user $n$ in channel $m$ and MC at iteration $t$}\\
{}{$(\sigma_{n,m}^t)^2, (\sigma_{MC}^t)^2$}&{}{Additive White Gaussian Noise (AWGN) parameter of XU $n$ in channel $m$ at $t$ and the MC, respectively}\\
{}{$\mathcal{N}_m^t(\boldsymbol{\kappa^t})$}&{}{the set of XUs that are allocated to the channel $m$ at $t$ with total number of $N_m^t(\boldsymbol{\kappa^t})$}\\

\hline
\end{tabular}
\end{center}
\vspace{-0.6cm}
\end{table}

\section{Problem Formulation}
\label{section: problem formulation}
We first formally describe our proposed system model and scenario. Then, we present the communication system and give the achievable transmission rates. Following that, we introduce the UL and DL objectives and piece them together and present the overall objective function of the communication model.

\subsection{System Model}
\label{sec:scenarios}
Consider the \textbf{asynchronous} \mbox{UL-DL} transmission system of $n$ XUs ($\mathcal{N}=\{1,2,\ldots,N\}$) sharing a MC with $m$ channels ($\mathcal{M} = \{1,2,\ldots,M\}$) within a building. A slotted time structure is applied by using the clock signal from the MC for synchronization~\cite{chenxudecentralized}, and each slot (transmission time interval (TTI)) contains two sub-intervals: uplink transmission time interval (UTTI) and downlink transmission time interval (DTTI). Each XU $n \in \mathcal{N}$ has its own data buffer, in which the XU will note how much data remains to be transmitted from the XU to the MC for digital replication. We assume that the data sizes of the images to be transmitted can be partitioned into subsets of any size~\cite{anysize}. For each UL, we set the UTTI as $\tau_u$ (in ms denoting milliseconds) to transmit data to the MC. At the start of each TTI, the MC will collect XU's information about the remaining data and channel gain, and then decide which one will offload data, arranging the channel access for them. Since the control information can be reliably completed through the in-band channel, and the size of control information is relatively small, we neglect the time for this control-message transfer, and only consider each TTI containing one UTTI and one DTTI. As we design our scenario to have many more users as compared to MC channels, some users may be arranged with no channel to upload data at each slot, so as to avoid channel congestion. And then, they need to wait to be allocated with channels to transmit data in the following slots. We assume that the uplink transmission powers are decided by the XUs' devices, and they are not optimized in this paper. Consequently, the energy consumption in the UL stage is not considered explicitly. In addition, it can be understood that by reducing the number of UTTIs used to transmit the whole data, we are also reducing UL energy consumption.

In the DL stage, the MC will select transmission power for users assigned with a channel in the current transmission iteration, to facilitate the data DL process. We set the limit of the downlink transmission time interval (DTTI) as $\tau_d$ (in ms). Assume that ${d'}_n^t$ is the transmission delay of XU $n$ in DL at time step $t$, if the DL delay exceeds the $\tau_d$ (${d'}_n^t > \tau_d$), we regard this circumstance as an XU's failed attempt to receive the DL data. If the data transmission for that round is considered a failed attempt, the data will have to be re-scheduled to be transmitted in the subsequent rounds.

To simplify the problem, we assume that the UL and DL processes are carried out alternately without overlap, and we use $\mathcal{T}=\{1,2,\ldots,T\}$ as the notation for the iterations taken to complete the tasks. It should be noted that each iteration contains both a UL and a DL process.

\begin{figure*}[t]
    \centering
    \includegraphics[width=0.9\linewidth]{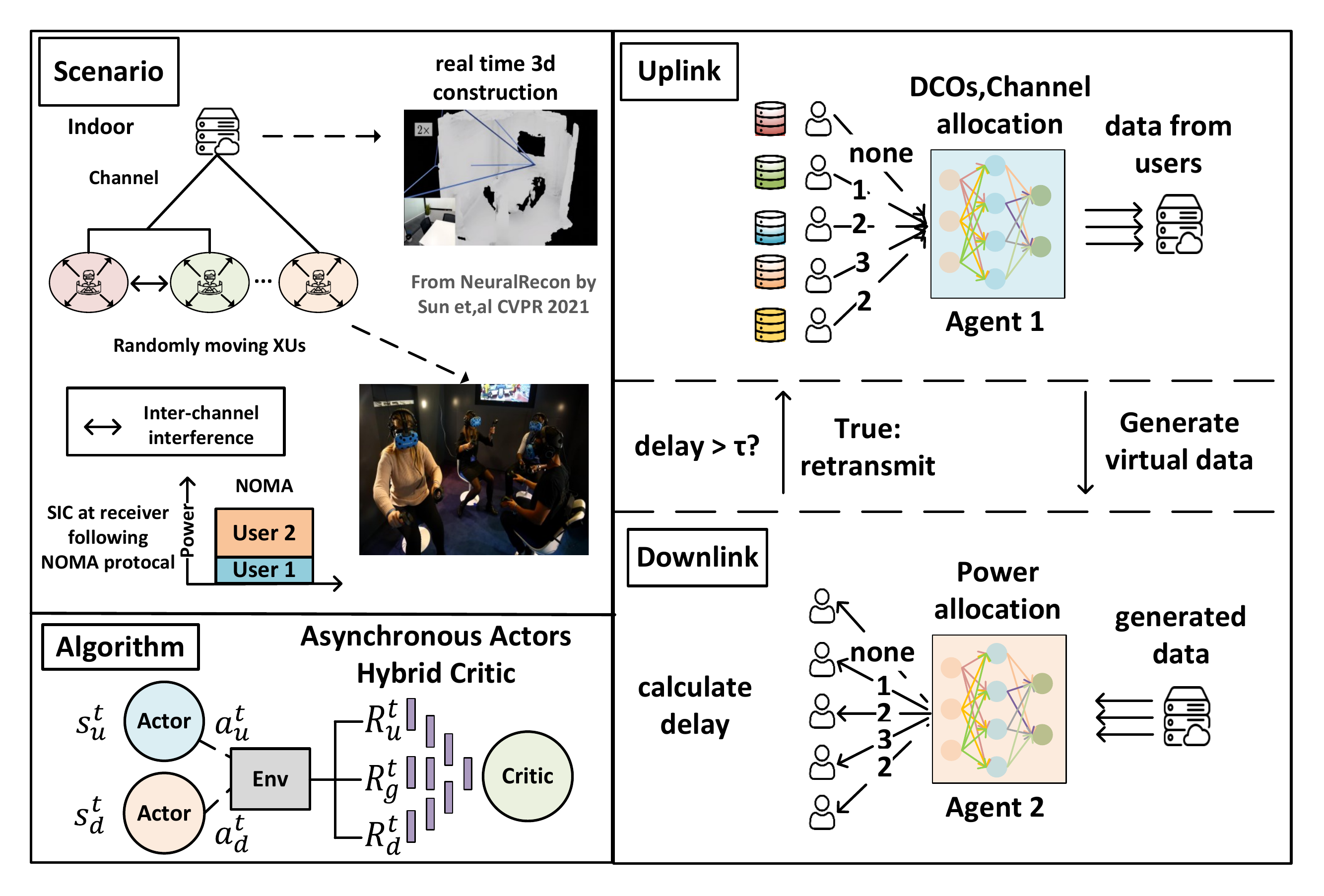}
    \caption{Our proposed system model. The top left of the figure illustrates our problem scenario, while the bottom left showcases a snippet of our proposed algorithm. The right of the figure highlights our transmission mechanism.}
    \label{fig:system}
    \vspace{-0.4cm}
\end{figure*}

We next introduce the communication system as well as the models of UL and DL stages, and tie together all descriptions above. The system model is shown in Fig.~\ref{fig:system}.

\subsection{Communication system}
We adopt the Non-Orthogonal Multiple Access (NOMA) technology for the communication model. In NOMA system, several users can be multiplexed on one channel by the successive interference cancellation (SIC) and superposition coding~\cite{ding2017survey}. As there are more XUs sharing fewer channels, we apply NOMA signal interference structure for both UL and DL transmissions, and consider the interference between XUs on the same channel. Below we give detailed descriptions of the achievable rates from each XU to MC in UL, and the rates from MC to each XU in DL.

For each $n\in\mathcal{N}$, we use $z_n^t$ to denote XU $n$'s Decision on Computation Offloading (DCO) at time step $t$. Specifically, $z_n^t=1$ means that XU $n$ will offload the computation to MC at time step $t$, whereas  $z_n^t=0$ means that XU $n$ will idle at time step $t$. Then we define a row vector $\boldsymbol{z}^t=[z_1^t, z_2^t, \ldots, z_N^t]$ to represent all $N$ XUs' DCO at time step $t$. We further define a $N\times T$ matrix $\boldsymbol{Z}$ such that the $t$th row is the vector $\boldsymbol{z}^t$ for $t \in \{1,2,\ldots,T\}$; i.e., the $t$th row and $n$th column of $\boldsymbol{Z}$ is $z_n^t$.

We let $\boldsymbol{\kappa}^t = [\kappa_1^t, \kappa_2^t, \ldots, \kappa_N^t]$ be the channel access arrangement, and $\boldsymbol{p'}^t=[{p'}_1^t, {p'}_2^t, \ldots, {p'}_N^t]$ be the downlink power allocation, where $\kappa_n^t=m$ ($m\in \mathcal{M}$) means XU $n$ is allocated to channel $m$ at time step $t$, and ${p'}_n^t$ is the power that the MC uses to communicate with XU $n$ at time step $t$. The uplink transmission power is $\boldsymbol{p}=[p_1, p_2, \ldots, p_N]$. Since we do not optimize $\boldsymbol{p}$ for the uplink, we consider $\boldsymbol{p}$ invariant with respect to time $t$ for simplicity (the extension to time-variant $\boldsymbol{p}$ just involves using more complex notation). 

In the NOMA system, we follow~\cite{NOMA} to consider that the decoders at the MC and XUs can recover the received signals from each channel through SIC, and multiple VUs can be multiplexed on each channel by superposition coding. The decoding sequence follows the rule explained in~\cite{NOMA}. To better illustrate the decoding protocols, we define $\mathcal{N}_m^t(\boldsymbol{\kappa}^t)$ as the set of $N_m^t(\boldsymbol{\kappa}^t)$ XUs that are allocated to communicate with MC through channel $m$ at time slot $t$, where ``$(\boldsymbol{\kappa}^t)$'' represents that $\mathcal{N}_m^t $ and $N_m^t$ are functions of $\boldsymbol{\kappa}^t$. Specifically,
\begin{align}
    \mathcal{N}_m^t(\boldsymbol{\kappa}^t):=\{n\in\mathcal{N}| \kappa_n^t=m\}, \text{ and }N_m^t(\boldsymbol{\kappa^t})=|\mathcal{N}_m^t(\boldsymbol{\kappa}^t)|.
\end{align}
\indent
For the UL NOMA, we order the XUs of $\mathcal{N}_m^t(\boldsymbol{\kappa}^t)$ in the descending order of the MC's received signals~\cite{NOMA} sent from these XUs. After the above process, we denote the resulting indices of XUs in $\mathcal{N}_m^t$ by $\Bar{u}_1,\Bar{u}_2,\ldots, \Bar{u}_{N_m^t}$, where we simplify $\mathcal{N}_m^t(\boldsymbol{\kappa}^t)$ and $N_m^t(\boldsymbol{\kappa}^t)$ as $\mathcal{N}_m^t$ and $N_m^t$. Then we have 
\begin{align}
    p_{\Bar{u}_1}|h_{\Bar{u}_1,m}^t|^2 \geq p_{\Bar{u}_2}|h_{\Bar{u}_2,m}^t|^2 \geq \ldots \geq p_{\Bar{u}_{N_m^t}}|h_{\Bar{u}_{N_m^t},m}^t|^2,
\end{align}
where $h_{n,m}^t$ is the channel gain between XU $n$ and the MC on channel $m$ at time step $t$ (this paper considers that uplink and downlink channel conditions are the same). Note that as each TTI is very short, we assume that the channel gain remains the same in each TTI (one UL plus one DL transmission), but varies across different TTIs. The detailed setting of $h_{n,m}^t$ will be explained in Section~\ref{section:num-set}. According to~\cite{NOMA}, for $n \in \mathcal{N}_m^t$, the achievable \textbf{uplink rate} $r_n^t$ for XU $n$ is:
\begin{align}
    r_n^t(\boldsymbol{z^t},\boldsymbol{\kappa}^t) \hspace{-2pt} = \hspace{-2pt} W_m\log_2 \hspace{-2pt} \left( \hspace{-2pt} 1 \hspace{-2pt} + \hspace{-2pt} \frac{p_n|h_{n,m}^t|^2}
    {\sum_{j=\iota+1}^{N_m^t}p_{\Bar{u}_j}|h_{\Bar{u}_j,m}^t|^2\hspace{-2pt} + \hspace{-2pt} W_m(\sigma_{MC}^t)^2} \hspace{-2pt} \right)\hspace{-2pt} ,
    \label{eq:ulrate}
\end{align}
where $\iota$ satisfies $\Bar{u}_{\iota}=n$ (i.e., XU $n$ is ranked at $\iota$th position in $\mathcal{N}_m^t$ according to descending order of received signals at the MC). $W_m$ denotes the bandwidth of channel $m$, and $(\sigma_{MC}^t)^2$ is the power spectral density of Additive White Gaussian Noise (AWGN) at the MC.

For the DL, we also order XUs in a descending order based on channel-to-noise ratios, and denote the resulting indices of XUs using channel $m$ as $\Bar{d}_1,\Bar{d}_2,\ldots,\Bar{d}_{N_m^t}$, satisfying~\cite{NOMA}:
\begin{align}
    \frac{|h^t_{\Bar{d}_1,m}|^2}{(\sigma_{\Bar{d}_1,m}^t)^2}\geq 
    \frac{|h^t_{\Bar{d}_2,m}|^2}{(\sigma_{\Bar{d}_2,m}^t)^2}\geq \ldots \geq \frac{|h^t_{\Bar{d}_{N_m^t,m}}|^2}{(\sigma_{\Bar{d}_{N_m^t,m}}^t)^2}.
\end{align}
Then the achievable \textbf{downlink rate} ${r'}_n^t$ for XU $n$ is:
\begin{align}
    {r'}_n^t\hspace{-1pt}(\boldsymbol{z^t},\hspace{-1pt}\boldsymbol{\kappa}^t,\hspace{-1pt}\boldsymbol{p'}^t) \hspace{-2pt}=\hspace{-2pt}
    W_m\hspace{-1pt}\log_2 \hspace{-2pt}\left(\hspace{-2.5pt} 1\hspace{-2pt}+\hspace{-2pt}\frac{{p'}_n^t|h_{n,m}^t|^2}
    {\sum\limits_{j=1}^{\iota'-1}\hspace{-1pt}{p'}_{\Bar{d}_{j}}|h_{n,m}^t|^2\hspace{-2pt}+\hspace{-2pt}W_m(\sigma_{n,m}^t)^2}\hspace{-2.5pt}\right)\hspace{-2.5pt},
    \label{eq:dlrate}
\end{align}
where $\iota'$ satisfies $\Bar{d}_{\iota'}=n$ (i.e., XU $n$ ranks $\iota'$th in channel $m$ in the descending order of channel-to-noise ratios), and $(\sigma_{n,m}^t)^2$ denotes the power spectral density of Additive White Gaussian Noise (AWGN) of XU $n$ in channel $m$ at time step $t$. Next, we will expound on the formulated problem.

\subsection{Uplink}
An important objective of the UL stage is to reduce the total delay in transmitting the data in all XUs' buffers to the MC. As described in Section~\ref{sec:scenarios} above, we aim to ensure the successful transmissions of these keyframes. Hence if the DL delay exceeds the DTTI limit $\tau_d$ (i.e., if ${d'}_n^t > \tau_d$), that set of data transmitted in that iteration is nullified and has to be re-transmitted. Re-transmissions will increase overall delay (i.e., the number of TTIs used for finishing the whole task), and it is considered as a system \textbf{unreliability}. This is because in real applications, more transmissions may cause more packet losses, more energy usage and longer delay. Hence, the UL transmission delay is influenced by the DL power allocation, because poor actions selected by the DL agent ($Agent_2$) will cause re-transmission. We let a UL agent ($Agent_1$) arrange XUs to the available channels. Considering that this is a multi-stage sequential optimization problem, an XU $n$ which is not arranged with any channel at time step $t$ will not partake in both UL and DL stages, contributing no interference to other XUs.

With the UL transmission rate $r_n^t(\boldsymbol{z^t},\boldsymbol{\kappa}^t)$ (abbreviated as $r_n^{t}$ below) defined in Eq.~(\ref{eq:ulrate}), the data transmitted from XU $n$ to MC at time step $t$ is:
\begin{align}
    D_n^t = \min (B_n^t, r_n^{t} \times \tau_u), ~\forall n\in\mathcal{N},
\end{align}
where $B_n^t$ means the remaining data of XU $n$ at time step $t$ to be transmitted.
This set of data will be digitized on the MC and sent back to XUs in the DL stage. The size of the rendered data is formulated as ${D'}_n^t=f_n(D_n^t)$. The function $f_n(\cdot)$ is the data size translation and rendering formula from 2D to 3D. The expression of $f_n(\cdot)$ depends on the specific Metaverse application that XU $n$ executes with the help from the MC.

The 3D real-time construction has already been applied with the local computation methods and obtained impressive performance. Even a single home edition GTX 2080 can handle this task in milliseconds~\cite{realtime3D1}, but we consider a server with much more computing power. Thus, this paper mainly studies the communication problem of this scenario, considering the asynchronous \mbox{UL-DL} communication, and seeks near-optimal channel resources and power allocation. We take the computation resources at the MC as sufficient and assume the digital replication execution time on the MC as negligible. Thus, the sub-target of UL agent in each iteration $t$ is to maximize the UL sum rate:
\begin{align}
    &O_u:\max\limits_{\boldsymbol{z}^t, \boldsymbol{\kappa}^t} \left ( \sum_{n\in \mathcal{N}} r_n^t \right), \\
    &s.t.~~\kappa_n^t \in \{1,2,\ldots,M\}, ~\forall n\in\mathcal{N},~\forall t\in\mathcal{T}, \nonumber \\
    &~~~~~~z_n^t=1, ~\forall n\in\mathcal{N},~\forall t\in\mathcal{T}. \nonumber
\end{align}

\subsection{Downlink}
In the DL stage, $Agent_2$'s main objective is to (i) minimize the total re-transmission counts and (ii) minimize energy spent for the DL transmission. As described in the previous section (\ref{sec:scenarios}), to ensure the reliability of the whole system, if the DL delay exceeds the DTTI limit $\tau_d$ (${d'}_n^t > \tau_d$) permitted for downlink, that set of data transmitted in that iteration is nullified and has to be re-transmitted. 

In contrast to the fixed uplink transmission power  in the UL stage, we adopt the variable $\boldsymbol{p'}^t=[{p'}_1^t, {p'}_2^t,\ldots,{p'}_N^t]$ as the power allocated to each XU by $Agent_2$ in the DL stage to be optimized. Therefore, the transmission delay ${d'}_n^t$ of each XU in DL stage is represented as:
\begin{align}
   {d'}_n^t = \frac{{D'}_n^t}{{r'}_n^t} = \frac{f_n(D_n^t)}{{r'}_n^t}, 
\end{align}
\noindent where ${r'}_n^t$ is short for the downlink transmission rate ${r'}_n^t(\boldsymbol{z^t},\boldsymbol{\kappa}^t,\boldsymbol{p'}^t)$. We note that each XU $n$ will have a different DL delay ${d'}_n^t$. To simplify the problem of having a variable start time for each subsequent transmission, we assume that all XUs have the same UTTI and DTTI limit synchronized by the MC, and the next UL turn will start immediately after all XUs finish the current DL transmission. 


In the desirable event that ${d'}_n^t$ doesn't exceed the DTTI limit $\tau_d$, this transmission is considered successful, and the remaining data in the XUs' buffers are as shown:
\begin{align}
    B_n^t = B_n^{t-1} - (1-I_n^t)D_n^t.
\end{align}
where $I_n^t$ is the failure flag:
\begin{align}
    I_n^t = 
    \begin{cases}
        0, &\text{if}~~{d'}_n^t \leq \tau_d. \\
        1, &\text{if}~~{d'}_n^t > \tau_d.
    \end{cases}
\end{align}
In addition, the energy $E^t$ used at time step $t$ is
\begin{align}
    E^t = \sum_{n \in \mathcal{N}:z_n^t=1} {p'}_n^t \times \min({d'}_n^t, \tau_d).
\end{align}

The building floor-plan is taken to be a rectangle with length $X$ and width $Y$, and the center of the rectangle has coordinates $(0,0)$. The location of XU $n$ is designed to be confined to the space. 
Therefore, the sub-targets of the DL agent in each iteration $t$ is to minimize the number of XUs exceeding the pre-defined DL time limit $\tau_d$ and MC energy expenditure:
\begin{align}
    &O_d:\min\limits_{\boldsymbol{p'}^t} \left ( w_n \sum_{n\in\mathcal{N}}I_n^t + w_e E^t \right ). \label{eq:dltarget}\\
    &s.t.~~{p'}_n^t \in [{p'}_{\min},{p'}_{\max}],~\forall n\in\mathcal{N},~\forall t\in\mathcal{T}. \nonumber
\end{align}
where $w_n$ and $w_e$ are the weights of each sub-target, and they are represented by the reward instead of being directly set as constants in our proposed DRL algorithm.

\subsection{Overall}
To sum up, our objectives are to minimize the total time used to complete the whole UL and DL processes, and to minimize the energy spent in the DL transmission. In our global objective, we do not include the energy spent on UL as we defined the UL output power as fixed. Therefore, the global objective can be written as:
\begin{subequations}
\begin{align}
    &\min\limits_{\boldsymbol{z}^t,\boldsymbol{\kappa}^t,\boldsymbol{p'}^t,T} 
    \Bigg\{w_1 \left [T \times \tau_u + \sum_{t=1}^{T} \min\left (\max\limits_{n\in\mathcal{N}}({d'}_n^t),\tau_d\ \right ) \right ]+\label{obj:eq1}\\
    &~~~~~~~~~~~~~~~w_2\left [\sum_{t=1}^{T}\sum_{n \in \mathcal{N}:z_n^t=1} {p'}_n^t \times \min ({d'}_n^t, \tau_d) \right ]+ \label{obj:eq2}\\
    &~~~~~~~~~~~~~~~w_3 \left [\sum_{t=1}^T\sum_{n\in\mathcal{N}}I_n^t\right] \Bigg\} \label{obj:eq3}\\
    &s.t.~\text{C1}:\sum_{t=1}^{T}(1-I_n^t)D_n^t=B_n^0, ~\forall n \in \mathcal{N}, \\
    &~~~~~\text{C2}:{p'}^t_n \in [{p'}_{\min}, {p'}_{\max}], ~\forall n\in\mathcal{N},~\forall t\in\mathcal{T}, \\
    &~~~~~\text{C3}:z_n^t \in \{0,1\}, ~\forall n\in\mathcal{N},~\forall t\in\mathcal{T},\\
    &~~~~~\text{C4}:\kappa_n^t \in \{1,..,M\}, ~\forall n\in\mathcal{N},~\forall t\in\mathcal{T},
\end{align}
\end{subequations}
where
\begin{align}
    {d'}_n^t = &\frac{f_n(\min(B_n^t, r_n^t \times \tau_u))}{{r'}_n^t}.
\end{align}
The term in (\ref{obj:eq1}) refers to the total time taken in \mbox{UL-DL} iterations to complete the task. The term in (\ref{obj:eq2}) denotes the energy spent in the DL transmission. The term in (\ref{obj:eq3}) is the transmission failure count. Note that we do not include the UL energy consumption in the formulated problem, because we fix the UL transmission power and time in each iteration. Also, minimizing the number of iterations can be understood as implicitly minimizing the energy consumption in UL. All the weights ($w_1,w_2,w_3$) will be reflected in the reward settings in Section~\ref{sec:rewardsetting} instead of being specific constants.

Constraint C1 means the whole system will finish with $T$ TTIs, where $T$ is also a variable in our optimization problem. In practice, we let the episode (one episode includes the entire execution of the above-mentioned process) continue if there is still data in any XU'
s data buffer to fulfill this constraint. C2 is the range of downlink transmission power allocated by MC for communicating with each XU. This constraint will be satisfied by setting the DL agent action space in practice. C3 and C4 specify the domains for the decisions on computation offloading and channel resource arrangement, respectively, which will be guaranteed by designing the UL agent action space. According to the above-formulated problem, we propose an innovative model-free reinforcement learning method. The reasons for not using convex optimization techniques and model-based reinforcement learning strategies are explained as follows.


\textbf{RL over Convex Optimization: }\label{cvx} \textbf{Firstly}, our defined problem scenario aims to be reflecting ``true-to-real world" . As such, our problem formulation (objective function and constraints) is not naturally convex. There are three options we considered: (1) \textit{Redefining our problem as a convex one. }This would certainly make the defined problem much easier to solve, but it does not represent an accurate model of the real world. A clear example of problem redefinition for the proposed problem would be the \textit{constraint relaxation} of the discrete MC-XU allocation into a continuous one. (2) \textit{Finding an approximate solution to the non-convex problem.} Solution will then only be an approximate solution and it is difficult to gauge how far it is from the optimal solution. (3) \textit{Adopting deep reinforcement learning techniques. } Deep reinforcement learning (RL) techniques may not provide the optimal solution. However, with sufficient model training, the RL agents are able to handle complex non-convex and sequential problems and provide near-optimal solutions. \textbf{Secondly}, the XUs' remaining data at each transmission iteration is sequential (collective) and changing, which makes the number of variables increase with $T$. This increases the solution search space indefinitely, rendering convex optimization techniques or heuristic search as infeasible approaches. Moreover, the discrete variables (DCOs and channel access) and the continuous variable (power) are highly coupled, which causes an Inseparable Mixed-integer Non-linear programming (MINLP) problem. This is NP-hard~\cite{NPhard} and tough to tackle.

\textbf{Model-free over Model-based RL: }\label{MBRL} Model-based (MB) mainly differs from model-free (MF) RL in that model-based requires the specification of the entire model, such as the transition probabilities. In other words, MB RL algorithms use a predictive model to select the optimal actions, whereas MF RL algorithms involve the training of a control policy. MB RL algorithms have brilliant performance in many scenarios. However, due to the predictive model requirement, these algorithms are hard to implement in communication problems. There are too many random variables and \mbox{unpredictable} changes in communication models. The model-based methods cannot accommodate the randomly evolving environment and are impractical for implementation due to the daunting computational complexity. This is also another reason why convex optimization approaches are not suitable for our proposed scenario.

Next, we demonstrate how the problem formulated above can be transformed into an RL problem.

\section{Reinforcement Learning Setting}
Deep reinforcement learning is a state-of-the-art technique to solve time-sequential optimization problems with randomly evolving environments. In this section, we will dissect our previously proposed problem and present the RL approach to solving it. The goal of a model-free RL algorithm is to find a near-optimal policy $\pi^*(a|s)$ of the sequential Markov Decision Process (MDP), which can guide the agent to select the best action $a$ under state $s$. This can also be extended to the multi-agent scenario. When designing an RL environment, we will set rewards to guide the agents to find a near-optimal solution to our problem. Thus, in the subsequent subsections, we discuss our design of the most important parts of RL: state design, action formulation, reward decomposition, and their reflection on our environment setting.
\label{environment}

\subsection{State} \label{sec:statesetting}
Although sophisticated states provide the agent with more information and a more comprehensive view, complex states can result in erratic training. Therefore, the amount of information to be perceived by each XU needs to be limited. Hence, weeding out less relevant variables is essential. Key attributes to include in the states are those which are collective or sequential. Therefore, the state $s_u^t$ in UL and state $s_d^t$ in DL stages are as follows:\\

\subsubsection{Uplink State $s_u^t$}
(i) The remaining data in each XU's buffer $B_n^t$, (ii) the UL transmission power of each XU's device $p_n$, and (iii) channel gain of each XU $n$ at TTI $t$: $h_{i,m}^t$.\\

\subsubsection{Downlink State $s_d^t$}
(i) The action by UL agent (combined by DCOs $\boldsymbol{z}^t$ and channel access management $\boldsymbol{\kappa}^t$, which will be explained in the following part), (ii) The data buffer of each XU $B_n^t$, to equip the DL agent the overall view of the remaining data to be transmitted, (iii) The data size after generated and rendering: ${D'}_n^t$, and (iv) channel gain in current TTI $h_{i,m}^t$.

\subsection{Action}
The action space defines the boundaries of possible actions our agents can take. An appropriate and relevant setting, as per our earlier discussed optimization variables, is crucial. In our scenario, the action $a_u^t$ and action $a_d^t$ are respectively uplink action and downlink action, explained below.  \\

\subsubsection{Uplink Action $a_u^t$}
The discrete action $a_u^t $ in UL includes the DCOs and channel assignment:
\begin{align}
    &a_u^t = \{\boldsymbol{z}^t;\boldsymbol{\kappa}^t\}.
\end{align}
For discrete action space RL, we need to encode the actions into discrete indicators. Thus, we integrate the DCO and channel assignment as: $\Gamma_n^t=\{0,1,2,\ldots,M\}$. $\Gamma_n^t=0$ means $z_n^t=0$ (i.e., XU $n$ does not offload the computation to MC at time step $t$). And $\Gamma_n^t=m$ for $m\in\mathcal{M}$ means $n$ is allocated to channel $m$.

We use a tuple in which there are $N$ elements corresponding to $N$ users and each element can take \mbox{$M+1$}  values, which corresponds to the number of MC channels, plus~1 for the possibility of a user not being assigned a channel. In practice, the discrete actions are in fact discrete indices to the DRL Agent. Therefore, we need to allocate consecutive but distinct indexes to each action. The most intuitive way of implementing this is to use $N$-bit-$(M+1)$-number to denote the tuple, and convert it into a decimal action index (similar to binary to decimal):
\begin{align}
    a_u^t = \sum_{n=1}^{N}\Gamma_n^t(M+1)^{n-1},~\forall t\in \{1,\cdots,T\}.
\end{align}

\begin{figure}[t]
    \centering
    \includegraphics[width=0.9\linewidth]{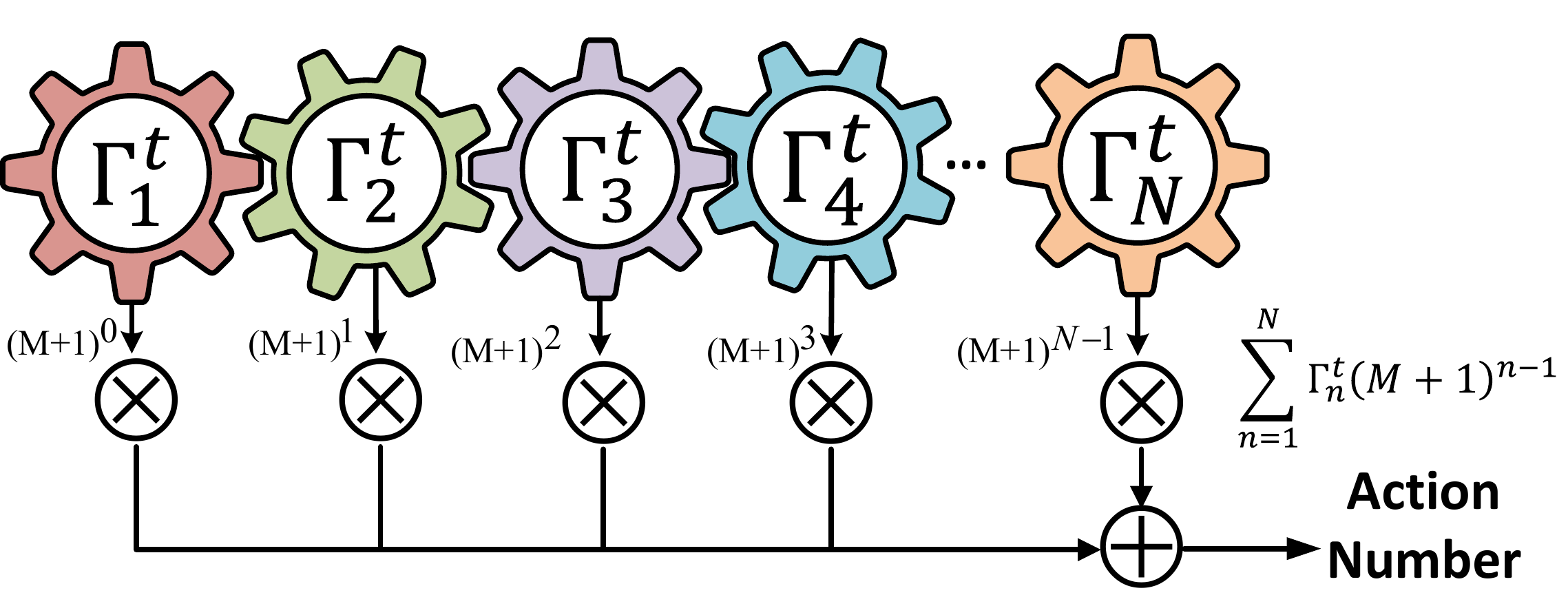}
    \caption{The encoding method for the uplink (UL) action.}
    \label{fig:actionencode}
    \vspace{-0.4cm}
\end{figure}

The encoding method is shown in Fig.~\ref{fig:actionencode}.

\subsubsection{Downlink Action}
On the other hand, the continuous action power allocation to each XU is:
\begin{align}
  a_d^t = \{{p'}_1^t,{p'}_2^t,\ldots,{p'}_N^t\},~\forall{p'}_n^t \in [{p'}_{\min}, {p'}_{\max}].  
\end{align}

\subsection{Reward} \label{sec:rewardsetting}
In our scenario, the two agents not only have their own role-specific goals, but also have to fulfill a global target that works in the combined best interest of both agents. In our proposed scenario, we can observe some trade-offs: (i) While $Agent_1$ aims to fulfill its task of uploading the data to MC in the shortest time possible, it has to observe the data downlink speed to ensure that the uploaded data size can be managed by the DL agent in the downlink transmission. (2) While $Agent_2$'s objective is to minimize the DL transmission delay, it has to manage the energy spent for transmission at the same time. Due to these trade-offs between and within agents, we have to establish a unified priority by introducing a global target that serves as the global information to both agents. In our proposed work, minimizing the overall total time spent on UL and DL transmission (including re-transmissions) are related to both agents, and hence, are adopted as the global target. We also let the proposed algorithm learn the global reward separately with a global branch (which will be discussed in Section~\ref{sec:AAHC}), to give the networks a more specific view of the global process.

Therefore, we have chosen to adopt both role-specific rewards and global rewards for our scene. We introduced additional self-defined rewards to guide the agents' training, as sparse rewards detract the training progress of RL agents. In practice, the rewards are structured as follows:\\

\subsubsection{Uplink Reward}
$R_u^t$ encompasses: (i) Upload efficiency penalty $R_{ur}^t$: a penalty will be given in each UL transmission in which a lower data transmission rate results in a bigger penalty.\\

\subsubsection{Downlink Reward}
$R_d^t$ encompasses: (i) Download efficiency penalty $R_{dr}^t$: a larger downlink delay results in a larger penalty. (ii) Energy expenditure penalty $R_{ene}^t$: a penalty will be issued to the agent for the expense of energy. (iii) Power allocation guide $R_{i,gu}^t$: a small penalty will be given when the XU that is not allocated a channel is assigned a power greater than 0. \\

\subsubsection{Global Reward}
\label{section:globalreward}
$R_g^t$ encompasses: (i) total delay penalty $R_{ite}^t$: a larger penalty will be added to every transmission iteration for both UL and DL agents, and  (ii) Re-transmission penalty $R_{i,re}^t$: Higher transmission failures and hence, higher re-transmission counts result in a larger penalty added to both UL and DL agent.\\ 

In a hybrid reward setting where rewards of multiple agents are considered, we recommend standardizing the values across the different contributing rewards to smaller scales, to ease the training. Therefore, note that all the rewards are shaped (weighted) to easy-for-training scales in this paper. The numerical reward settings in this paper for reference are as follows:
\begin{itemize}
    \item \textbf{Uplink} -- upload efficiency penalty:\\
    $R_{ur}^t = -\sum_{n=1}^N \frac{(1-\frac{D_n^t}{B_n^0})}{N}$. This reward denotes the average XUs' transmitted portions of their total data in buffers, and the range of this reward is $[-1,0]$. 
    \item \textbf{Downlink} -- download efficiency penalty:\\
    $R_{dr}^t = -\min(\sum_{n=1}^N\frac{{d'}_n^t}{\tau_d N},1)$. This reward is the average XUs' ratio of the actual required time to tolerant DL transmission delay. The range is $[-1,0]$.
    \item \textbf{Downlink} -- energy penalty:\\
    $R_{ene}^t = -\sum_{n=1}^N\frac{{p'}_n^t-{p'}_{\min}}{({p'}_{\max}-{p'}_{\min})N} \times 0.5$. This is the average XUs' ratio of the allocated power to the maximum DL transmission power. Its range is $[-0.5,0]$.
    \item \textbf{Downlink} -- power allocation guide:\\
    $R_{i,gu}^t = -0.2$ if user $n$ in channel $0$ is allocated to power.
    \item \textbf{Global} -- total system delay penalty:\\
    $R_{ite}^t = -1$ for every iteration.
    \item \textbf{Global} -- re-transmission penalty:\\
    $R_{i,re}^t=-0.5$ for every re-transmission. The re-transmissions also lead to an increased number of iterations. Therefore, we do not give the re-transmission a too large penalty.
\end{itemize}

\section{Methodology}
\label{algorithm}
In this section, we introduce our proposed novel algorithm, Asynchronous Actors Hybrid Critic (AAHC). We will first introduce our inspiration behind this algorithm, and then introduce the preliminary algorithm: Proximal Policy Optimization (PPO). Finally, we will detail the derivation of AAHC from PPO, the structure of AAHC, and discuss the training mechanism.

\subsection{Inspiration}
In our problem definition, we aim to minimize the overall task delay (including re-transmission delay) and transmission energy consumption to fulfill the ``Green Metaverse'' demand while achieving each agent's objectives. Given the multitude of objectives, reinforcement learning agents struggle in achieving them and may fail to achieve convergence. We propose a novel multi-input, multi-objective (output) Critic which aims to be able to handle complex scenarios such as the one presented in our work, and simplify the objectives into bite-size challenges. Our Critic's objectives can be broken down as such: The UL Critic branch guides the UL agent in (i) increasing the UL data transmission rate in every UL. The DL Critic branch guides the DL agent in (ii) decreasing the DL transmission delay and (iii) minimizing MC DL transmission energy consumption. We employ an additional, overarching branch within our Critic which handles the global objective, (iv) to reduce the overall time taken to complete the task (UL and DL), and (v) total energy spent by MC for the DL transmission, as these are important objectives of our system. It is crucial to note that agent policies that minimize transmission failure and re-transmission could potentially assist in decreasing the overall system transmission delay.

Our multi-input, multi-objective Critic is inspired by Hybrid Reward Architecture (HRA)~\cite{van2017hybrid}, which has not been used for solving problems related to wireless communications to the best of our knowledge. HRA can utilize the domain knowledge by decomposing the reward into simpler parts, which has been demonstrated by extensive experiments in multiple fields, e.g., video games~\cite{HRAgame}. However, unlike our proposed AAHC, HRA does not consider the multi-agent setting and the asynchronous interaction between agents. Our reward is decomposed across two agents, and uses a similar way of updating the value network (Critic) by the weighted sum of the loss functions. In the next sub-section, we will introduce the preliminary RL algorithm, Proximal Policy Optimization (PPO), which is chosen as the backbone of our AAHC algorithm.


\subsection{Preliminary: Proximal Policy Optimization (PPO)}
PPO is a state-of-the-art, effective RL algorithm, which has been actively used in wireless communication research\cite{li2020trajectory}. PPO is a suitable RL algorithm to tackle our proposed scenario as PPO can handle both discrete and continuous action spaces through fitting different output heads on the Actor network. Next we will expound PPO, focusing on its three main underlying features: (i) Policy gradient, (ii) Importance sampling and (iii) Policy constrain.

\textbf{Based on policy gradient methods.} The widely used policy gradient method computes an estimator and embeds it into a stochastic gradient ascent algorithm to maximize the expected reward:
\begin{align}
    J(\theta)               &= \sum_\tau \pi_\theta(\tau) R(\tau), \label{eq:1} \\ 
    \nabla_{\theta} J(\theta) &= \sum_\tau \nabla_{\theta} \pi_{\theta}(\tau) R(\tau)  \nonumber\\
                            &= \sum_\tau \pi_{\theta}(\tau) \nabla_{\theta} log \pi_{\theta}(\tau) R(\tau)  \nonumber \\ 
                            &= \mathbb{E}_{\tau\sim\pi_\theta}[\nabla_{\theta} log\pi_\theta(\tau) R(\tau)], \label{eq:2}
\end{align}
where $\pi_\theta$ is a stochastic policy, $R(...)$ denotes the reward function, and $\tau$ denotes the trajectories including $(s^0, a^0, ..., s^t, a^t)$. Recent works use an advantage function instead of a reward function to make training more stable. Thus, we rewrite Eq. (\ref{eq:1}) into:
\begin{align}
    J(\theta) = \mathbb{E}_{\tau\sim\pi_\theta}[A(\tau)]. \label{eq:3}
\end{align}
Note that this expected value $\mathbb{E}=[...]$ represents the average value of the sampled data.

\textbf{Use of importance sampling.} Importance sampling (IS)\cite{owen2000safe} is a method where an expectation with respect to a target distribution is approximated from another distribution. Hence, IS is an important trick adopted in PPO as it allows PPO to use different policies for sampling and evaluating trajectories, increasing the overall sample efficiency~\cite{schulman2017proximal}.

Here we use $\pi_\theta$ as the policy for evaluation and $\pi_{\theta'}$ as the policy for sampling data for training, and Eq. (\ref{eq:3}) can be rewritten as:
\begin{align}
    \mathbb{E}_{\tau\sim\pi_\theta}[A(\tau)]= \mathbb{E}_{\tau\sim\pi_{{\theta'}}} \left [\frac{\pi_\theta(\tau)}{\pi_{{\theta'}}(\tau)} A(\tau) \right ]. \label{eq:5}
\end{align}
However, in practice, we use state-action pairs instead of trajectories to update the gradient. Thus, the objective function of the Actor can be written as:
\begin{align}
    J(\theta)& =\mathbb{E}_{(s^t,a^t)\sim\pi_{{\theta'}}} \left [\frac{\pi_\theta(s^t,a^t)}{\pi_{{\theta'}}(s^t,a^t)} A^t \right ], \nonumber\\
    &\approx \mathbb{E}_{(s^t,a^t)\sim\pi_{{\theta'}}} \left [\frac{\pi_\theta(a^t|s^t)}{\pi_{{\theta'}}(a^t|s^t)} A^t \right ],
\end{align}
where $A^t$ is short for $A(s^t,a^t)$. Note that we use the $\approx$ instead of the = symbol as calculating the exact probabilities of $\pi_{\theta}$ and $\pi_{\theta'}$ is impractical. This is because some states within our proposed scenario occur infrequently. Therefore, we assume that $\pi_{\theta}(s^t) = \pi_{\theta'}(s^t)$~\cite{schulman2017proximal}.

\textbf{Add KL-divergence penalty.}
After switching to $\pi_{{\theta'}}$ for data sampling, there remains an issue of unequal variances. Although Eq.s (\ref{eq:3}) and (\ref{eq:5}) have the same expectations, their variances are very different, as shown below:
\begin{align}
    &Var_{\tau\sim\pi_\theta} [A(\tau)] \nonumber \\
    &=\mathbb{E}_{\tau\sim\pi_\theta}[A^2(\tau)] - (\mathbb{E}_{\tau\sim\pi_\theta}[A(\tau)])^2, \label{eq:var1}\\
    &Var_{\tau\sim\pi_{{\theta'}}} \left [\frac{\pi_\theta(\tau)}{\pi_{{\theta'}}(\tau)} A(\tau) \right ]\nonumber \\
    &=\sum_{\tau\sim\pi_{\theta'}} \frac{\pi_\theta^2(\tau)}{\pi_{\theta'}^2(\tau)} A^2(\tau) \pi_{{\theta'}}(\tau)  - 
    \left (\sum_{\tau\sim\pi_{\theta'}} \frac{\pi_\theta(\tau)}{\pi_{{\theta'}}(\tau)} A(\tau) \pi_{\theta'}(\tau) \right )^2 \nonumber\\
    &=\sum_{\tau\sim\pi_{\theta'}} \frac{\pi_\theta^2(\tau)}{\pi_{\theta'}(\tau)} A^2(\tau) -
    \left (\sum_{\tau\sim\pi_{\theta'}} \pi_\theta(\tau) A(\tau) \right )^2 \nonumber\\
    &=\mathbb{E}_{\tau\sim\pi_\theta} \left [ \frac{\pi_\theta(\tau)}{\pi_{{\theta'}}(\tau)} A^2(\tau) \right ] - (\mathbb{E}_{\tau\sim\pi_\theta}[A(\tau)])^2. \label{eq:var2}
\end{align} 
From the two variances (\ref{eq:var1}) and (\ref{eq:var2}), we can see that the distance between the distributions $\theta$ and $\theta'$ can not be large. Therefore, PPO adds a KL divergence penalty to the Actor objective function to constrain the distance:
\begin{align}
    &J(\theta) = \mathbb{E}_{(s^t,a^t)\sim\pi_{{\theta'}}}[r(\theta) A^t] \label{Actorobj} \\
    &s.t.~~D_{KL}(\pi_{\theta}||\pi_{{\theta'}}) \leq \sigma, \label{trustregion}
\end{align}
where $r^t(\theta)=\frac{\pi_\theta(a^t|s^t)}{\pi_{{\theta'}}(a^t|s^t)}$ is the probability ratio between old and new policies. $D_{KL}(\cdot||\cdot)$ denotes the Kullback--Leibler (KL) divergence for measuring the distance between $\pi_\theta$ and $\pi_{{\theta'}}$.

These strategies make PPO an effective and reliable RL algorithm to tackle wireless communication optimization problems. Nevertheless, this KL divergence is impractical to calculate in practice as this constraint is imposed on every observation. Thus, in~\cite{schulman2017proximal}, the objective function is finally represented as:
\begin{align}
    \mathbb{E}_{(s^t,a^t)\sim\pi_{{\theta'}}}[f^t(\theta, A^t)], \label{eq:obj}
\end{align}
where
\begin{align}
    f^t(\theta, A^t)=\min\{r^t(\theta)A^t,
    clip(r^t(\theta), 1-\epsilon, 1+\epsilon)A^t\}.
\end{align}

The problem in~(\ref{eq:obj}) is solved by gradient ascent, therefore, the gradient is finally written as:
\begin{align}
    \Delta\theta = \mathbb{E}_{(s^t,a^t)\sim\pi_{{\theta'}}}[\nabla_{\theta} f^t(\theta,A^t)]. \label{eq:Actorobj}
\end{align}

\textbf{Critic loss.} In terms of the Critic, PPO uses a Critic with an identical network to the Actor, just like in other Actor-Critic algorithms. We use Generalized advantage estimation (GAE)~\cite{GAE} to calculate the advantage in practice. Thus, the loss function is formulated in~\cite{JO1} as:

\begin{align}
    &L(\phi) = (V_\phi(s^t)-V^t_{target})^2, \label{eq:Criticloss} \\
    &V^t_{target} = A^{GAE}+\gamma V_{\phi'}(s^{t}).
    \label{eq:targetV}
\end{align}

$A^{GAE}$ is the advantage calculated using GAE, which will be explained in the next part, and $V(s)$ is the widely used state-value function~\cite{sutton1999policy}, which is estimated by a learned Critic network with parameter $\phi$. We update $\phi$ by minimizing the $L(\phi)$, and the parameter $\phi'$ of the target state-value function periodically with $\phi$. Using target value is a prevailing trick in RL, which has been used in many algorithms~\cite{mnih2013playing,lillicrap2015continuous}.

\subsection{Asynchronous Actors Hybrid Critic (AAHC)} \label{sec:AAHC}
This paper proposes a novel structure AAHC that selects the state-of-art RL algorithm PPO as the backbone. In our work, we equipped the AAHC with a discrete-action space Actor, a continuous-action space Actor, and a multi-head hybrid Critic. Different from existing algorithms like independent multi-agent RL in~\cite{IPPO} that uses two separate independent agents, and the widely used CTDE-based algorithm like MAPPO~\cite{JO1} that uses the concatenation of states and actions in different stages, AAHC uses two asynchronous Actors and a hybrid Critic with three branches to learn the information in UL, DL, and global stages separately, and better utilizes the domain knowledge from the separate states, actions and rewards in different stages.
\begin{figure*}[h]
    \centering
    \includegraphics[width=0.85\linewidth]{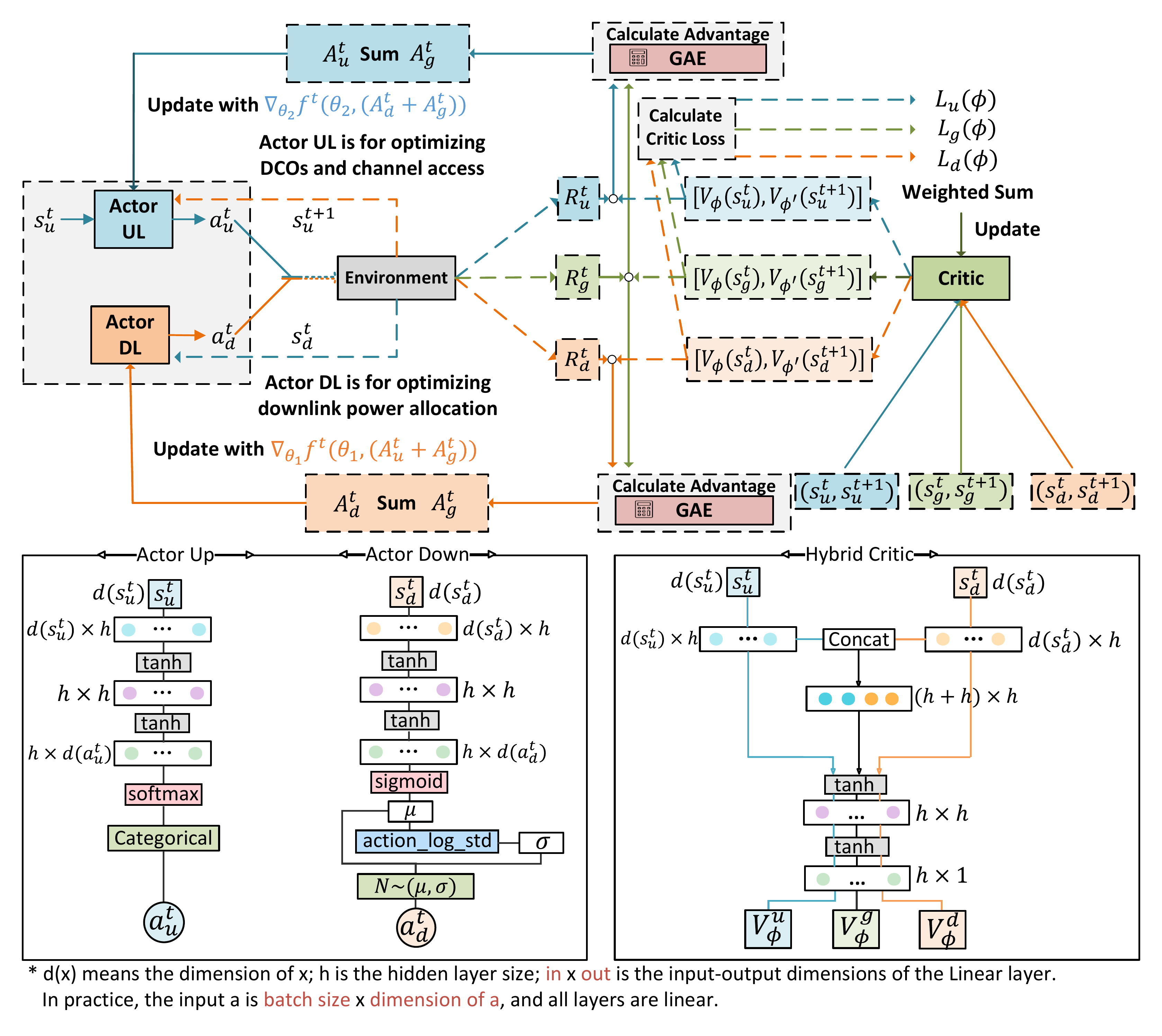}
    \caption{Asynchronous Actor Hybrid Critic (AAHC) structure. Top of the figure illustrates the architecture of AAHC. Below are the networks of the Uplink Actor, Downlink Actor, and Hybrid Critic, respectively.}
    \label{fig:alg}
    \vspace{-0.4cm}
\end{figure*}

\textbf{Function process}: In each episode, the initial state $s_u^0$ will be observed by the UL Actor, which will output the selected action $a_u$. Then, the current UL transmission can be accomplished with the selected $a_u$, and the environment will provide a feedback reward $R_u$ for the action $a_u$ and the DL state $s_d$ at the current stage. Following the UL transmission, the DL Actor will generate the power allocation $a_d$ in the DL stage upon observing $s_d$. The $a_d$ will be acted on the environment to accomplish the DL transmission task, and obtain the DL reward $R_d$ for the choice of $a_d$, and global reward $R_g$ for the whole iteration. These rewards ($R_u,R_d,R_g$) will be used to generate the advantages ($A_u,A_d,A_g$) by GAE~\cite{GAE} for updating the Actors, and the state-values for UL, DL, global and loss functions of Critic will be calculated. This process repeats until the end of an episode. The above-mentioned process is illustrated in Fig.~\ref{fig:alg}. Next, we will explain the mechanisms of how Actors and Critic update, respectively.

\textbf{Asynchronous Actors}:
In AAHC, there are two Actors, one is responsible for making decisions on computation offloading and arranging channel resources in the UL stage, and the other is responsible for allocating transmission power in the DL stage. Apart from their asymmetric task, the action space types for both UL and DL agents, and hence policy parameterizations, are dissimilar. Thus, we adopt the action space methods proposed by Sutton~\cite{sutton1999policy}. We compute learned probabilities for each of the many actions for the Actor in the UL transmission stage, and learn the probability distribution of the policy for the Actor in DL transmission stage. In practice, continuous action values are chosen from a normal (Gaussian) distribution.

Each agent attempts to achieve its own role-specific and global objectives. However, both agents are designed to not have the ability to observe the other's role-specific objective as the success of the other agent's objectives is not within an agent's control. An example would be that $Agent_1$ is rewarded for having an overall short UL transmission time, while the UL transmission time is not within $Agent_2$'s control (which is DL power selection). In this case, allowing $Agent_2$ to receive rewards based on $Agent_1$'s control can be ``confusing" and even ``conflicting" to $Agent_2$'s policy training.

In Eq.~(\ref{eq:Actorobj}), we established the policy gradient for PPO Actor, and in AAHC we have the gradients $\Delta\theta_{1}$, $\Delta\theta_{2}$ of $Agent_1$ and $Agent_2$ as:
\begin{align}
    &\Delta\theta_1 = \mathbb{E}_{(s^t_u,a^t_u)\sim\pi_{{\theta_1}'}}[\nabla_{\theta_1} f^t(\theta_1,(A_u^t+A_g^t)] ,
    \label{eq:gradient1}\\
    &\Delta\theta_2 = \mathbb{E}_{(s^t_d,a^t_d)\sim\pi_{{\theta_2}'}}[\nabla_{\theta_2} f^t(\theta_2,(A_d^t+A_g^t)],
    \label{eq:gradient2}
\end{align}
where $s^t_u$, $s^t_d$ denote the states at time step $t$ in UL and DL stages, respectively. $A_u, A_d, A_g$ are the UL, DL and global advantage functions.

In terms of the \textbf{advantage function}, most techniques compute it with a learned state-value function $V(s)$, and generalized advantage estimation (GAE)~\cite{GAE} is undoubtedly one of the most renowned methods. Running the policy for $\bar{T}$ time steps and using them to update is a widely accepted method popularized in~\cite{Min16}. Following their paradigm, we use the truncated version of GAE as:
\begin{align}
    &A_u^t = \delta_u^t + (\gamma\lambda)\delta_u^{t+1}+...+(\gamma\lambda)^{\bar{T}-1}\delta_u^{t+\bar{T}-1}, \\
    &\text{where}~~~\delta_u^t=R_u^t+\gamma V_{\phi'}(s_u^{t+1})-V_{\phi'}(s_u^t).
\end{align}
and $A_d,A_g$ are as the same. $\bar{T}$ specifies the length of given trajectory segment, and $\gamma$ specifies the discount factor, and $\lambda$ denotes the GAE parameter. 
\\

\textbf{Hybrid Critic}:
In our problem, the objectives of each agent are very different, while there remains an overall arching unified goal. Thus, we propose the Hybrid Critic to aid the estimation of the value function in the hybrid reward problem. This Hybrid Critic has three heads and the value is divided into three parts $V_\phi^u, V_\phi^d, V_\phi^g$. The value function can be estimated by the three-head value network $V_\phi$ as shown below:
\begin{align}
    &V_\phi^u = V_\phi(s_u), \\
    &V_\phi^d = V_\phi(s_d), \\
    &V_\phi^g = V_\phi(\{s_u; s_d\}).
\end{align}
Then, we calculate the three losses $L^u(\phi), L^d(\phi), L^g(\phi)$ with the advantages and values, and sum the weighted losses of each head according to Eq.~(\ref{eq:Criticloss}) and Eq.~(\ref{eq:targetV}) as the loss function of the Hybrid Critic:
\begin{align}
    &L^u(\phi) = (V_\phi(s_u^t)-A_u^t-\gamma V_{\phi'}(s_u^{t}))^2, \\
    &L^d(\phi) = (V_\phi(s_d^t)-A_d^t-\gamma V_{\phi'}(s_d^{t}))^2, \\
    &L^g(\phi) = (V_\phi(\{s_u^t;s_d^t\})-A_g^t-\gamma V_{\phi'}(\{s_u^{t};s_d^{t}\}))^2, \\
    &L(\phi) = w_u \times L^u(\phi) + w_d \times L^d(\phi) + w_g \times L^g(\phi), \label{eq:HCloss}
\end{align}
\hspace{-5pt}
where $\phi$, $\phi'$ are the weights of the network and the target network, and $w_u, w_d, w_g$ are the weights of each head's loss that are all set as  $1$ in the experiments. We update the Hybrid Critic through Eq.~(\ref{eq:HCloss}) to improve the estimate of the hybrid value function. The intricacies of our algorithm are illustrated in Fig.~\ref{fig:alg}. And the algorithm flow is in Algorithm.~\ref{alg:AAHC}.

\begin{figure}[!t] 
        \renewcommand{\algorithmicrequire}{\textbf{Initiate:}}
        \renewcommand{\algorithmicensure}{\textbf{Output:}}
        \begin{algorithm}[H]
            \caption{\label{alg:AAHC} Our proposed Asynchronous Actors Hybrid Critic (AAHC) algorithm.}
            \begin{algorithmic}[1]
                \REQUIRE uplink Actor parameter $\theta_1$, downlink Actor parameter $\theta_2$, Critic parameter $\phi$ and target network $\phi'$, initial state $s_u^0$, $s_u^t \leftarrow s_u^0$;
                \FOR{iteration = $1,2...$}
                    \STATE $Agent_1$ execute action according to $\pi_{\theta_1^{'}}(a_u^t|s_u^t)$;
                    \STATE Get reward $R_u^t$ and next downlink state $s_d^{t+1}$;
                    \STATE $Agent_2$ execute action according to $\pi_{\theta_2^{'}}(a_d^t|s_d^t)$;
                    \STATE Get reward $R_d^t, R_g^t$ and next uplink state $s_u^{t+1}$;
                    \IF{iteration $\geq 2$}
                    \STATE Sample ($s_u^t, a_u^t, R_u^t, s_u^{t+1}, s_d^t, a_d^t, R_d^t, R_g^t, s_d^{t+1}$) iteratively;
                    \ENDIF
                    \STATE $s_d^t \leftarrow s_d^{t+1}$, $s_u^t \leftarrow s_u^{t+1}$;
                    \STATE Compute advantages \{$A_u^t,A_d^t,A_g^t$\} and target values \{$V_{u,targ}^t,V_{d,targ}^t,V_{g,targ}^t$\} using current Hybrid Critic; 
                    \FOR{$k$ = $1,2,\ldots,K$}
                        \STATE Shuffle the data's order, set batch size $bs$;
                        \FOR{$j=0,1,\ldots,\frac{\text{trajectory length}}{bs}-1$}
                            \STATE Compute gradient for uplink, downlink Actors by Eq.~(\ref{eq:gradient1}) and (\ref{eq:gradient2});
                            \STATE Update Actors by gradient ascent;
                            \STATE Update Critic with MSE loss using Eq.~(\ref{eq:HCloss});
                        \ENDFOR
                        \STATE Assign target network parameters $\phi' \leftarrow \phi$ for every $C$ steps;
                    \ENDFOR
                \ENDFOR
            \end{algorithmic}
        \end{algorithm}
\vspace{-0.6cm}
\end{figure}

\section{Experiments}
\label{experiment}

In this section, we conduct several experiments to highlight the outstanding performance achieved by our proposed AAHC algorithm. We compare our algorithm performance against baseline models: (i) iterative independent RL~\cite{IPPO}, (ii) and iterative CTRL framework, based on commonly adopted metrics. The numerical settings and extensive experimental results are illustrated as well.

\subsection{Baseline}
To demonstrate the superiority of our proposed algorithm, we adopt commonly used baseline models. All baseline models are similar to our proposed model in that they contain two separate agents which work asynchronously, and that they are fitted with the PPO algorithm as their backbone.

\textbf{iteRL.}
The most intuitive way of using RL in such a cooperative interactive environment is to implement two single standard iterative RL (iteRL) agents, similar to~\cite{IPPO}. However, in~\cite{IPPO} they use two iterative deep Q networks, while we use two more advanced PPO networks for this baseline. In terms of the reward, we give the UL agent $R_U^t=R^t_u+R^t_g$, and the DL agent $R_D^t=R^t_d+R^t_g$. In other words, these two agents are not entirely separated but are still affiliated through the global reward. This baseline is to testify the performance if we apply no modification to the structure.

\textbf{CTRL.}
The widely used, standard Centralized Training with Decentralized Execution (CTDE) based algorithms like MAPPO~\cite{lowe2017multi} cannot be used right out-of-the-box for our scenario in the execution stage, as the agents in our scenario are required to select actions asynchronously rather than at the same time. In order to compare our proposed methods with the CTDE-based MAPPO, we adapt the Centralized training reinforcement learning (CTRL) algorithm for comparison. We use a centralized Critic with two Actors, and in this baseline, we use the sum of rewards $R=R^t_u+R^t_d+R^t_g$ as the reward received by the Critic, instead of computing three different losses and using the sum-loss for updating the Critic. This baseline is to examine the performance if no hybrid reward structure is embedded in the Critic.

\textbf{Random Allocation. }
In addition to the above-mentioned algorithms, we implement an algorithm that assigns both UL and DL random channel allocation and power selection, respectively. The random channel assignment and power selection algorithm will serve as an intuitive baseline.

\begin{figure}[t]
    \hspace{-15pt}   \includegraphics[width=1.05\linewidth]{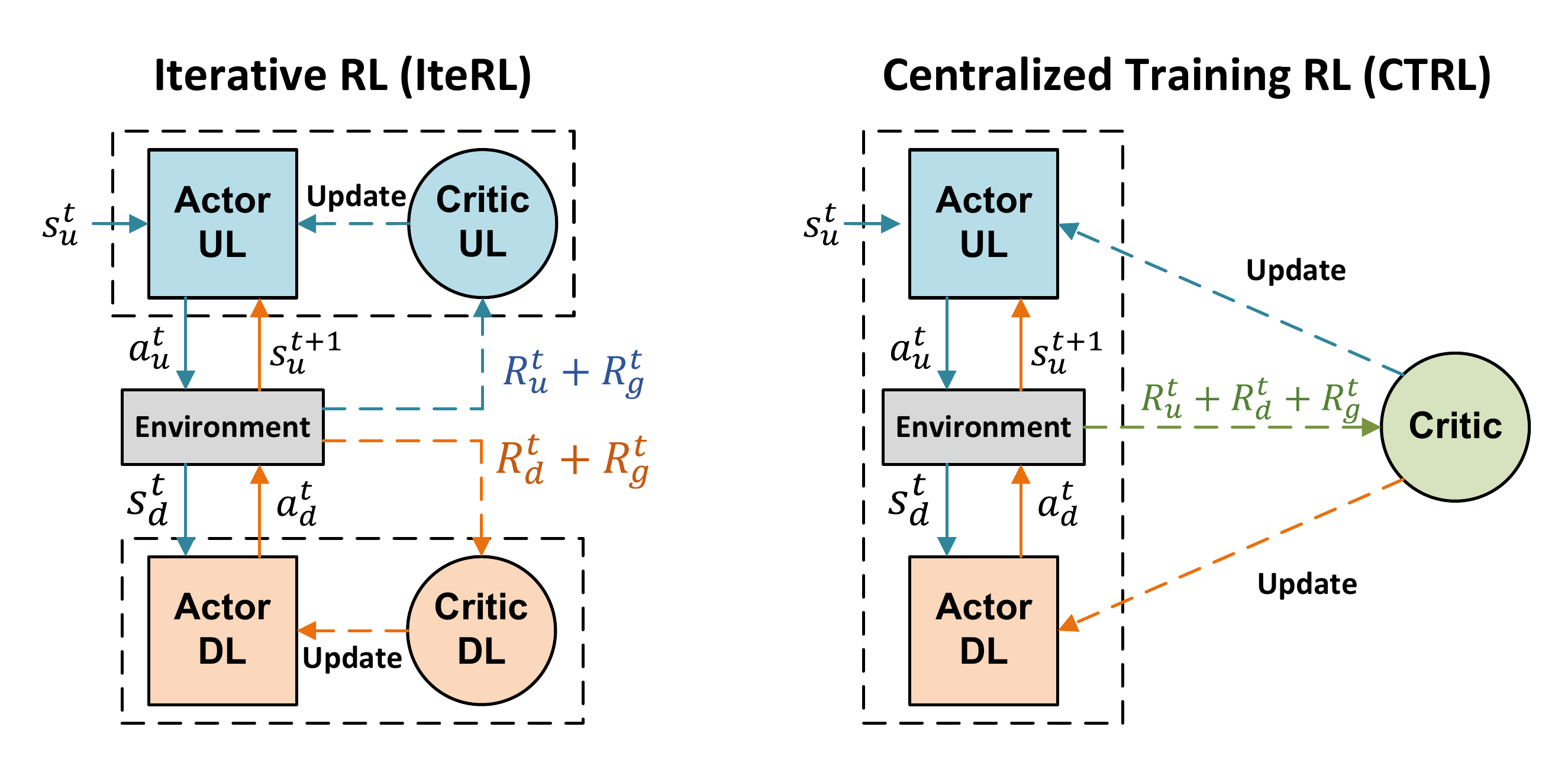}
    \caption{Illustrations of iteRL and CTRL. iteRL uses two separate DRL agents with the UL agent taking in $R_u^t+R_g^t$, and the DL agent using $R_d^t+R_g^t$ to update the networks. CTRL only has one critic, which takes in the \mbox{$R_u^t+R_d^t+R_g^t$} for updating the networks.}
    \label{fig:alg}
    \vspace{-0.4cm}
\end{figure}

\subsection{Metrics (KPIs)}
To fairly compare the performance of the algorithms tackling our proposed scenario, we introduce several commonly-adopted metrics as key performance indicators (KPIs).

Given that we are using an RL approach to tackle our proposed problem, the most obvious performance metric would be the rewards. The objective of DRL algorithms is to obtain as high accumulated rewards in one episode as possible, and the rewards are directly related to multiple objectives, which reflect the algorithms' abilities to handle the proposed problem. Therefore, the training rewards can well serve as an intuitive and overall performance of the algorithms. Specifically, there are three main rewards in our proposed scenario: (i) \textbf{Uplink reward}, which stands for the uplink efficiency penalty, (ii) \textbf{Downlink reward}, which encapsulates the downlink efficiency and energy expenditure penalty, (iii) \textbf{Global reward}, which encompasses the total delay and re-transmission penalty. Aside from the rewards, the training efficiency of an algorithm is another important metric. It reflects on an algorithm's ability to learn optimal policies quickly. Therefore, we illustrate the (iv) \textbf{Training time} of different methods.

In our proposed scenario, we consider the (v) \textbf{Total latency} taken to complete the digital twinning task as the most important objective. Hence, it makes complete sense to include the total time taken including the latency caused by re-transmissions. (vi) \textbf{Re-transmission percentage} (re-transmission times divided by the number of total transmission iterations) reflects the reliability of the system and directly impacts the total time taken for UL and DL transmissions. To satisfy the ``Green Metaverse'' requirement, we are concerned with (vii) the \textbf{Energy consumption}. Additionally, we use (viii) the \textbf{Maximum uplink rate} of all XUs in one iteration to testify if the UL agent can handle the channel access task and be influenced by the DL transmission.

Note that we log the number of transmission iterations (UL and DL) during training instead of the total time (in milliseconds) as it is more clear and more intuitive.

\subsection{Experimental settings}
\label{section:num-set}
For our experiments, the bandwidth $W_m$ for every channel is simulated to be $10$ Ghz~\cite{6Gmagzine}, and all the Gaussian noise power spectral densities are simulated to be $-174$ dBm/Hz. To simplify the simulation, we set the data augmentation function $f_n(\cdot)$ as a proportional function with slope $c_n^t$ (i.e., ${D'}_n^t=c_n^t\times D_n^t$), where $c_n^t$ is sampled from a uniform distribution [5,15] in this paper. The domain of $c_n^t$ is referenced from the experimental results from demos of NeuralRecon~\cite{realtime3D1} and Monster Mash~\cite{monstermash}, which are two impressive 3D reconstruction techniques that are abreast of the times. The initial buffer sizes $B_n^0$ and uplink transmission power of each XU are uniformly selected from $[10,20]$ Megabits (Mb) and $[3,10]$ Watts, respectively. The XUs and MC are uniformly located in a $100\times100~m^2$ indoor space. To ensure the adaptability of our method, these variables vary in every episode, and the solutions in each episode are not the same. The DTTI limit and UTTI in each iteration are simulated to $1.5$ms and $0.5$ms~\cite{changyangURLLC}, respectively. And the minimum and maximum downlink transmission powers ${p'}_{\max}$ are $0$ and $20$ Watt, respectively. The max training steps (iterations) in one episode are $100$, and the total training steps are $2$ million steps\footnote{As PPO is an on-policy algorithm that only uses the latest trajectories sampled from the current stochastic policy, it will fill the trajectory buffer first and use it for training, then empty the buffer and refill. Therefore, the training steps here are not the actual training times, but the sample times. And the training times should be equal to training steps divided by buffer capacity (trajectory length).}. As the random values influence the experiment outcome to a certain degree, we conducted each experiment using \textbf{global random seeds from 0 to 10} to ascertain a consistent and reliable study. We then draw the error bands to better quantify the performance of each algorithm. The detailed implementation and hyper-parameters are explained in Appendix~\ref{appendix:implementation}.


In terms of channel gain, Ibrahim \textit{{et al.}}~\cite{ricianRIS} studied the model of 6G indoor reconfigurable intelligent surface (RIS). Similar to them, we use the Rician fading as the small-scale fading. It is given by:
\begin{align}
    &h_{i,m}^t = \sqrt{\beta_n^t}g_{i,m}^t,
\end{align}
where
\begin{align}
    &\beta_n^t = \beta_0 (L_n^t)^{-\alpha}, \\
    &g_{i,m}^t = \sqrt{\frac{K}{K+1}}\bar{g} + \sqrt{\frac{1}{K+1}}\tilde{g}, \\
    &L_n =\sqrt{(X_n-X_{MC})^2 + (Y_n-Y_{MC})^2 + H^2}.
\end{align}
$(X_n,Y_n)$ is the location of XU $n$, and H is the height. $L_n^t$ represents the distance between XU $n$ and the MC, and $\beta_n^t$ represents the large-scale channel gain of XU $n$ at iteration $t$. $\bar{g}$ and $\tilde{g}$ stand for the Line-Of-Sight (LOS) component and the Non-LOS (NLOS) component, respectively. Here, $\tilde{g}$ follows the standard complex normal distribution $\mathcal{C}\mathcal{N}(0,1)$ distribution. $\beta_0$ denotes the channel gain at the reference distance $L_0 = 1$m, and $\alpha$ denotes the path-loss exponent which is simulated as $2$. The Rician factor $K$ is simulated as $3$. 

Here, \textbf{we use the notation $m-n$ to denote the scenario where there are $m$ channels and $n$ extended reality users (XUs)}. In conducting our experiments, we fix the number of channels at 3 and vary the number of XUs across 4 to 8, for comparison. 


\begin{figure*}[h]
\centering
\subfigtopskip=2pt
\subfigbottomskip=2pt

\subfigure[3-8 global reward.]{
\begin{minipage}[t]{0.24\linewidth}
\centering
\includegraphics[width=1\linewidth]{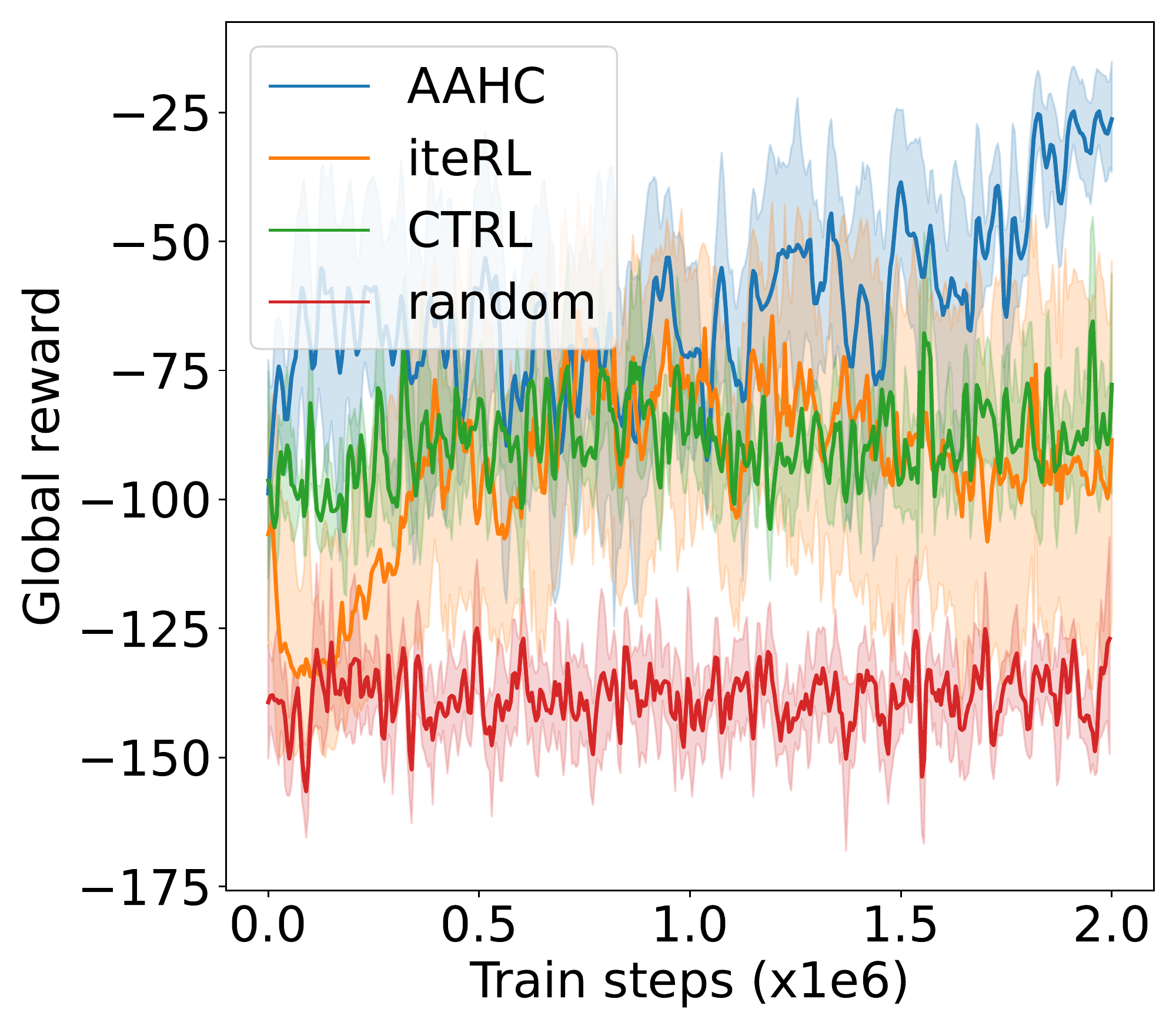}
\label{fig:38Greward}
\vspace{-10mm}
\end{minipage}
}%
\subfigure[3-8 uplink reward.]{
\begin{minipage}[t]{0.24\linewidth}
\centering
\includegraphics[width=1\linewidth]{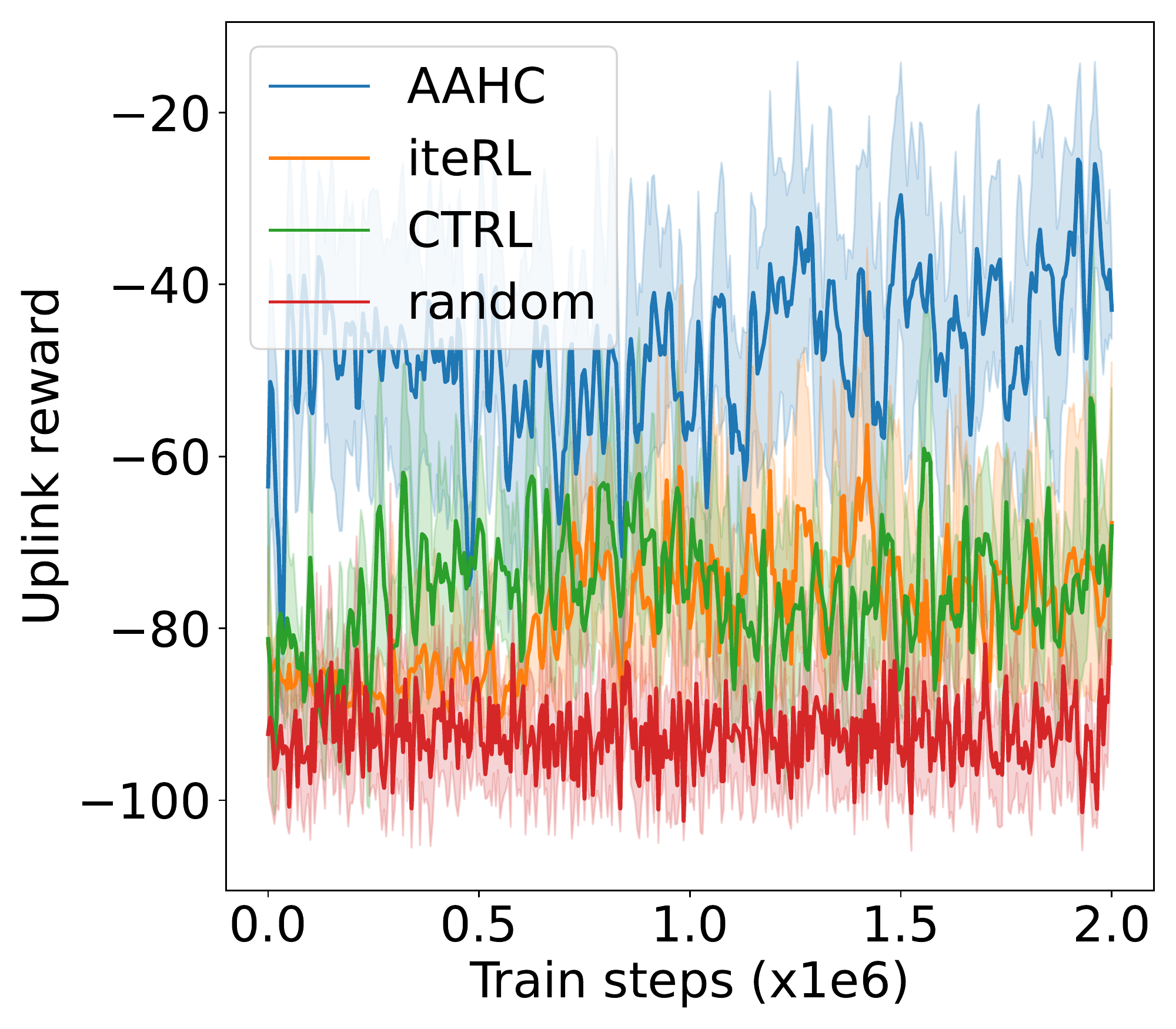}
\label{fig:38Ureward}
\vspace{-10mm}
\end{minipage}%
}%
\subfigure[3-8 downlink reward.]{
\begin{minipage}[t]{0.24\linewidth}
\centering
\includegraphics[width=1\linewidth]{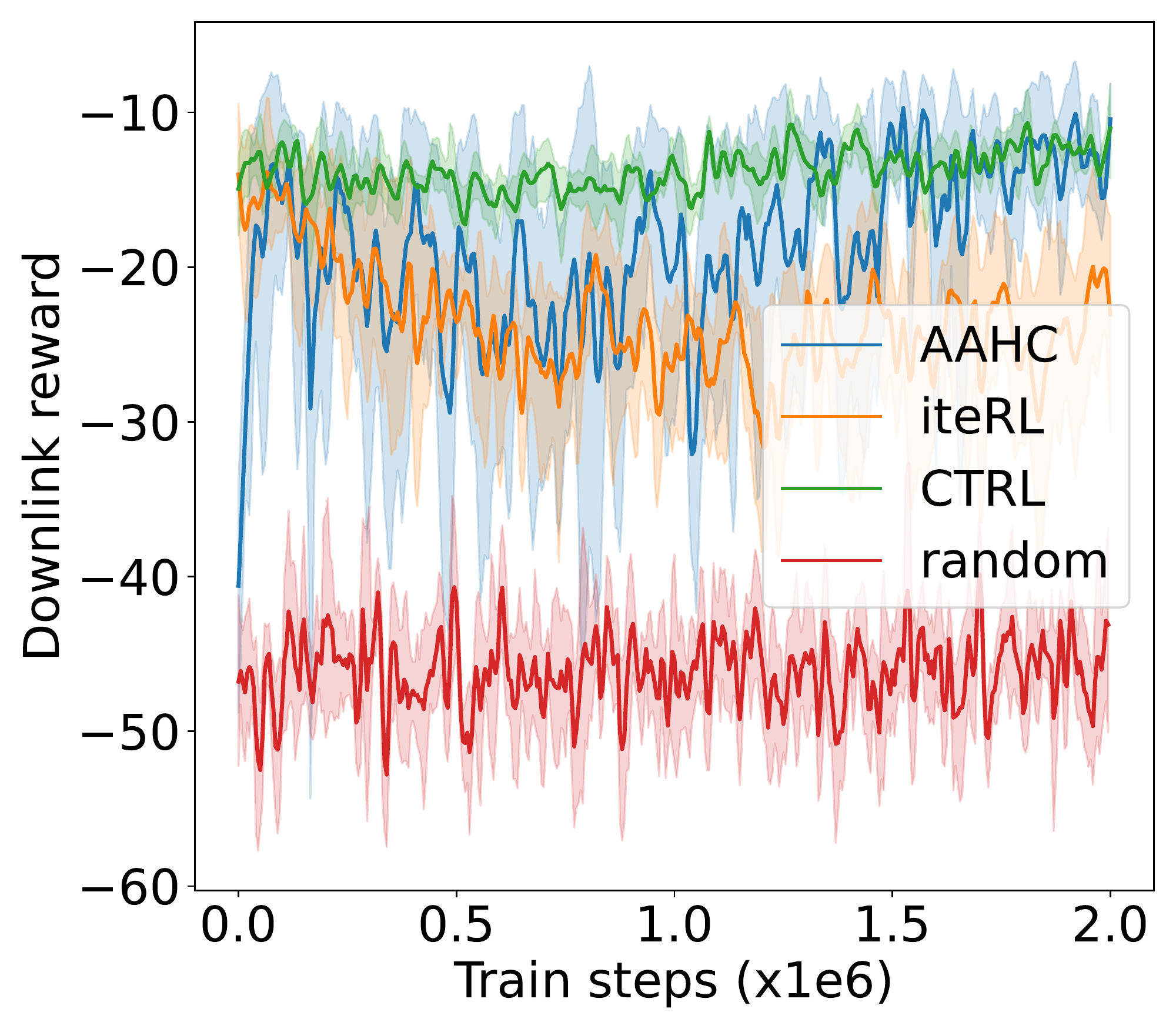}
\label{fig:38Dreward}
\vspace{-10mm}
\end{minipage}%
}%
\subfigure[3-8 number of iterations.]{
\begin{minipage}[t]{0.24\linewidth}
\centering
\includegraphics[width=1\linewidth]{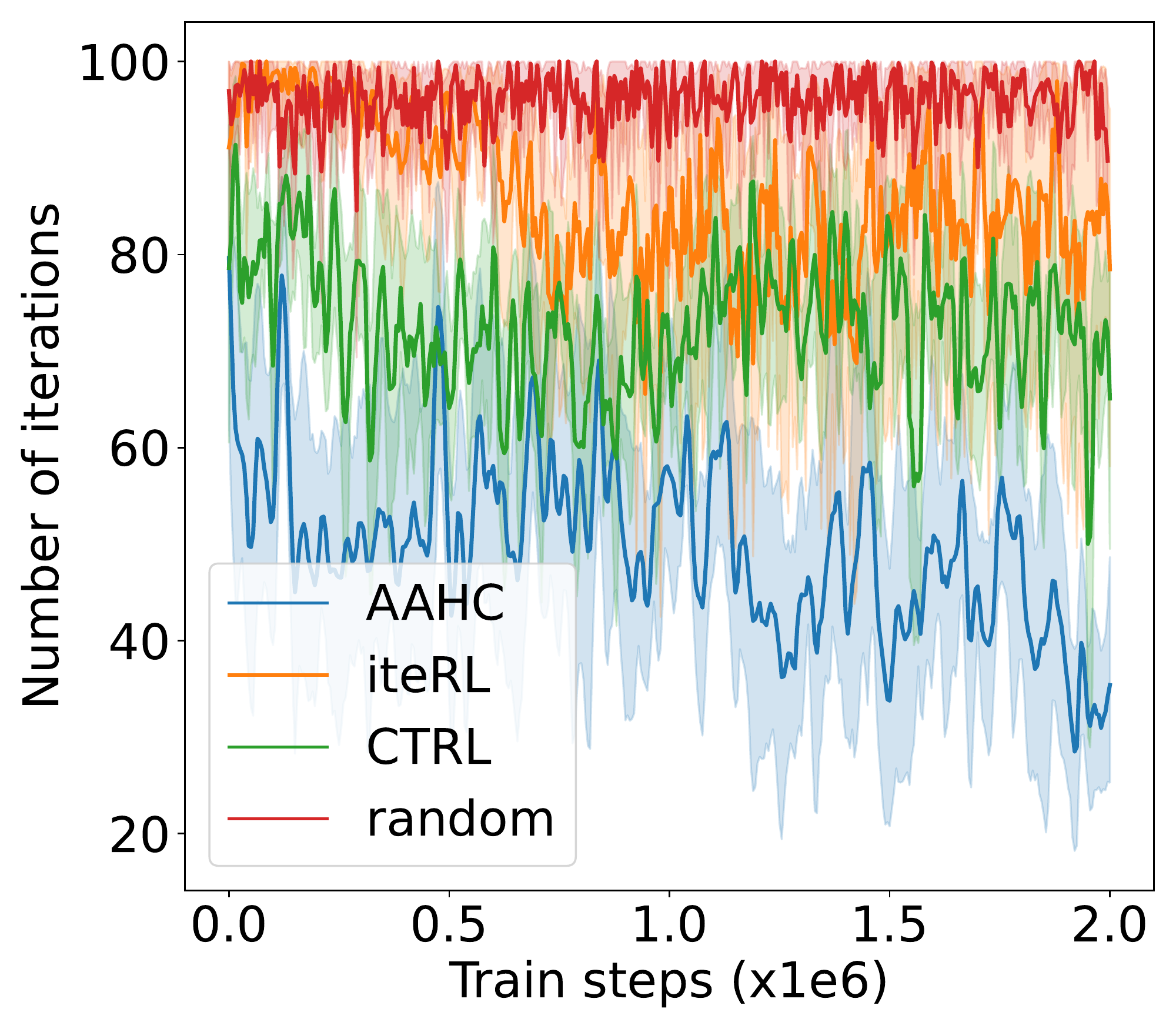}
\label{fig:38steps}
\vspace{-10mm}
\end{minipage}
}%

\subfigure[3-8 re-transmission percentage.]{
\begin{minipage}[t]{0.24\linewidth}
\centering
\includegraphics[width=1\linewidth]{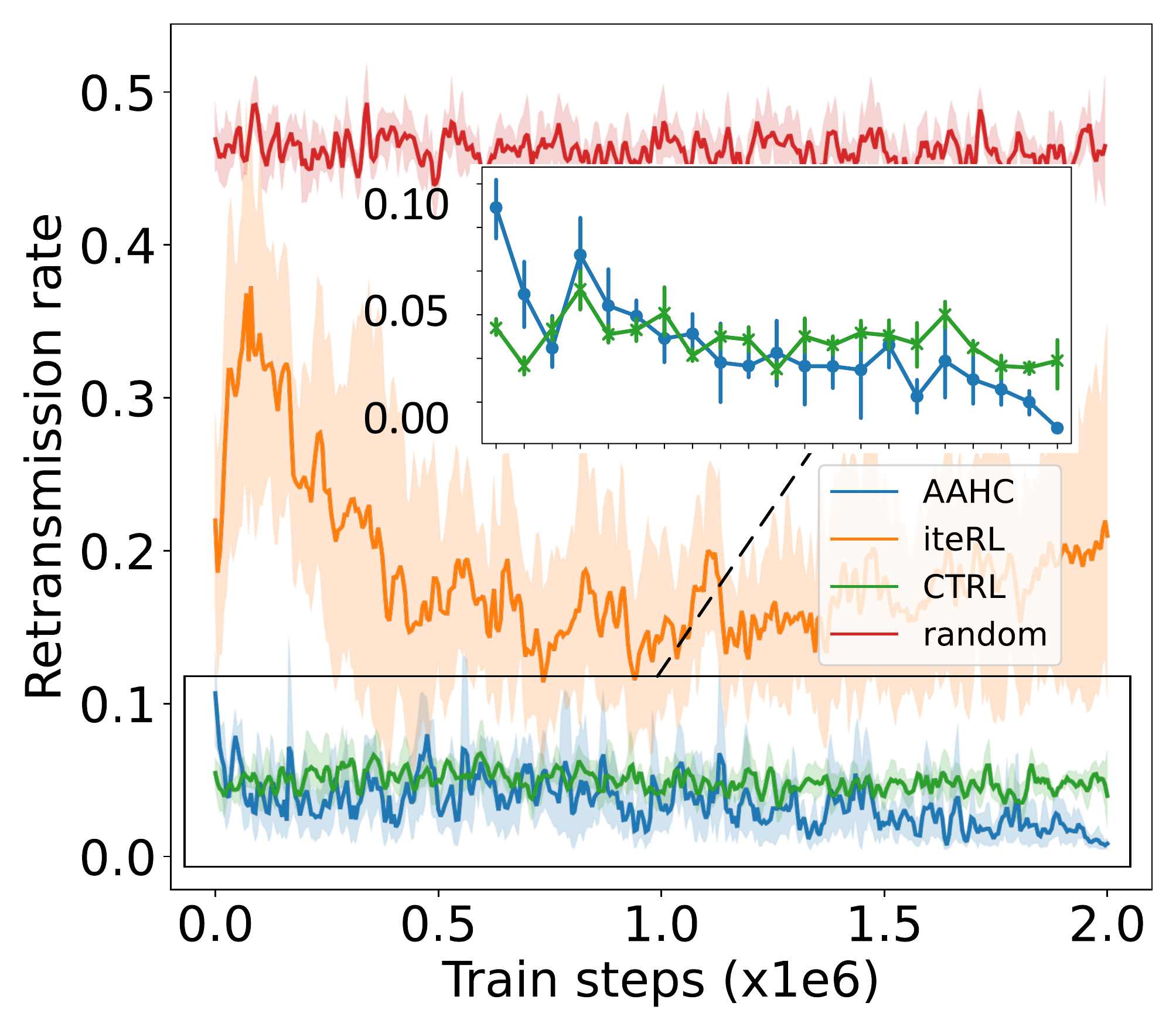}
\label{fig:38retrans}
\vspace{-10mm}
\end{minipage}
}%
\subfigure[3-8 max uplink rate per step.]{
\begin{minipage}[t]{0.24\linewidth}
\centering
\includegraphics[width=1\linewidth]{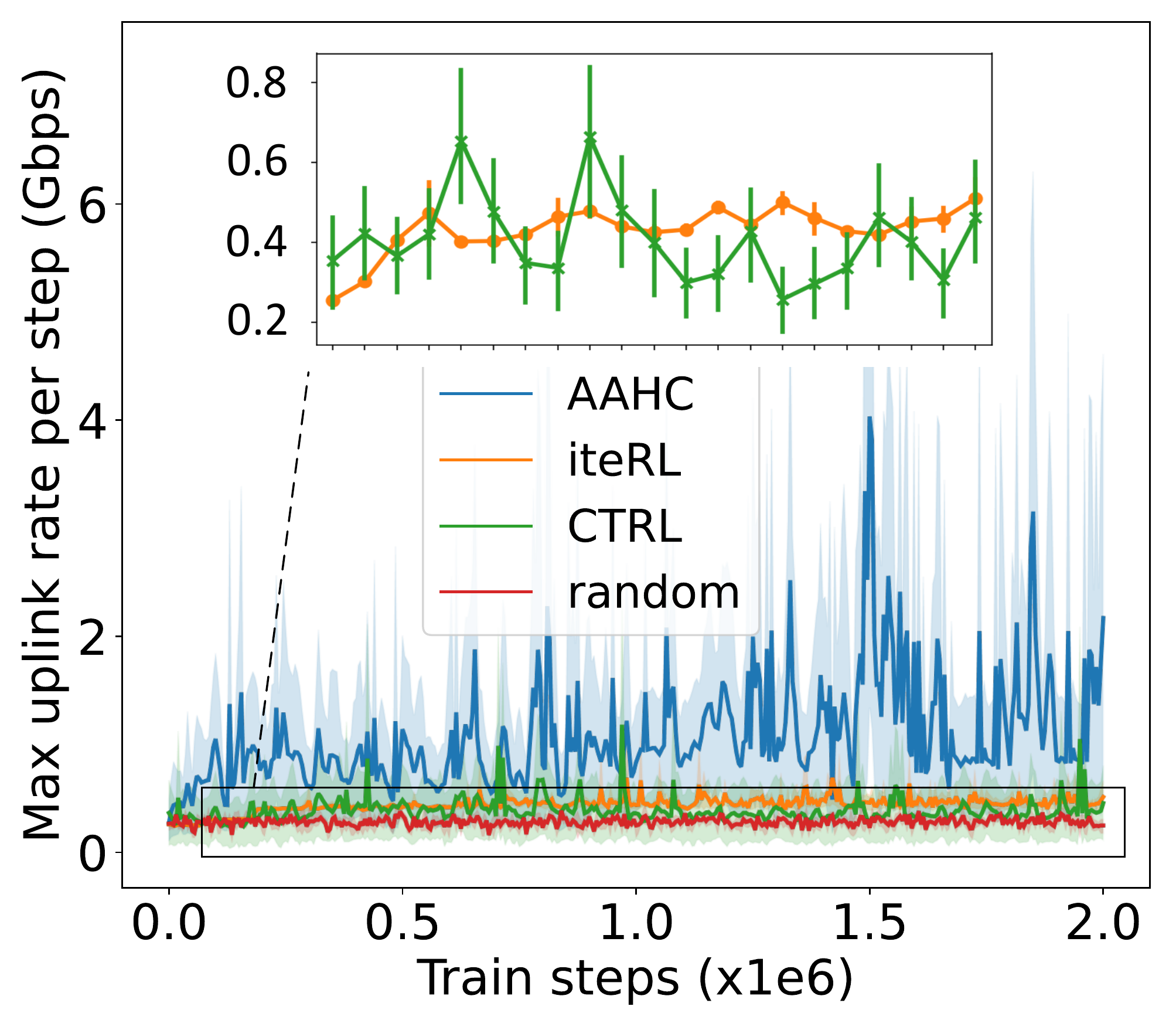}
\label{fig:38uprate}
\vspace{-10mm}
\end{minipage}%
}%
\subfigure[3-8 energy consumption.]{
\begin{minipage}[t]{0.24\linewidth}
\centering
\includegraphics[width=1\linewidth]{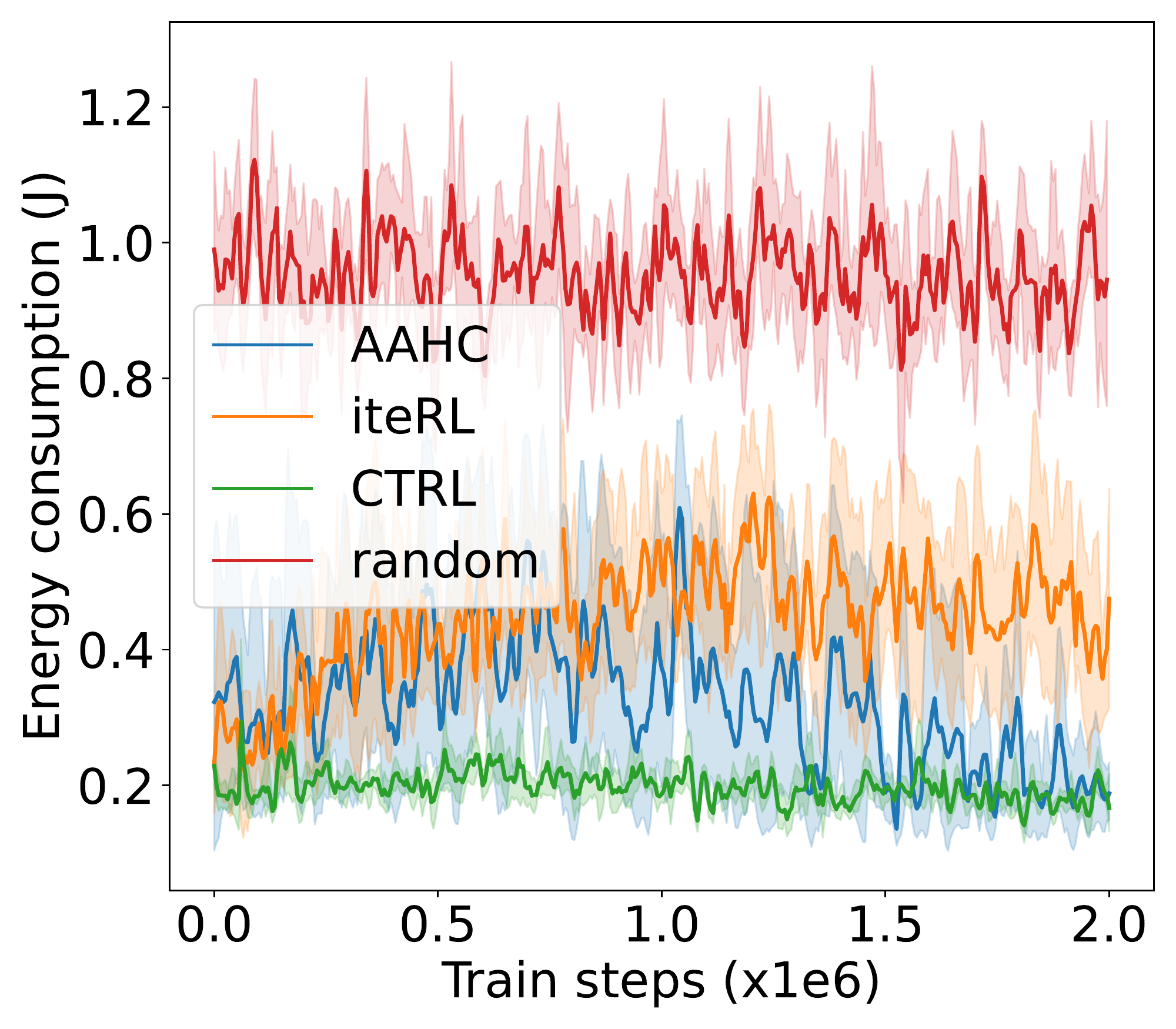}
\label{fig:38energy}
\vspace{-10mm}
\end{minipage}%
}%
\subfigure[Train time \& Execute time.]{
\begin{minipage}[t]{0.24\linewidth}
\centering
\includegraphics[width=1\linewidth]{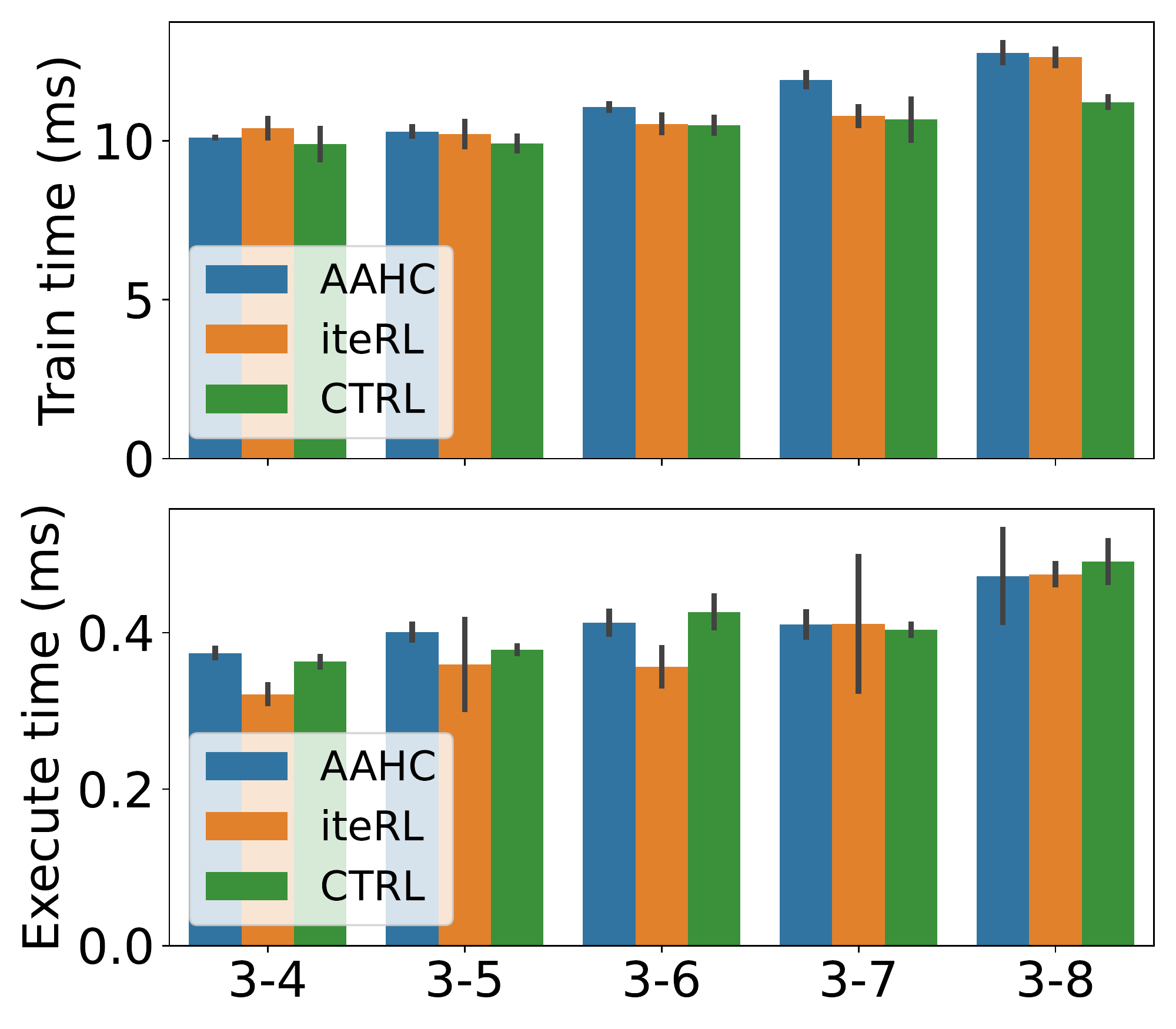}
\label{fig:traintime}
\vspace{-10mm}
\end{minipage}%
}%

\caption{Training in 3-8 scenarios. Considering the randomly evolving environment, all experiments are conducted with global random seeds from 0-10, and the error bands are drawn.}
\vspace{-0.5cm}
\end{figure*}

\begin{figure}[t]
    \centering
    \includegraphics[width=0.9\linewidth]{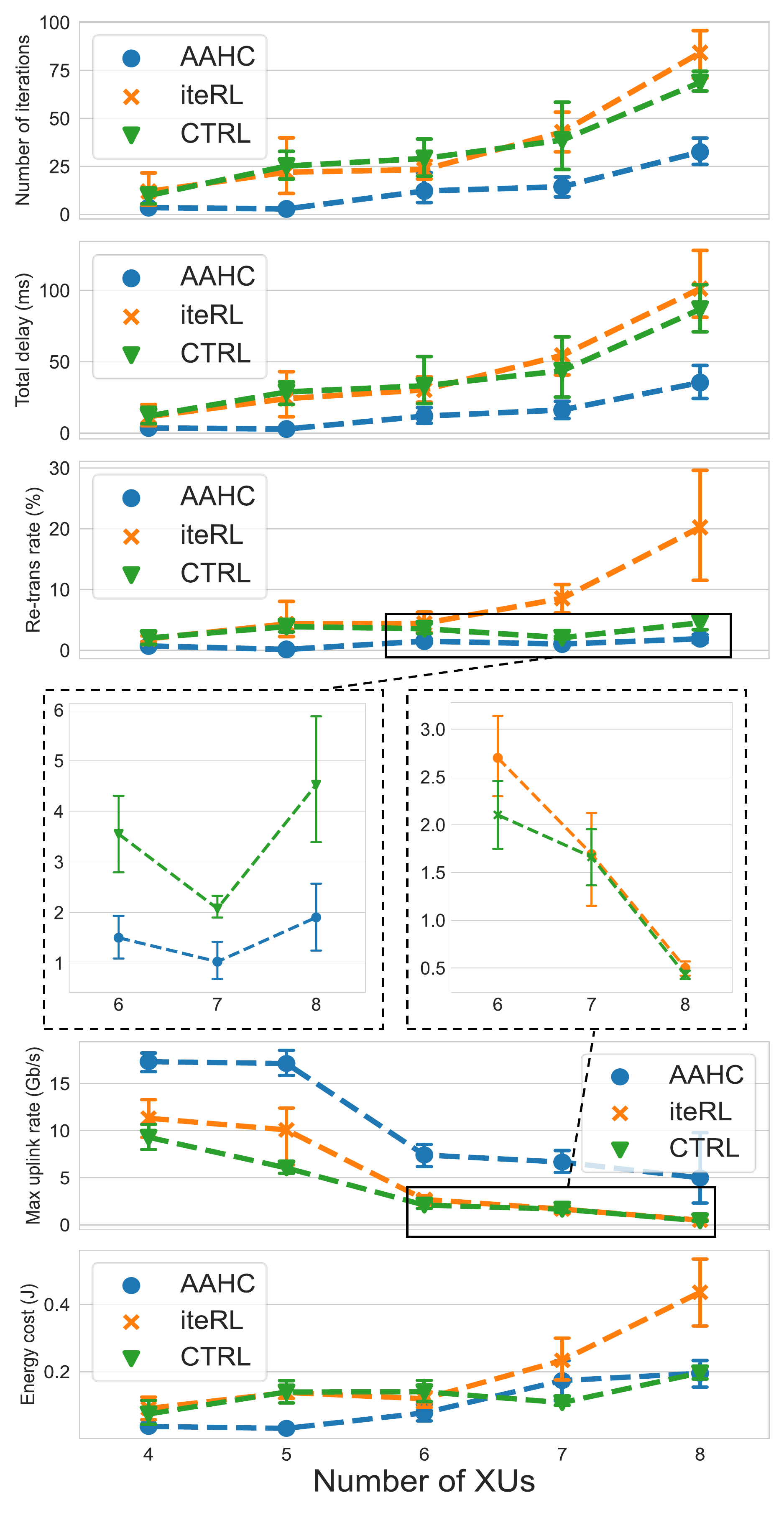}
    \caption{Metrics with different numbers of XUs.}
    \label{fig:metrics}
    \vspace{-0.5cm}
\end{figure}

\begin{figure}[t]
\centering
\subfigtopskip=2pt
\subfigbottomskip=2pt

\subfigure[3-8 heat map after 50000 training steps with AAHC.]{
\begin{minipage}[t]{0.9\linewidth}
\centering
\includegraphics[width=1\linewidth]{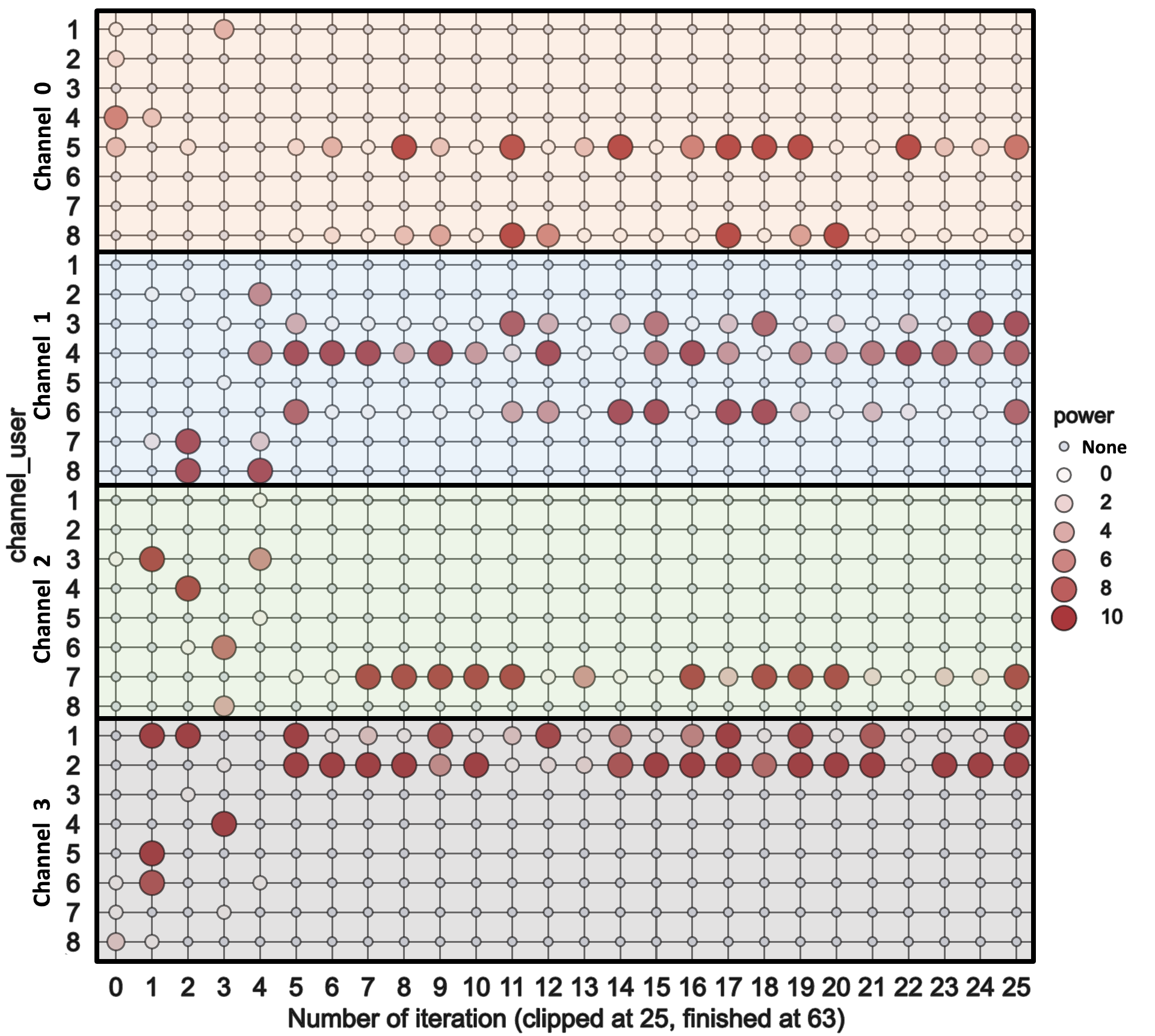}
\label{fig:heat50}
\vspace{-10mm}
\end{minipage}%
}%

\subfigure[3-8 heat map after 2 million training steps with AAHC.]{
\begin{minipage}[t]{0.9\linewidth}
\centering
\includegraphics[width=1\linewidth]{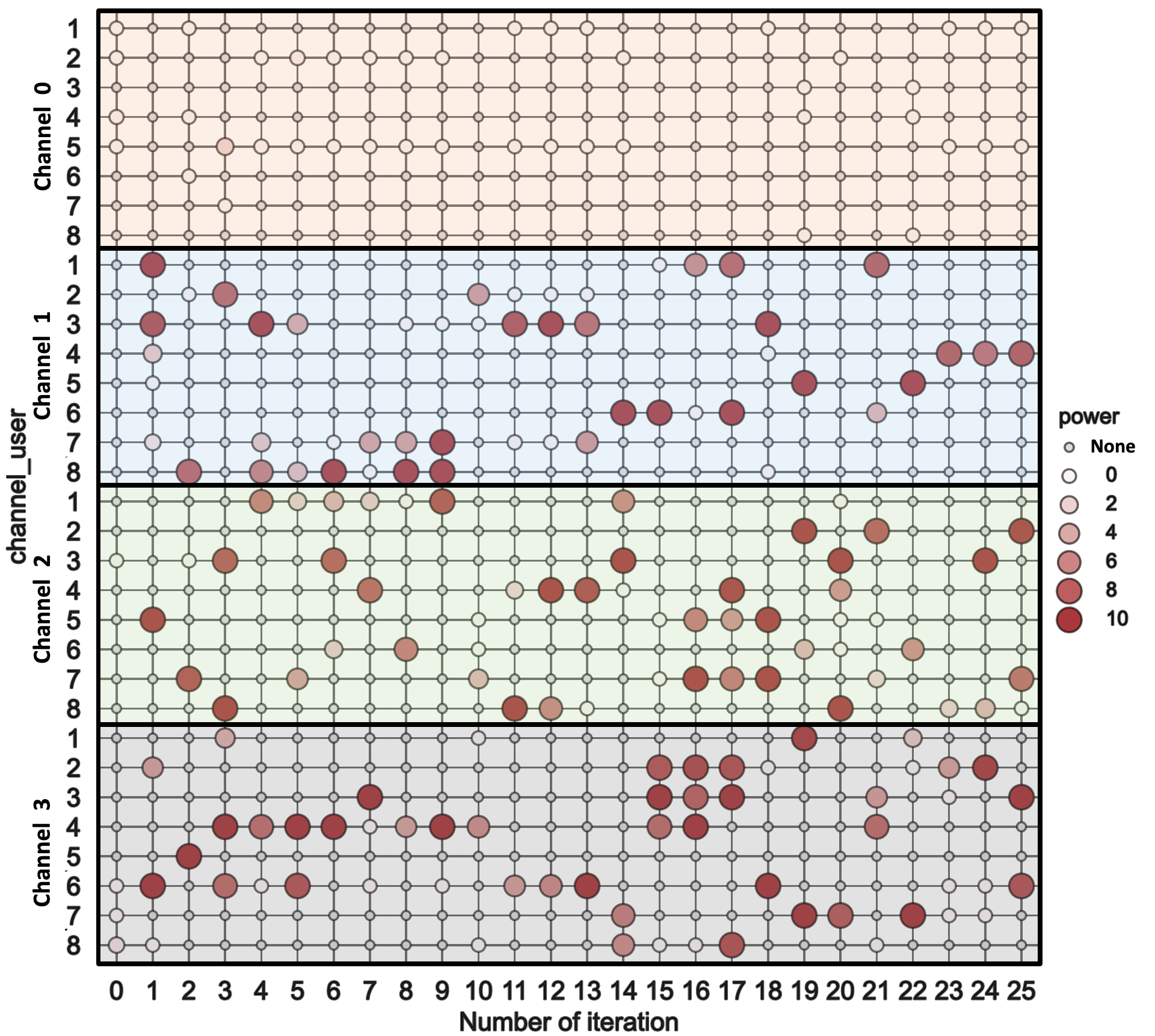}
\label{fig:heat200}
\vspace{-10mm}
\end{minipage}%
}%

\caption{3-8 heat maps with AAHC.}
\label{fig:heat}
\vspace{-0.5cm}
\end{figure}

\subsection{Results} \label{secResults}
In this part, we first choose to only display the experiment results of the most complex scenario: the $3-8$ configuration, to show the model convergence and characteristics. The complete set of experiments which include all other configurations are shown in Appendix~\ref{app:experiment}. Then, the metrics (KPIs) across different configurations are evaluated and discussed. The overall numerical results are shown in Table~\ref{table:results}.

\subsubsection{Train-time model performance in 3-8 configuration}
For ease of discussion, we compared the performance of our proposed method against other algorithms in the most complex setting: the $3-8$ configuration. Although the action space in this configuration is extremely huge, AAHC has demonstrated superior performance in handling the task when compared to the iteRL and CTRL baseline models.

In terms of rewards, AAHC has achieved the highest uplink, downlink, and global reward across the training episodes, when compared to the baselines. The global reward encompasses the total delay and re-transmission penalty. Despite having gradually increasing global rewards, the maximum uplink transmission rate of XUs is not increasing steadily across the training episodes. We observe from Fig.~\ref{fig:38Ureward} and Fig.~\ref{fig:38uprate} that the maximum uplink rate reaches a peak at about 1.5 million training steps and then decreases sharply. Then, it increases slowly from thereon, while the global reward increases almost steadily (Fig.~\ref{fig:38Greward}) in this stage. We can speculate that although the UL agent is able to achieve a higher max uplink rate, it may bring about higher transmission failure (re-transmission) counts, which can result in lower rewards. Therefore, AAHC learns to slowly decrease the max uplink rate after 1.5 million steps to avoid unacceptable re-transmissions.

The MC DL transmission energy consumption initially increases in the early training episodes, and subsequently decreases in the later training episodes. We believe that the agent attempts to maximize reward by minimizing transmission failure (and hence the overall total time taken), through increasing DL transmission power output and lowering re-transmission counts. However, this is at the expense of having a higher energy consumption which negatively impacts the global reward. Subsequently, it is likely that the agent learns to control the MC DL power output, using significantly lower power while maintaining re-transmission counts at a decently low rate. Overall, this improves the downlink reward and demonstrates the prowess of our proposed model in handling the complex scenario and reward structure.

Across the different performance metrics, iteRL performs the worst. The improvement in reward attainment of the UL agent across the training episodes is at the expense of the DL agent's performance. This poorer DL agent performance is reflected in its increasing DL transmission energy consumption, while the improving UL agent's performance is reflected in its overall decreasing re-transmission percentage, across the training. Although the global reward increases slightly across the training episodes, it achieves a much lower reward when compared to our proposed AAHC and CTRL algorithms. From the above observations, this signifies that the agents in the iteRL are non-cooperative and do not achieve an overall good channel arrangement and downlink power selection when compared to the AAHC and CTRL algorithms in this scenario. Nevertheless, the iteRL algorithm still achieves relatively good performance in less complex configurations such as $3-4$ in Fig.~\ref{fig:complete}. 

The CTRL algorithm fails to find an optimal solution in the complex $3-8$ configuration as the UL agent chose to keep the UL transmission rates at a low value so that the DL re-transmission percentage, and hence re-transmission percentage and energy consumption stays low. This sub-par performance is reflected in its considerably low eventual uplink reward and consistently high downlink reward. This problem likely results from the ability of the UL agent to perceive and consider the DL agent's reward and objective. In such a case, the UL agent is partially rewarded based on the DL agent's actions and it is unable to decipher how its action influences its own objectives. Hence, it could have fallen into a local optimal when it finds that decreasing the uplink rate can lead to a relatively higher overall reward because of the lower re-transmission counts and energy consumption.

\textbf{Complexity analysis.} We also analyze the computational complexity with the aid of real training and execution time illustrations. Let $(L_u, L_d, L_c)$, $(X_u^l, X_d^l, X_c^l)$ be the total number of layers, and the number of neurons in layer $l$'s of the three networks: UL Actor, DL Actor, and hybrid Critic. Let $d(s)$ for state $s$ be the input dimension, which is proportional to the state dimension described in Section~\ref{sec:statesetting}. Then we have $s_u^t$ and $s_d^t$ for UL Actor and DL Actor respectively in the $t$th training step. Here we first analyze the complexity in a single training step. In AAHC, one training step involves the updates of two Actors and the update of one Critic, and the hybrid Critic has three input-output branches. Therefore, the complexity of the $t$th training step can be derived as:
\begin{align}
    &O(B\{d(s_u^t)X_u^1 + d(s_d^t)X_d^1 + [d(s_u^t)+d(s_d^t)+d({s_u^t;s_d^t})]X_c^1\nonumber\\ 
    &+ \sum_{l=1}^{L_u-1}X_u^lX_u^{l+1} + \sum_{l=1}^{L_d-1}X_d^lX_d^{l+1} + 3\sum_{l=1}^{L_c-1}X_c^lX_c^{l+1}\}),\nonumber
\end{align}
where $B$ is the batch size decided in experiments. According to \cite{complexity}, the total computation complexity should be the complexity in one training step multiplied by the total steps used for convergence, which is hard to quantify. To better show the complexity, we illustrate the training times in one step and execution time (both agents in one iteration) of different algorithms under different scenarios in Fig.~\ref{fig:traintime}. We can observe from it that AAHC, iteRL, and CTRL all have similar training time and execution time in a single step. Although the hybrid Critic in AAHC needs to take in three different states and calculate three values, it needs only one back-propagation with the summed losses, while iteRL needs two back-propagation for two separate Critics. CTRL is slightly faster during training, but the difference is not obvious. As the execution stage is not related to the Critics, they are expected to have similar time in a single execution (evaluate) step. In practice, the training stage with huge computational complexity can be performed offline on the MC first. The training and execution are all conducted on a GTX 2080 Ti.

\begin{table}[t]
\centering
\caption{Overall results}
\label{table:parameter}
\vspace{-0.2mm}
\scalebox{0.95}{
 \footnotesize\begin{tabular}{cccccc}
\hline
\makecell{Scenario} & \makecell{Number of\\iterations} & \makecell{Total\\delay (ms)} & \makecell{Re-trans\\rate (\%)} & \makecell{Max uplink\\ rate (Gbps)} & \makecell{Energy \\cost (J)} \\ \hline
\multicolumn{6}{c}{AAHC} \\ \hline
$3$-$4$ & $3.5$ & $3.7$ & $0.52$ & $17.08$ & $0.07$\\
$3$-$5$ & $2.9$ & $3.2$ & $0.13$ & $18.33$ & $0.03$\\
$3$-$6$ & $12.1$ & $11.8$ & $1.43$ & $7.63$ & $0.08$\\
$3$-$7$ & $14.3$ & $16.1$ & $1.01$ & $6.95$ & $0.17$\\
$3$-$8$ & $31.1$ & $34.9$ & $1.86$ & $3.23$ & $0.19$\\ \hline

\multicolumn{6}{c}{iteRL} \\ \hline
$3$-$4$ & $11.9$ & $11.5$ & $1.91$ & $8.36$ & $0.09$\\
$3$-$5$ & $21.9$ & $23.9$ & $4.30$ & $12.46$ & $0.14$\\
$3$-$6$ & $23.2$ & $30.1$ & $4.44$ & $2.75$ & $0.12$\\
$3$-$7$ & $42.8$ & $54.3$ & $8.49$ & $1.94$ & $0.23$\\
$3$-$8$ & $83.9$ & $101.1$ & $20.26$ & $0.52$ & $0.44$\\ \hline

\multicolumn{6}{c}{CTRL} \\ \hline
$3$-$4$ & $9.9$ & $12.0$ & $2.11$ & $8.77$ & $0.04$\\
$3$-$5$ & $25.2$ & $29.0$ & $3.92$ & $5.73$ & $0.14$\\
$3$-$6$ & $29.2$ & $33.6$ & $3.55$ & $2.38$ & $0.15$\\
$3$-$7$ & $38.7$ & $43.9$ & $2.21$ & $1.88$ & $0.11$\\
$3$-$8$ & $69.2$ & $86.7$ & $4.54$ & $0.44$ & $0.20$\\ \hline

\multicolumn{6}{c}{random} \\ \hline
$3$-$4$ & $73.5$ & $114.9$ & $29.32$ & $3.71$ & $0.22$\\
$3$-$5$ & $81.4$ & $129.2$ & $32.94$ & $4.32$ & $0.28$\\
$3$-$6$ & $92.6$ & $176.0$ & $40.13$ & $1.96$ & $0.42$\\
$3$-$7$ & $94.7$ & $183.7$ & $44.23$ & $1.01$ & $0.57$\\
$3$-$8$ & $97.2$ & $180.1$ & $49.33$ & $0.32$ & $1.09$\\ \hline
\end{tabular}
}
\label{table:results}
\vspace{-0.5cm}
\end{table}

\subsubsection{Metric performance across different configurations}
As our model performance results are taken from a constantly improving RL model, we take the performance results of each RL model by evaluating the trained model on a new but identical environment and averaging the results. The results are shown in Fig~\ref{fig:metrics}.

In general, the total delay, re-transmission times, and energy consumption all increase with the rise in the number of users. Nevertheless, as compared to the other baseline RL models, AAHC displays greater capability in finding a near-optimal solution in more complicated scenarios (i.e., configuration $3-8$). This capability can be seen in the much slower increment in total delay and re-transmission counts for the AAHC, as the number of users increases, when compared to the iteRL and CTRL methods. 

Nevertheless, as the number of XU rises, the training complexity of the problem rises precipitously. When compared to AAHC, CTRL and iteRL methods fail to find satisfactory solutions in the $3-7$, $3-8$ scenarios as the total delays are 2 or 3 times that of AAHC. Furthermore, according to the error bars drawn in Fig.~\ref{fig:metrics}, AAHC also has the lowest variance under different random seeds. Therefore, we infer from the results that AAHC is more efficient and stable than CTRL and iteRL, especially in more complicated scenarios. AAHC learns a conservative policy of restraining the UL rate to avoid re-transmission, reducing the total time delay. This demonstrates that our proposed methods enable the agents to have a global view through the shared losses, instead of directly sharing the extra state or action information from each other. This is not unexpected because in such a hybrid-reward scenario, giving the agent extra information from the other agent is not always advisable. In this scenario, the energy consumption in each DL can just be influenced by the actions of the downlink agent, and it is not related to UL agent. Thus, the information of DL energy consumption is redundant to UL agent, and it will even impinge on the action evaluation of UL agent.

To better visualize the improvement during training, we sample trajectories from 1 evaluation episode from each of two pre-trained AAHC models with identical initial settings. The two AAHC models differ in that one of the pre-trained models has been pre-trained for 2 million iterations, while the other has undergone disrupted training in the early iterations. As shown in Fig.~\ref{fig:heat}, the y-axis \textit{channel-user} denotes the allocation of users to the channel, with a sizeable node referring that a particular user having been assigned to the specified channel. Nodes with larger sizes and deeper colors denote that the DL agent allocates more power for the DL transmission to the specified user in a particular channel. The x-axis indicates the number of training iterations.

In the initial stages of training (i.e., 50000 iterations), as illustrated in Fig.~\ref{fig:heat50}, the policy is poorly trained and is unable to select good actions under states. It is unsurprising, as agents lack exploration in the early stages of training. Coupled with being faced with a complex scenario, the agent struggles to make optimal decisions under various states. As a consequence, the actions chosen by the agent lacked pliability and variability when faced with different states. Evidently, the lack of variability in chosen actions by the agent is inconsistent with the fact that the environment is constantly evolving, in which channel gains and XUs' locations are changing in each time step. Furthermore, the 50,000 iterations pre-trained model exhibit in-efficiency by continuously allocating power to XUs that are arranged with no channel (channel 0). The 2 million iterations pre-trained model performed much better, as shown in Fig.~\ref{fig:heat200}. The user-channel arrangement is much more dispersive, and the decision-makings by the two agents exhibit much more variability in response to granular changes in the state. Compared to the 50,000 iterations pre-trained model, which uses 63 iterations to finish the whole task, the 2 million iterations pre-trained model needs only 25 iterations.


\section{Conclusion}
\label{conclusion}
In this work, we have investigated an asynchronous hybrid-reward joint optimization problem, the real-time 3D reconstruction in xURLLC over wireless communication with multiple XR users, where the uplink and downlink are considered in tandem. We formulate the problem as an asynchronous multi-agent reinforcement learning task and propose the novel AAHC algorithm. Multiple KPIs such as the total delay, energy consumption, and re-transmission percentage are studied to fulfill the reliability and low latency of our communication system. And extensive experiments demonstrate that AAHC has more granular views on each agent and performs better in asynchronous and hybrid tasks with a preferable training time. We hope this work can provide more insights into the asynchronous cooperative tasks, as they are common in communication problems, and important to guarantee the reliability of the whole system.


%


\ifCLASSOPTIONcaptionsoff
  \newpage
\fi

\renewcommand{\baselinestretch}{.96}
 \renewcommand{\refname}{~\\[-20pt]References\vspace{-5pt} }

{\small
\bibliographystyle{IEEEtran}

}
\renewcommand{\baselinestretch}{1}

\appendices

\section{Implementation details}
\label{appendix:implementation}
For all these experiments, Adam is used as the optimization algorithm. The discount factor $\gamma$ and GAE factor $\lambda$ are fixed at $0.99$ and $0.95$. The batch size is set to $64$. The learning rates for uplink Actor, downlink Actor, and hybrid Critic are $10^{-4}, 10^{-4}, 5\times10^{-5}$, respectively. The entropy coefficient is set as $10^{-3}$.
As the essential objective of RL is to maximize the expected return (reward), the reward setting is always crucial. We highly recommend standardizing the values across kinds of rewards, or the parameter tuning will become a daunting task, and the large losses will make it slow to converge.

In terms of the activation functions, we recommend trying to use tanh instead of ReLU first, especially in the simpler network (in fact, networks in this kind of scenario where there is only deep neural network (DNN) but not convolutional neural network (CNN) are always much simpler than those in computer vision domains). Because tanh is zero-centered. Hence we can easily map the output values as strongly negative, neutral, or strongly positive. In other words, this reduces the difficulty of reward settings. However, tanh in hidden layers faces the problem of vanishing gradient. Thus, we should be careful when using this.

\section{Additional Experiments} \label{Appendix:Additional:experiments}

\begin{figure}[h]
\centering
\subfigtopskip=2pt
\subfigbottomskip=2pt

\renewcommand*{\thesubfigure}{}
\subfigure[3-4 re-transmission percentage.]{
\begin{minipage}[t]{0.48\linewidth}
\centering
\includegraphics[width=1\linewidth]{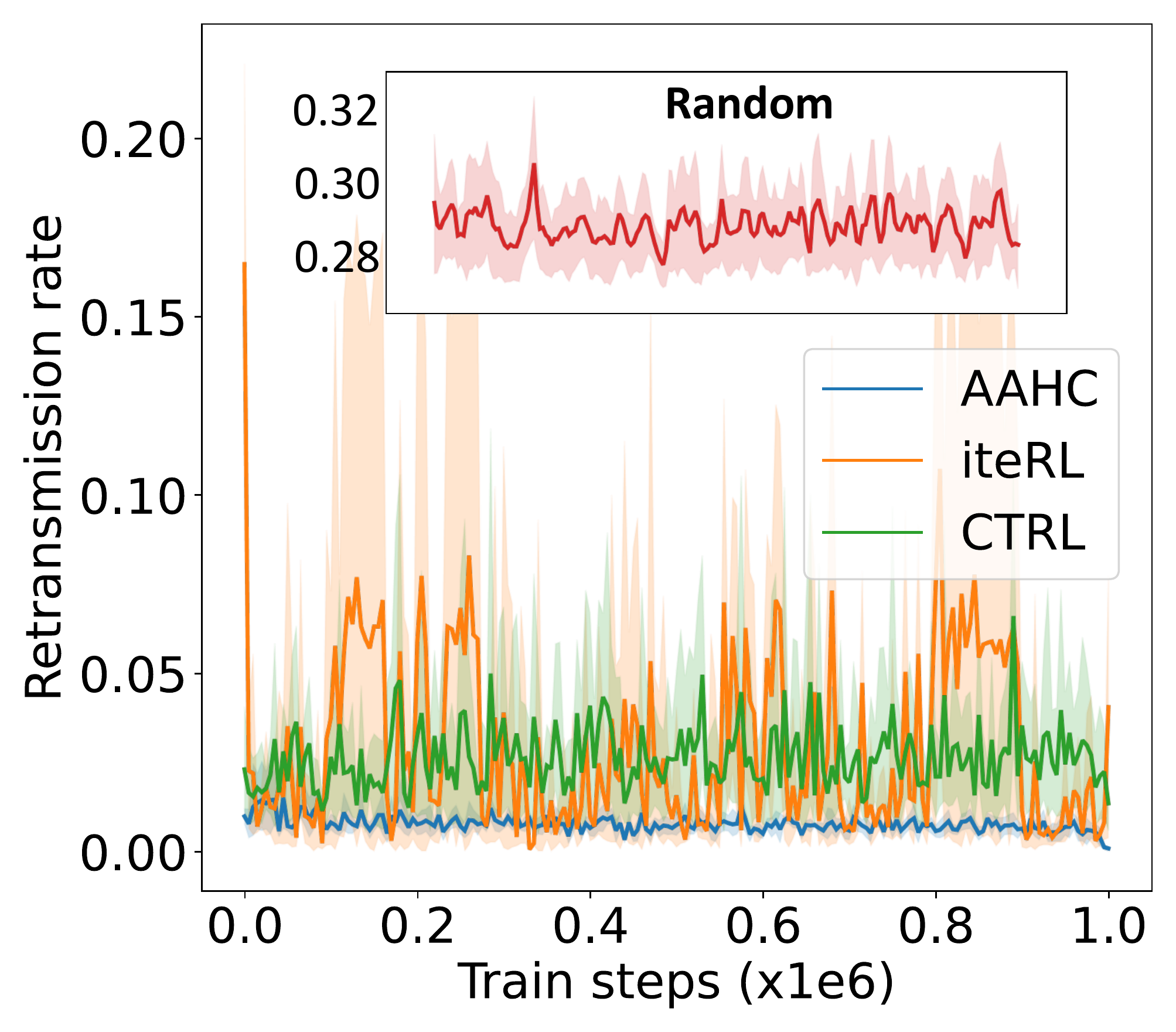}
\vspace{-10mm}
\end{minipage}%
}%
\subfigure[3-5 re-transmission percentage.]{
\begin{minipage}[t]{0.48\linewidth}
\centering
\includegraphics[width=1\linewidth]{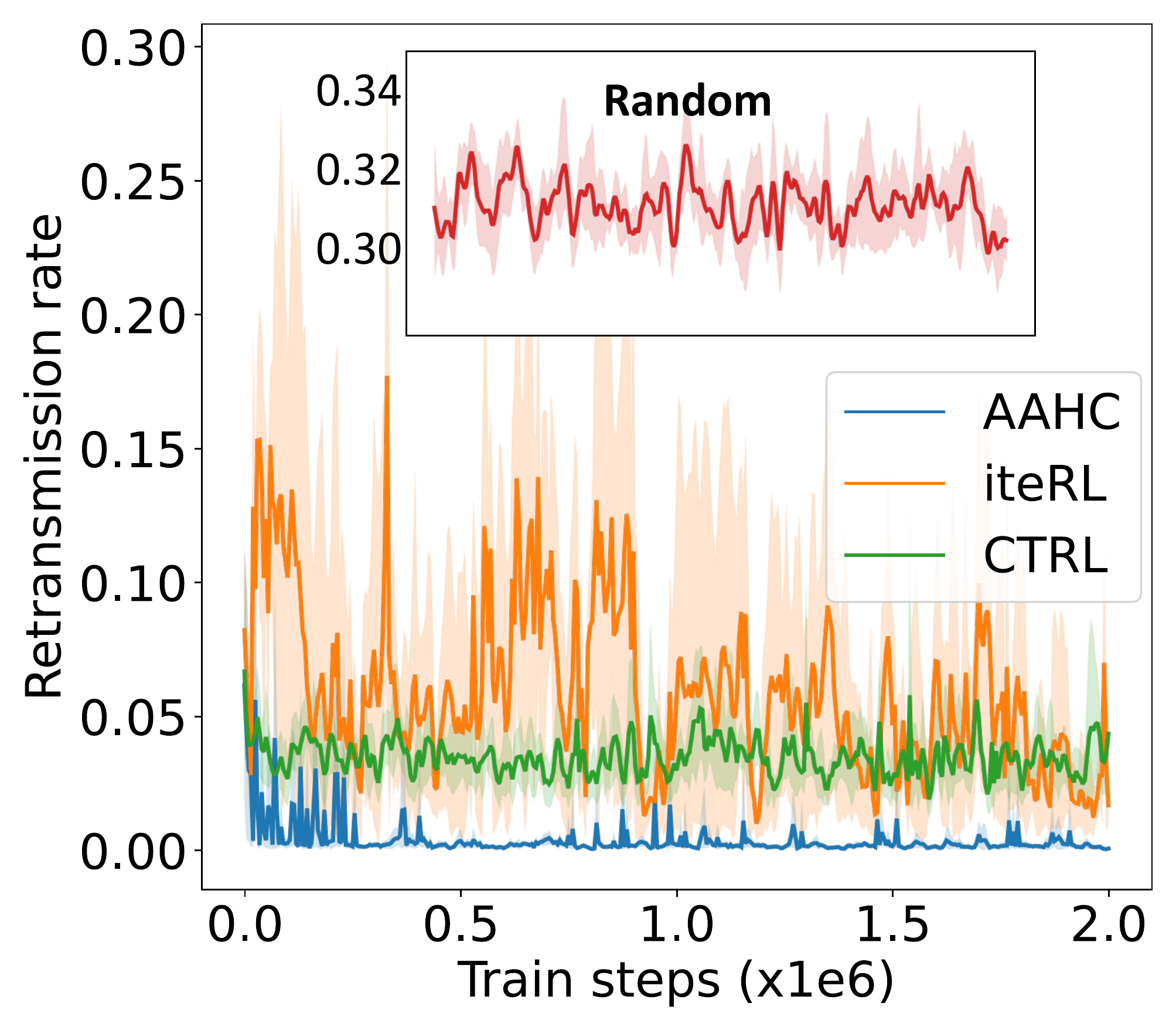}
\vspace{-10mm}
\end{minipage}%
}%

\subfigure[3-6 re-transmission percentage.]{
\begin{minipage}[t]{0.48\linewidth}
\centering
\includegraphics[width=1\linewidth]{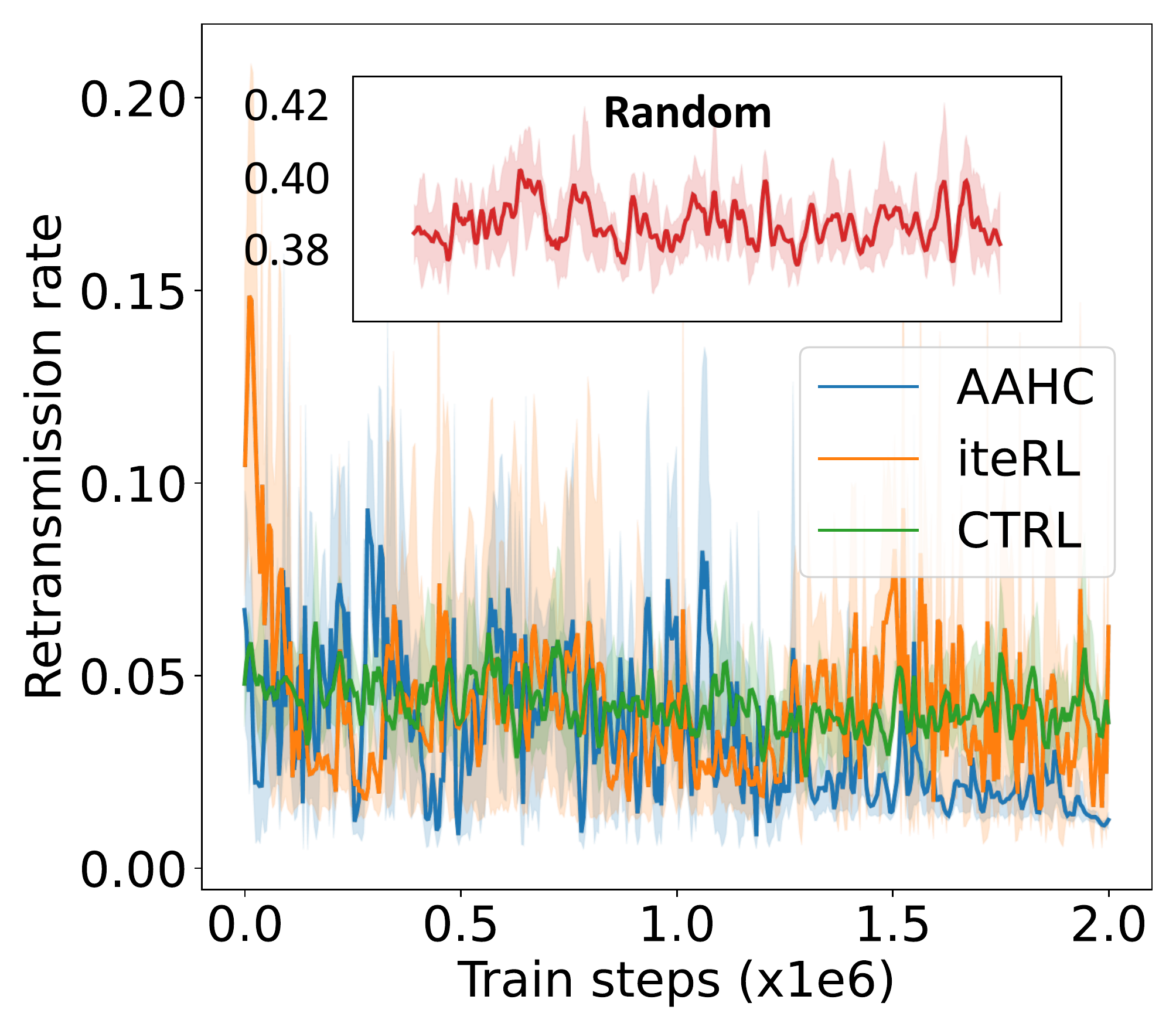}
\vspace{-10mm}
\end{minipage}
}%
\subfigure[3-7 re-transmission percentage.]{
\begin{minipage}[t]{0.48\linewidth}
\centering
\includegraphics[width=1\linewidth]{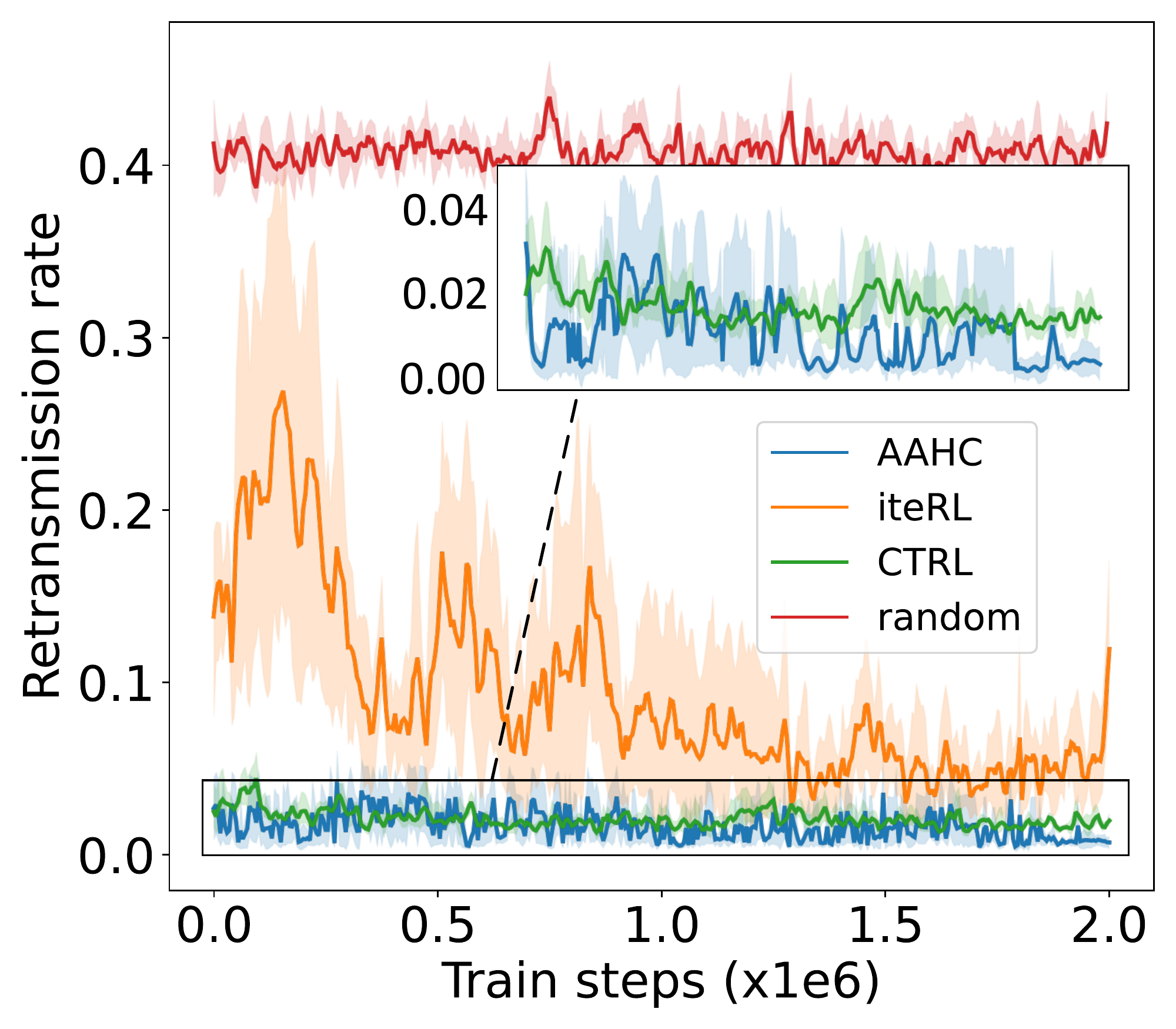}
\vspace{-10mm}
\end{minipage}
}%
\caption{Re-transmission percentage in 3-4 to 3-7 scenarios.} \label{fig7}
\vspace{-0.5cm}
\end{figure}

\begin{figure*}[t]

\centering
\subfigtopskip=2pt
\subfigbottomskip=2pt
\renewcommand*{\thesubfigure}{}

\subfigure[3-4 total iterations.]{
\begin{minipage}[t]{0.2\linewidth}
\centering
\includegraphics[width=1\linewidth]{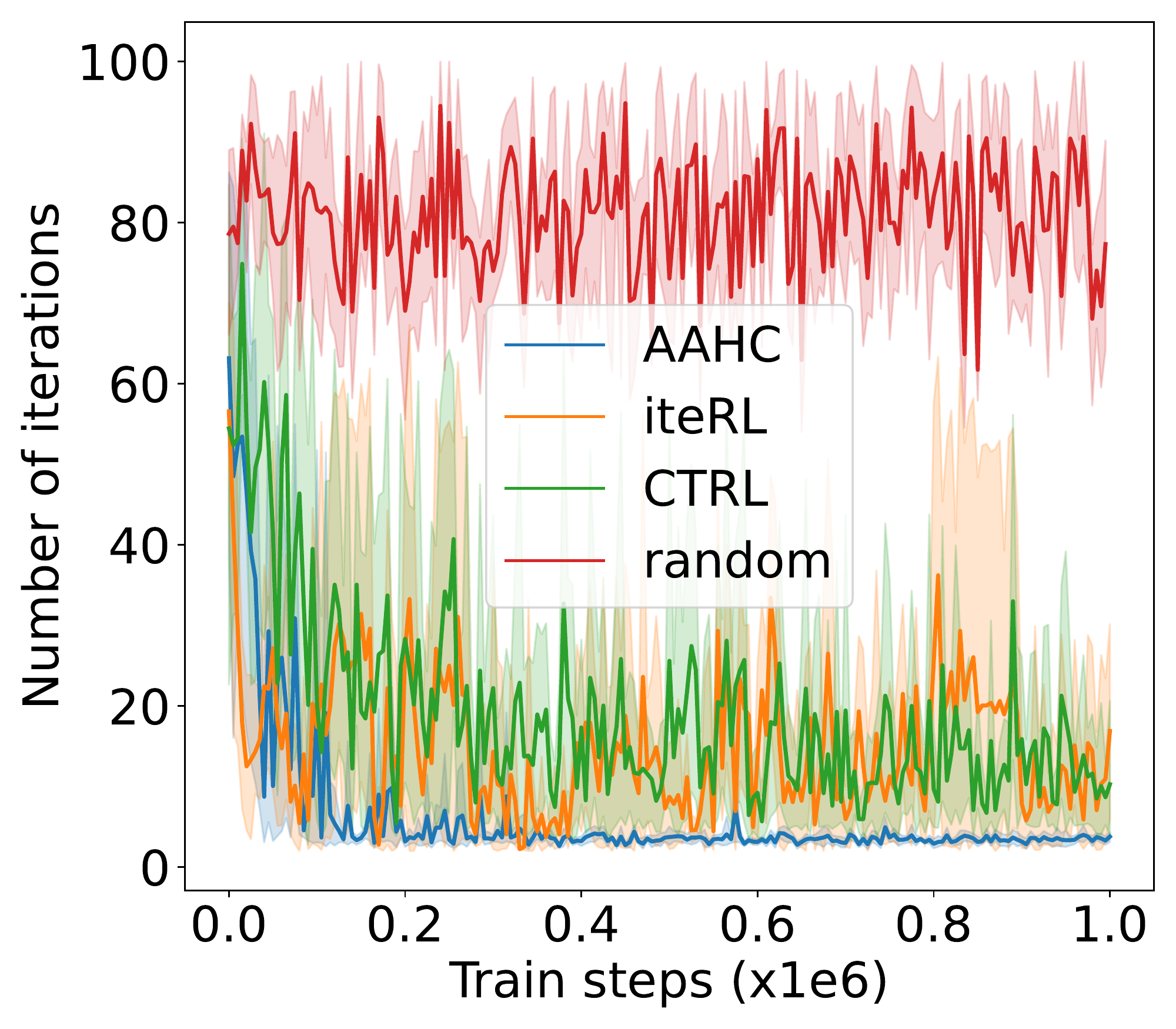}
\vspace{-10mm}
\end{minipage}%
}%
\subfigure[3-5 total iterations.]{
\begin{minipage}[t]{0.2\linewidth}
\centering
\includegraphics[width=1\linewidth]{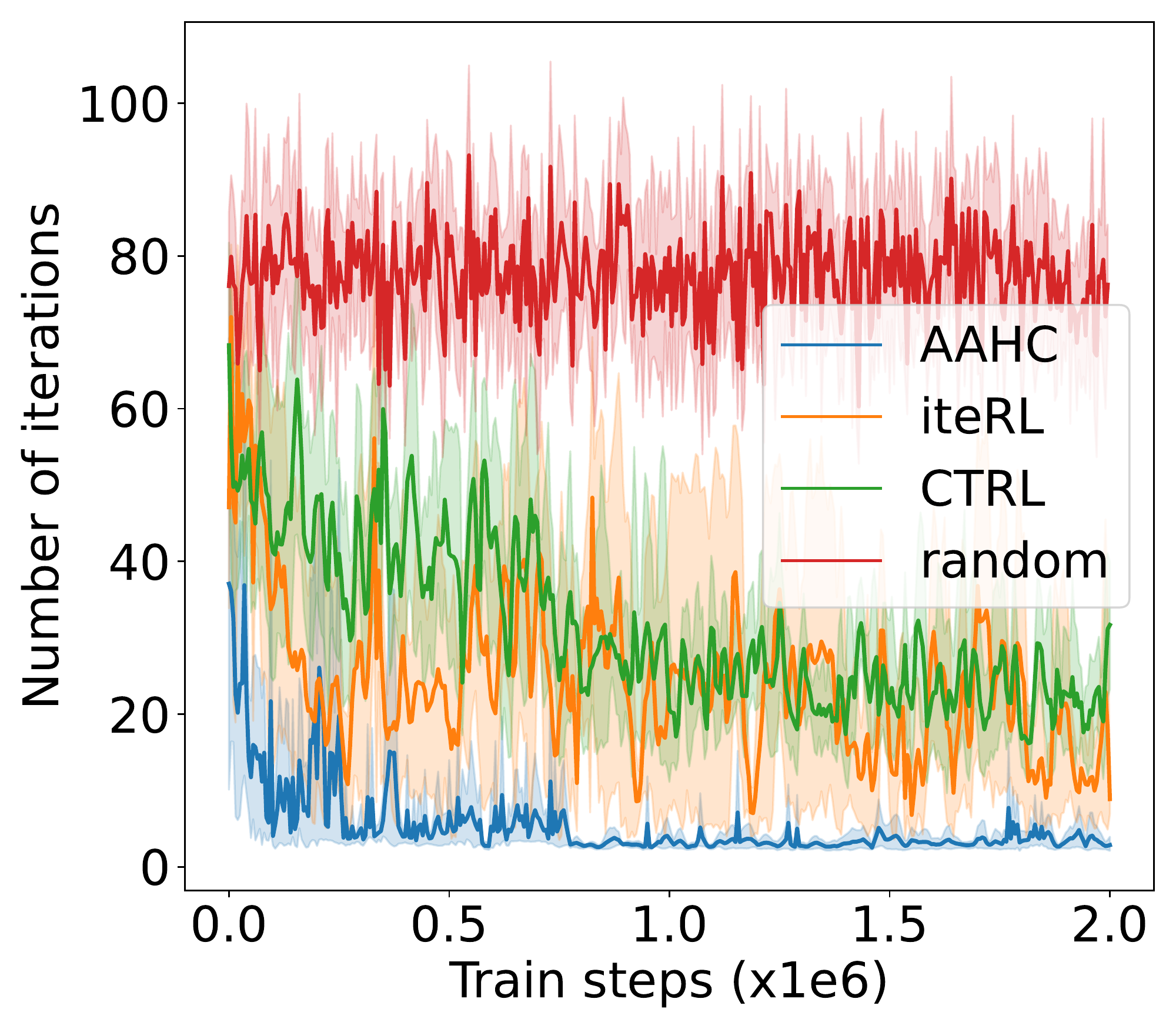}
\vspace{-10mm}
\end{minipage}%
}%
\subfigure[3-6 total iterations.]{
\begin{minipage}[t]{0.2\linewidth}
\centering
\includegraphics[width=1\linewidth]{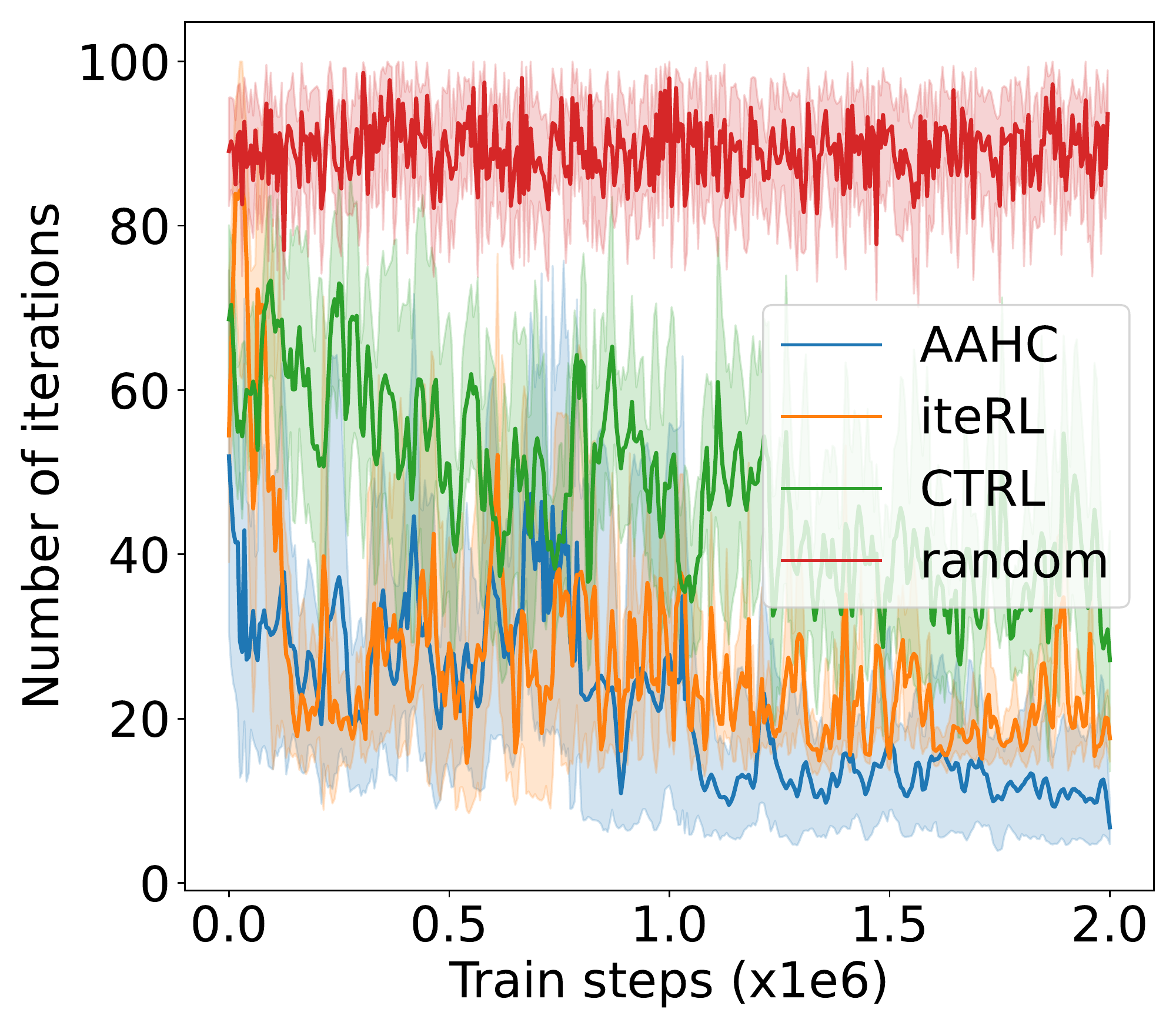}
\vspace{-10mm}
\end{minipage}
}%
\subfigure[3-7 total iterations.]{
\begin{minipage}[t]{0.2\linewidth}
\centering
\includegraphics[width=1\linewidth]{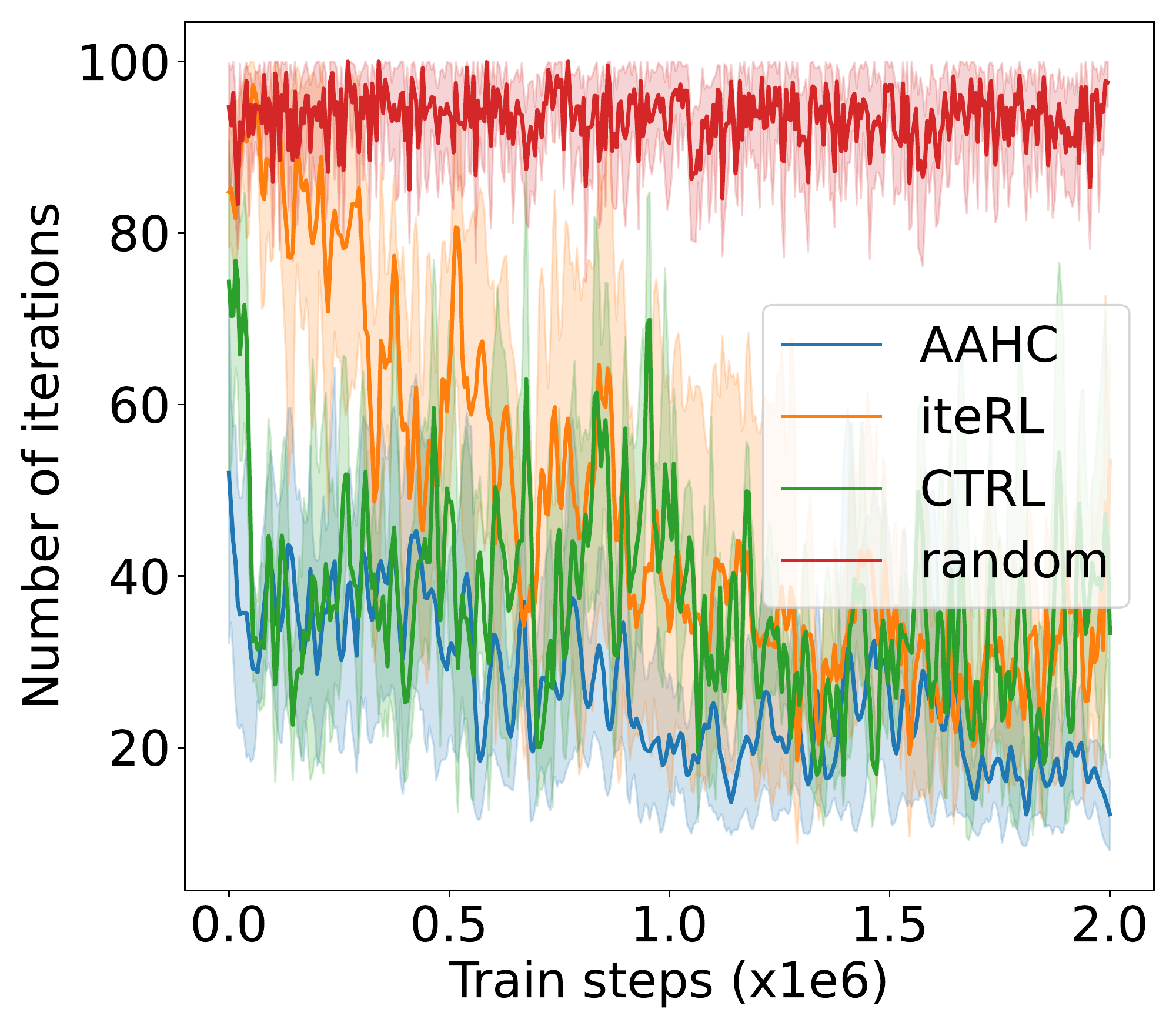}
\vspace{-10mm}
\end{minipage}
}%
\subfigure[3-8 total iterations.]{
\begin{minipage}[t]{0.2\linewidth}
\centering
\includegraphics[width=1\linewidth]{experiment/38steps.pdf}
\vspace{-10mm}
\end{minipage}
}%

\subfigure[3-4 global reward.]{
\begin{minipage}[t]{0.2\linewidth}
\centering
\includegraphics[width=1\linewidth]{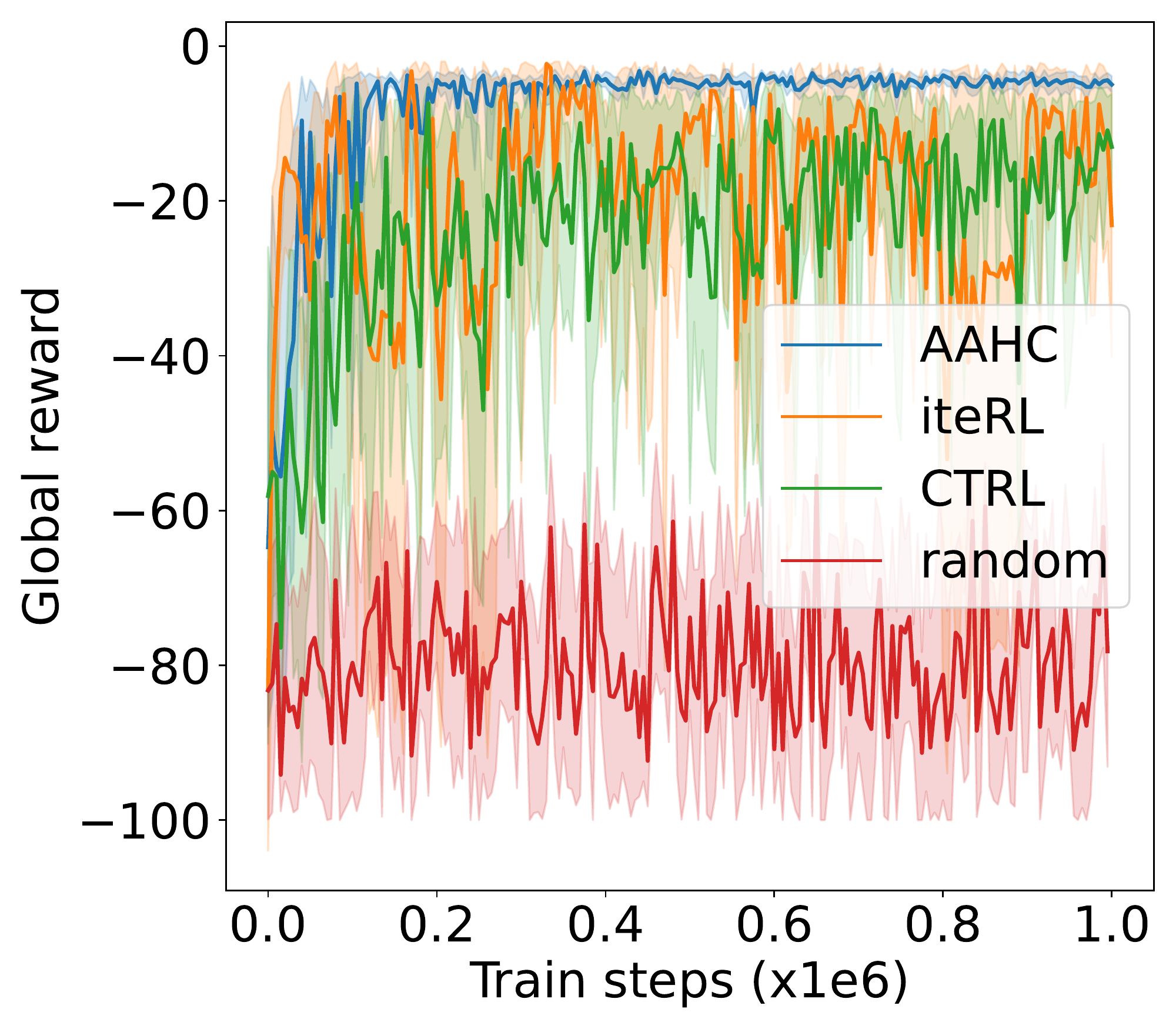}
\vspace{-10mm}
\end{minipage}%
}%
\subfigure[3-5 global reward.]{
\begin{minipage}[t]{0.2\linewidth}
\centering
\includegraphics[width=1\linewidth]{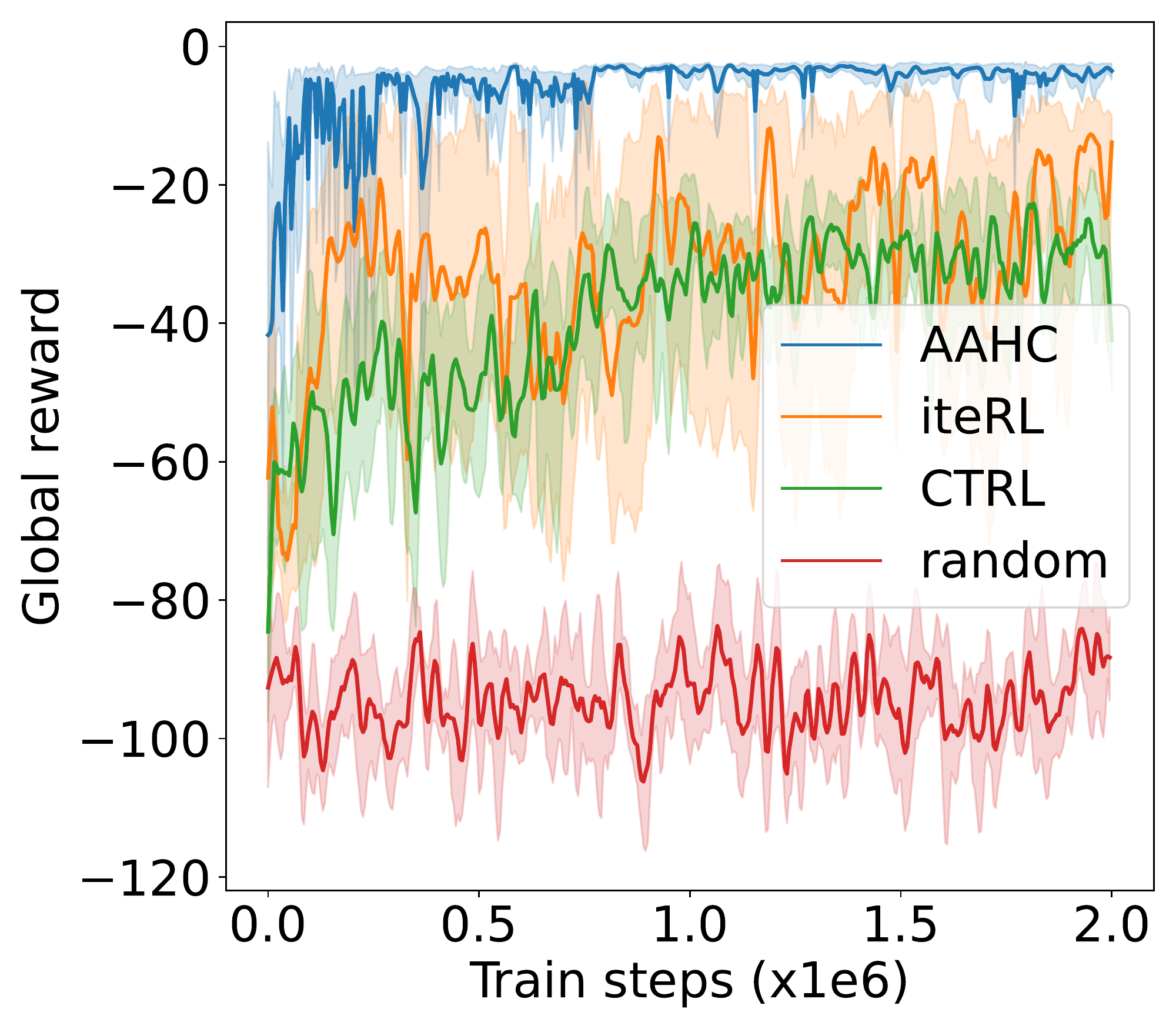}
\vspace{-10mm}
\end{minipage}%
}%
\subfigure[3-6 global reward.]{
\begin{minipage}[t]{0.2\linewidth}
\centering
\includegraphics[width=1\linewidth]{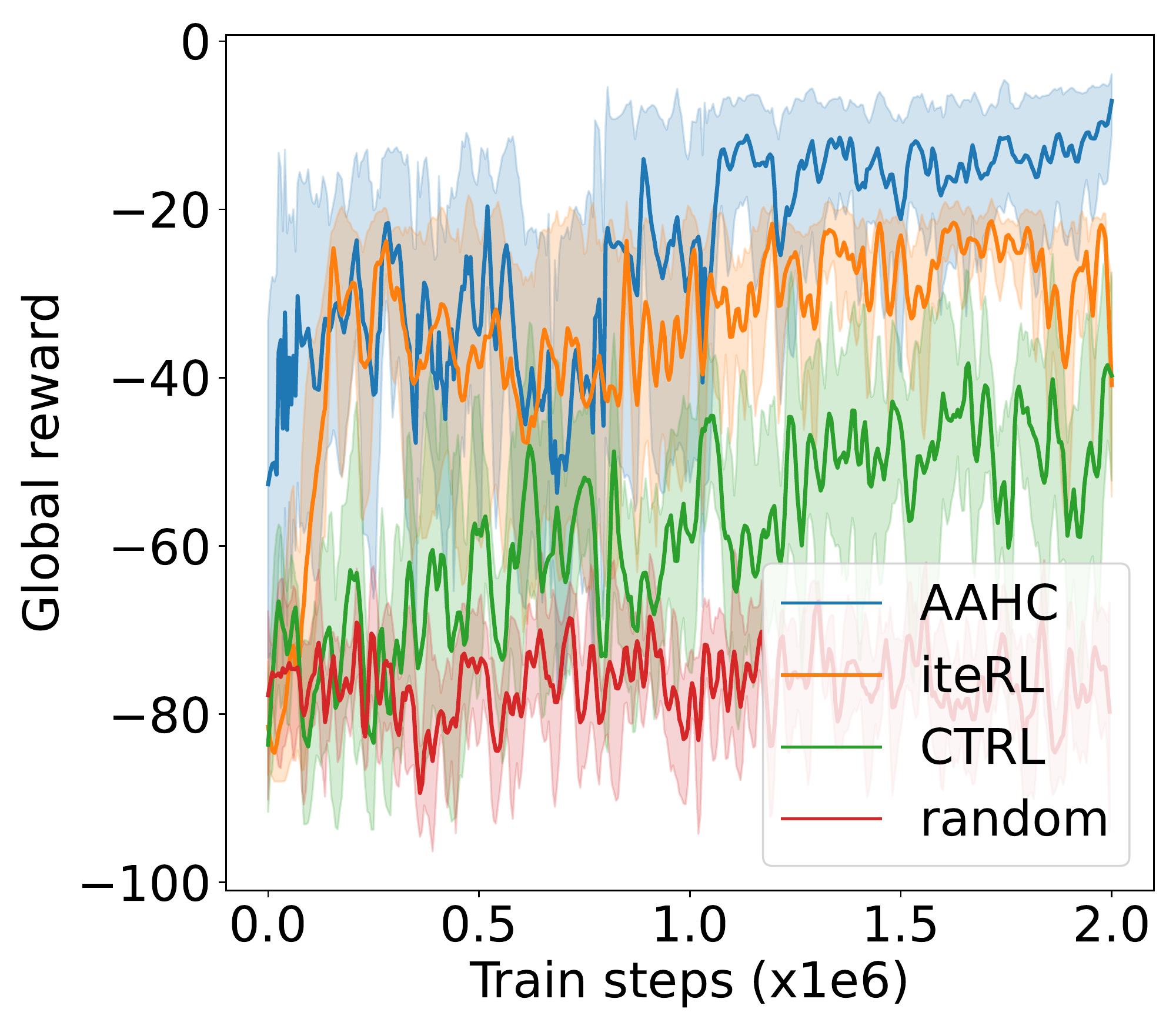}
\vspace{-10mm}
\end{minipage}
}%
\subfigure[3-7 global reward.]{
\begin{minipage}[t]{0.2\linewidth}
\centering
\includegraphics[width=1\linewidth]{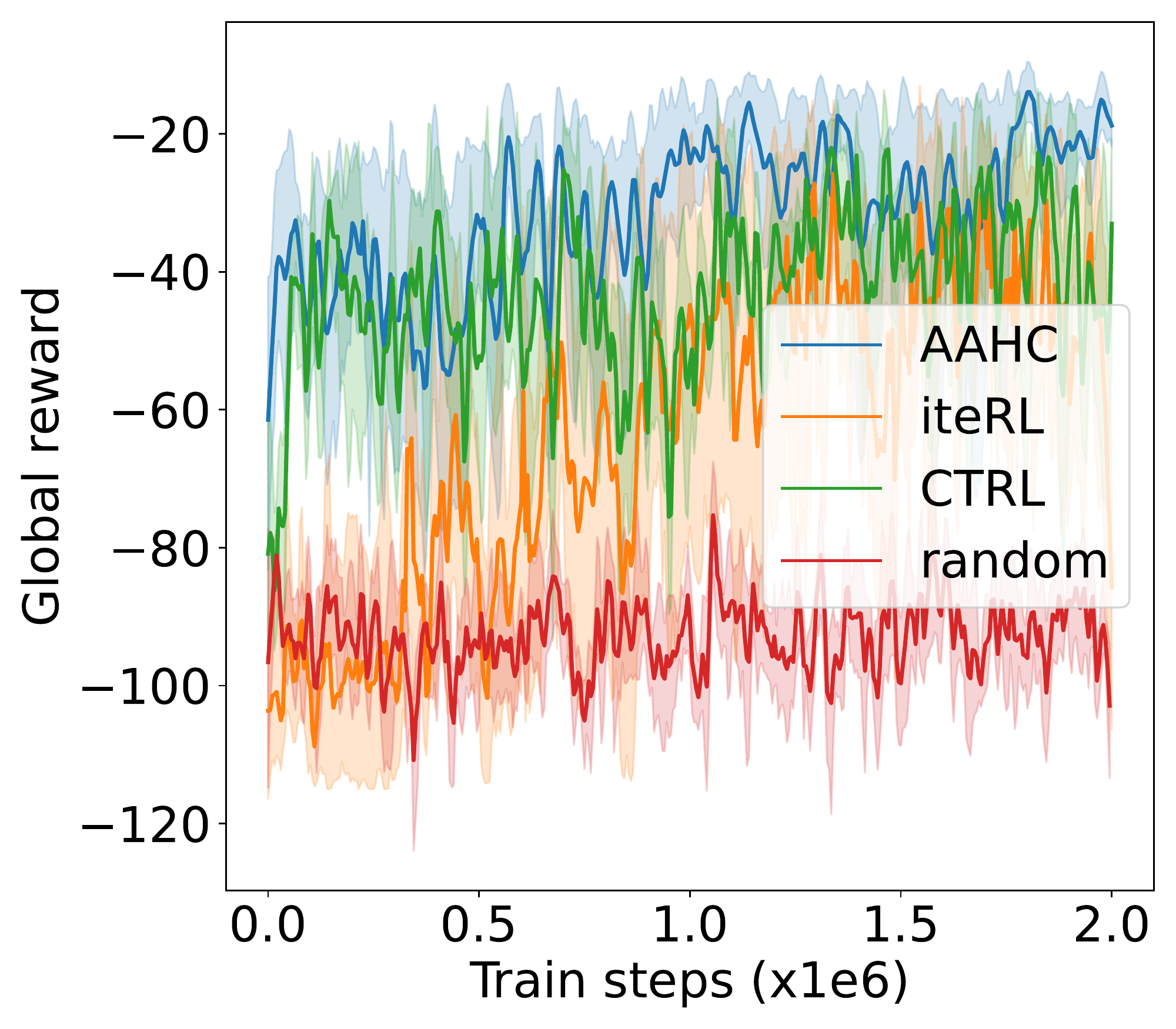}
\vspace{-10mm}
\end{minipage}
}%
\subfigure[3-8 global reward.]{
\begin{minipage}[t]{0.2\linewidth}
\centering
\includegraphics[width=1\linewidth]{experiment/38Greward.pdf}
\vspace{-10mm}
\end{minipage}
}%

\subfigure[3-4 max uplink rate.]{
\begin{minipage}[t]{0.2\linewidth}
\centering
\includegraphics[width=1\linewidth]{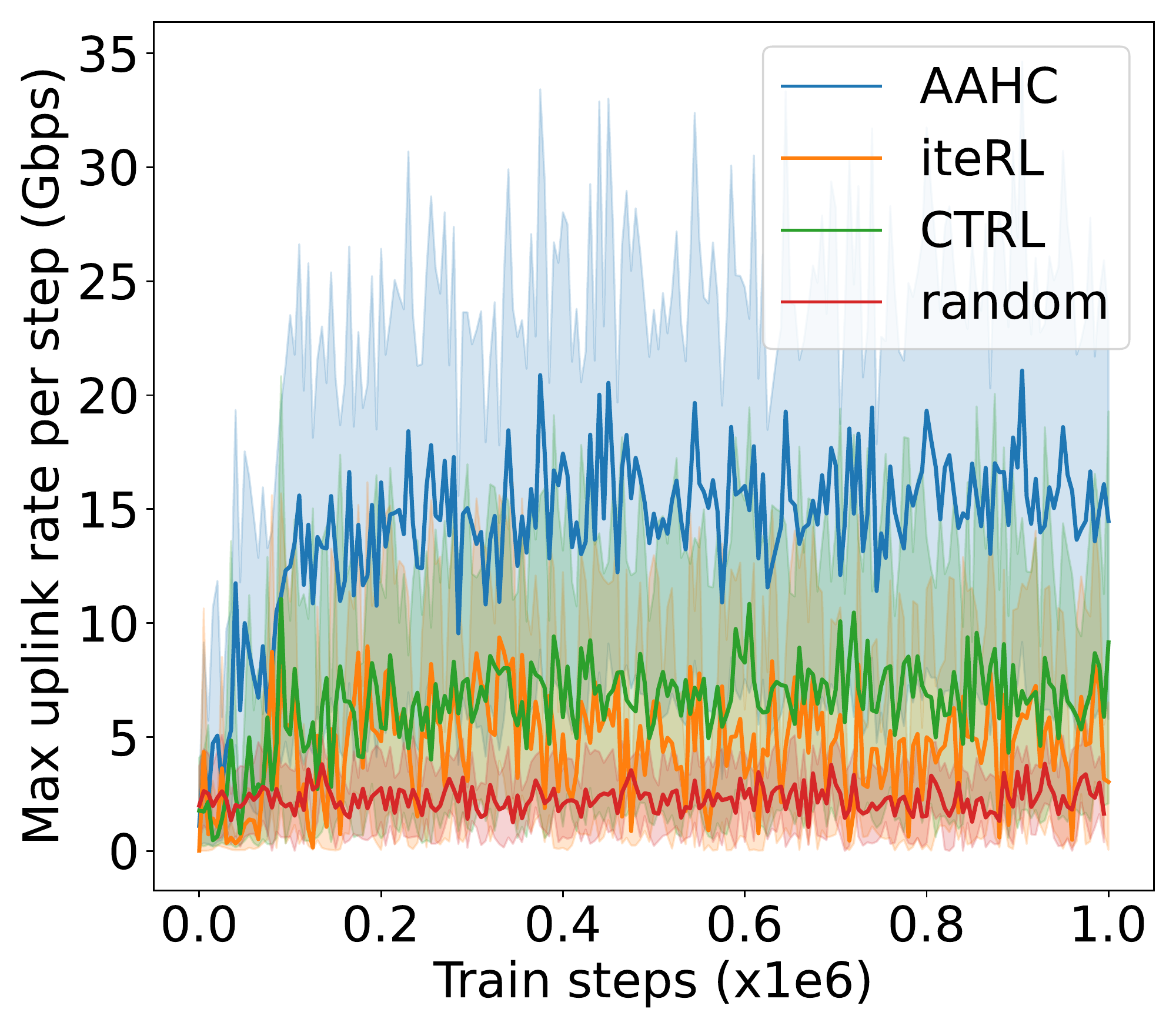}
\vspace{-10mm}
\end{minipage}%
}%
\subfigure[3-5 max uplink rate.]{
\begin{minipage}[t]{0.2\linewidth}
\centering
\includegraphics[width=1\linewidth]{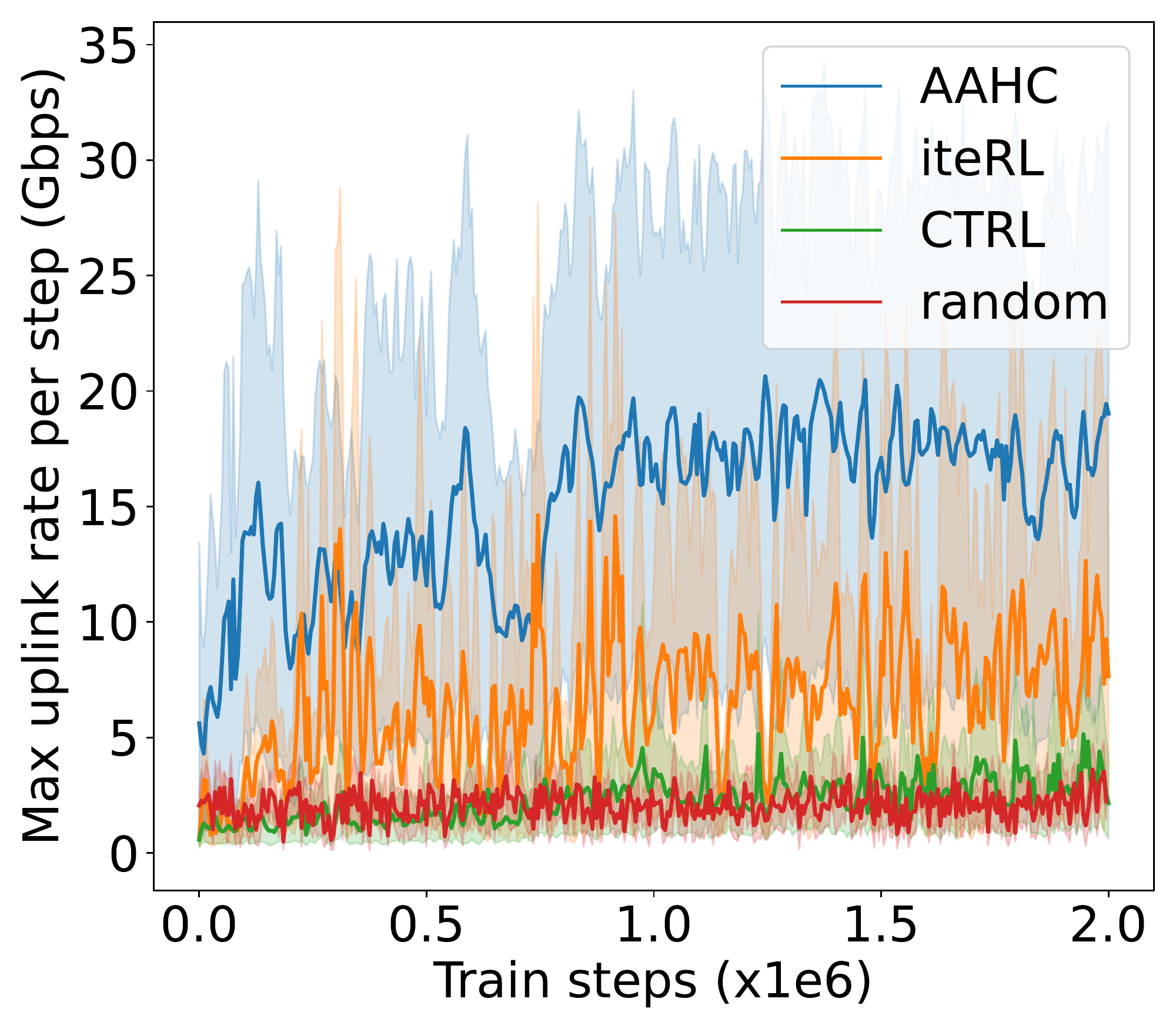}
\vspace{-10mm}
\end{minipage}%
}%
\subfigure[3-6 max uplink rate.]{
\begin{minipage}[t]{0.2\linewidth}
\centering
\includegraphics[width=1\linewidth]{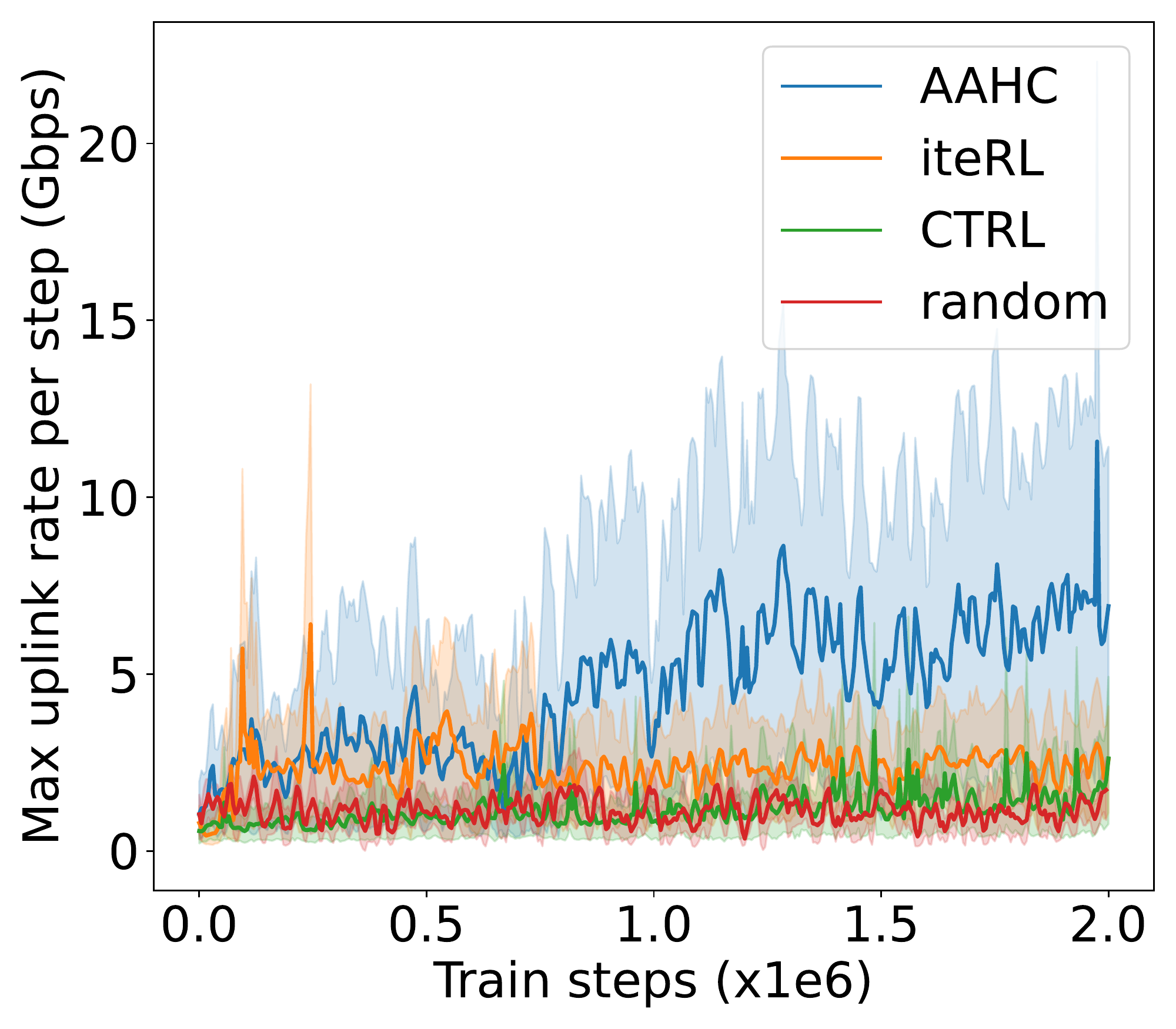}
\vspace{-10mm}
\end{minipage}
}%
\subfigure[3-7 max uplink rate.]{
\begin{minipage}[t]{0.2\linewidth}
\centering
\includegraphics[width=1\linewidth]{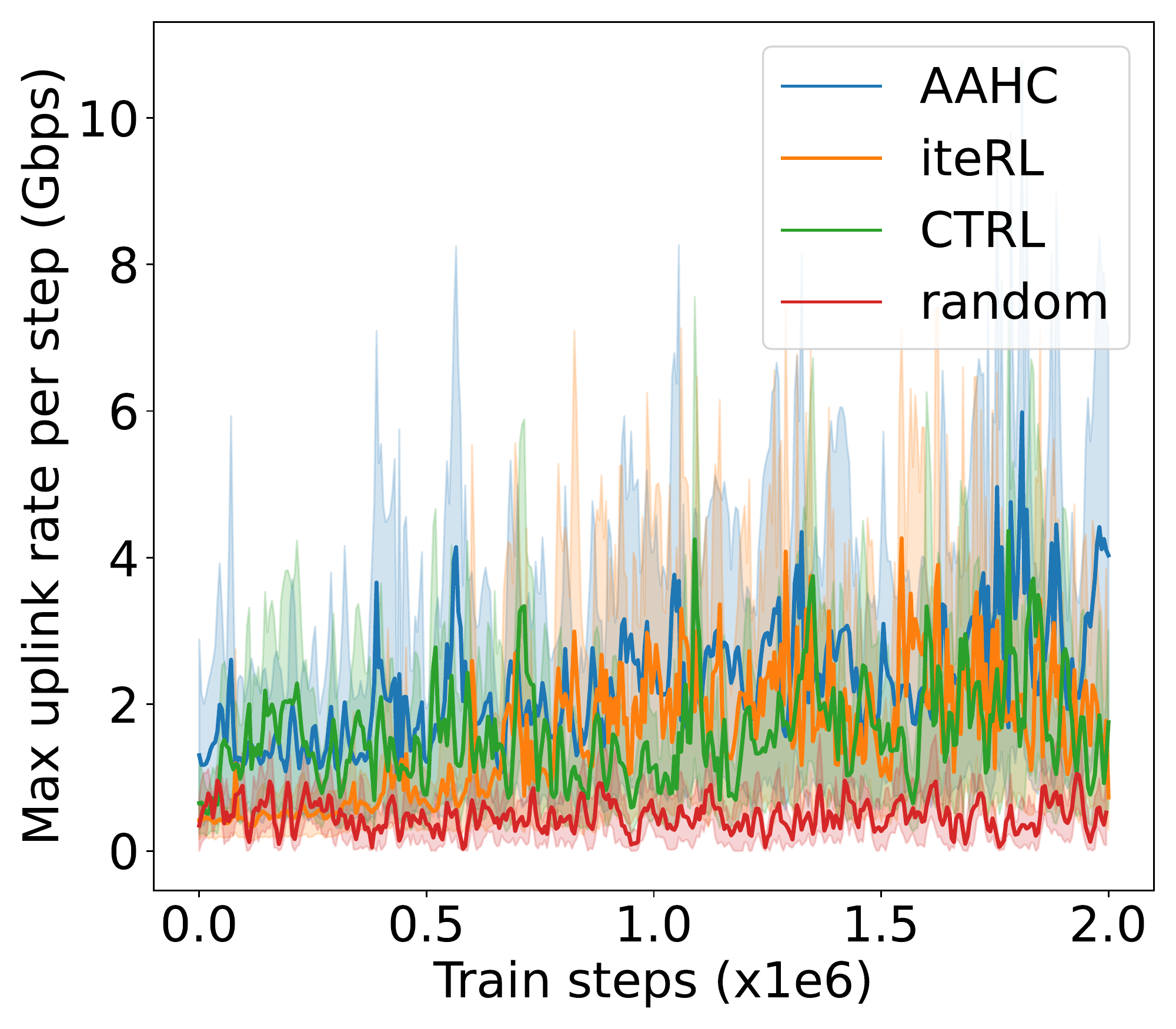}
\vspace{-10mm}
\end{minipage}
}%
\subfigure[3-8 max uplink rate.]{
\begin{minipage}[t]{0.2\linewidth}
\centering
\includegraphics[width=1\linewidth]{experiment/38max_rate.pdf}
\vspace{-10mm}
\end{minipage}
}%

\subfigure[3-4 energy consumption.]{
\begin{minipage}[t]{0.2\linewidth}
\centering
\includegraphics[width=1\linewidth]{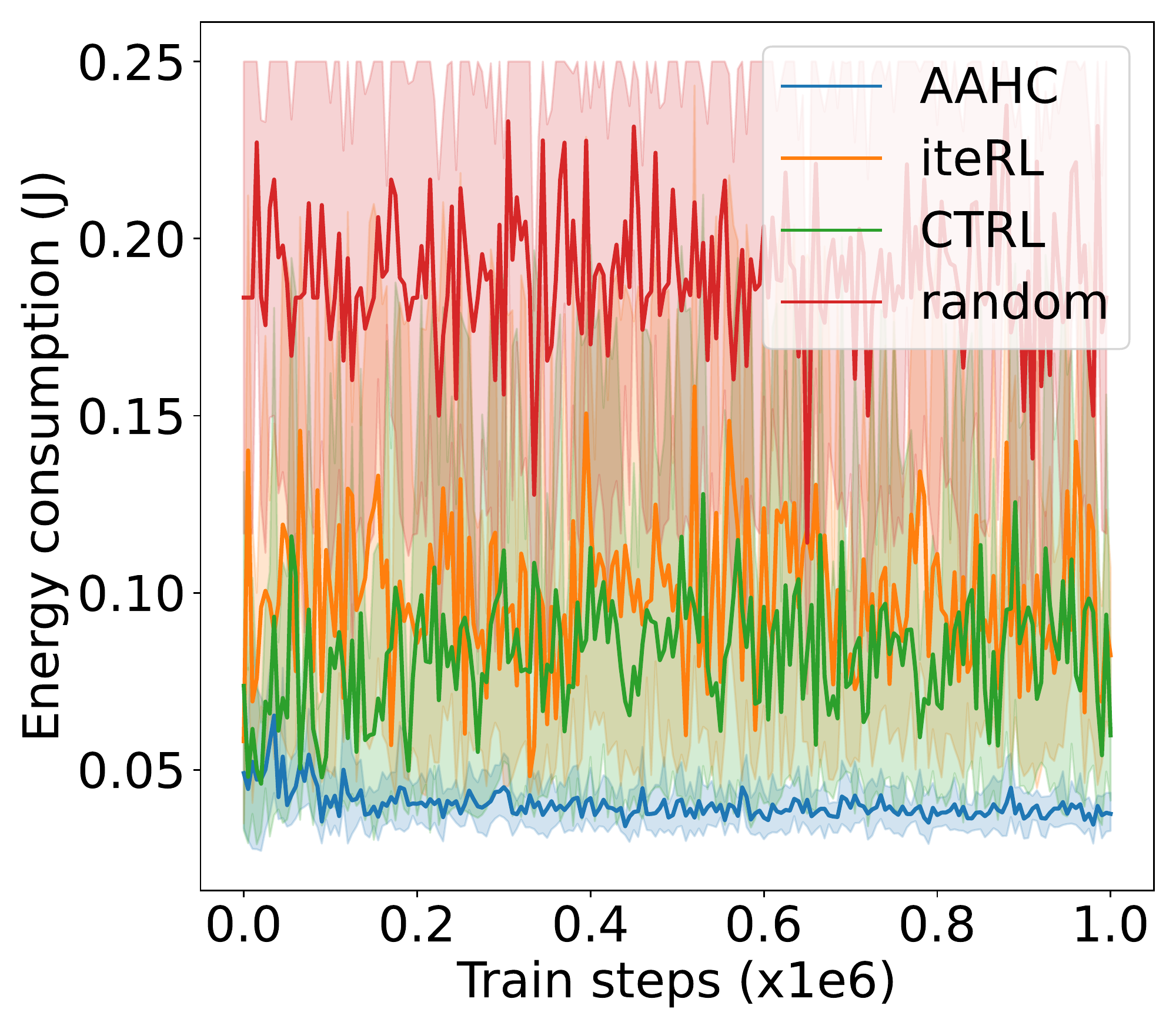}
\vspace{-10mm}
\end{minipage}%
}%
\subfigure[3-5 energy consumption.]{
\begin{minipage}[t]{0.2\linewidth}
\centering
\includegraphics[width=1\linewidth]{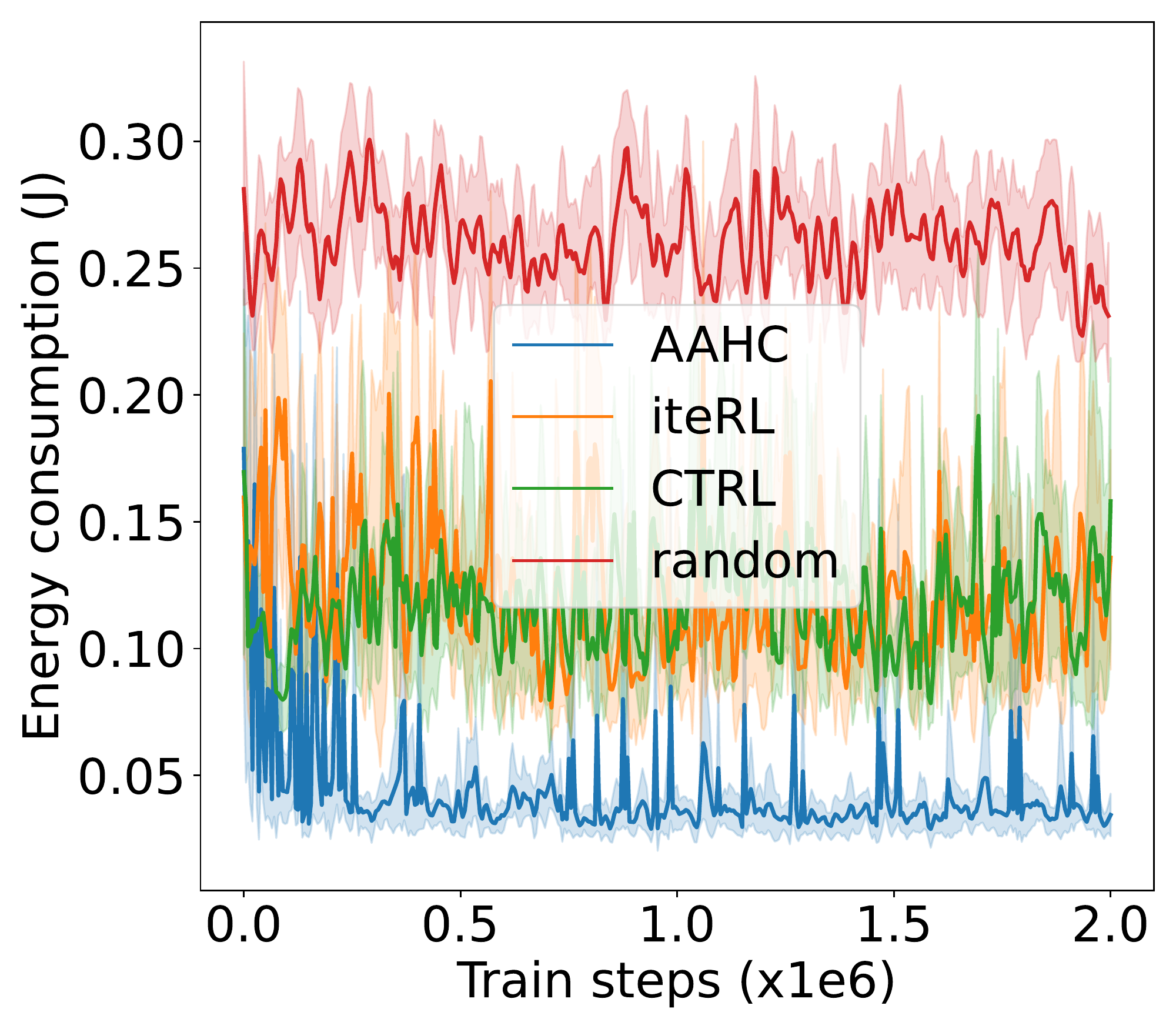}
\vspace{-10mm}
\end{minipage}%
}%
\subfigure[3-6 energy consumption.]{
\begin{minipage}[t]{0.2\linewidth}
\centering
\includegraphics[width=1\linewidth]{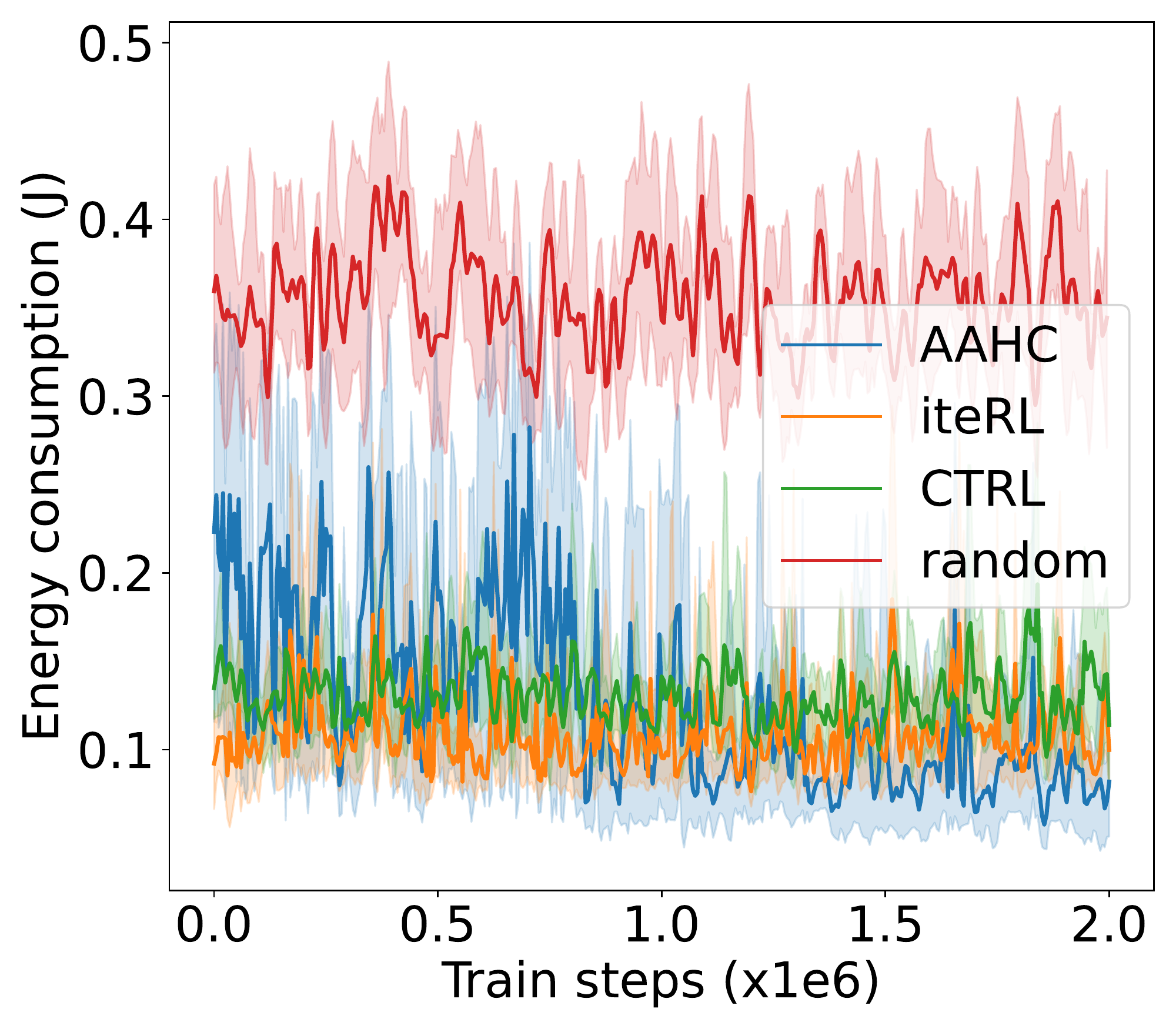}
\vspace{-10mm}
\end{minipage}
}%
\subfigure[3-7 energy consumption.]{
\begin{minipage}[t]{0.2\linewidth}
\centering
\includegraphics[width=1\linewidth]{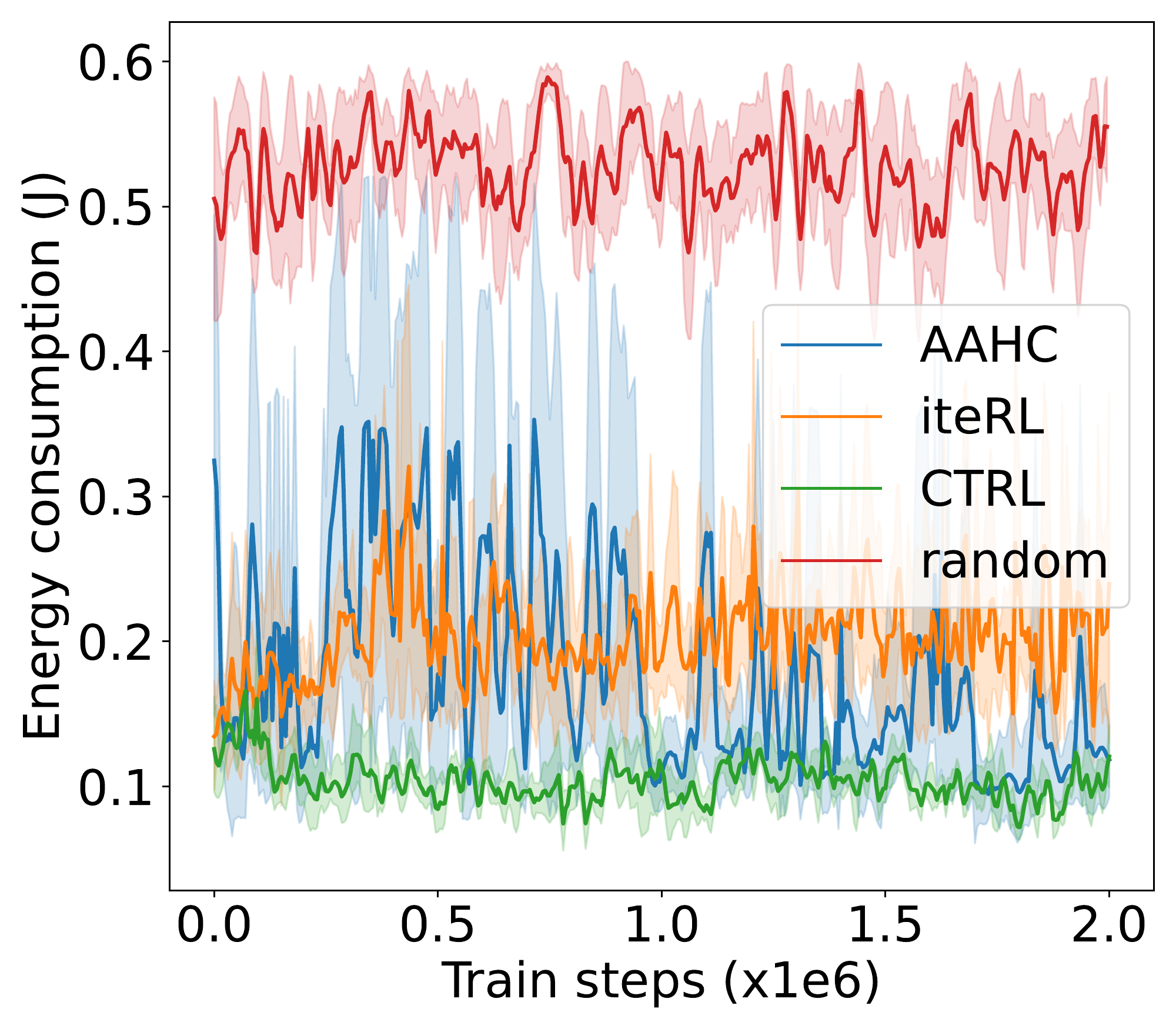}
\vspace{-10mm}
\end{minipage}
}%
\subfigure[3-8 energy consumption.]{
\begin{minipage}[t]{0.2\linewidth}
\centering
\includegraphics[width=1\linewidth]{experiment/38energy.pdf}
\vspace{-10mm}
\end{minipage}
}%

\subfigure[3-4 uplink reward.]{
\begin{minipage}[t]{0.2\linewidth}
\centering
\includegraphics[width=1\linewidth]{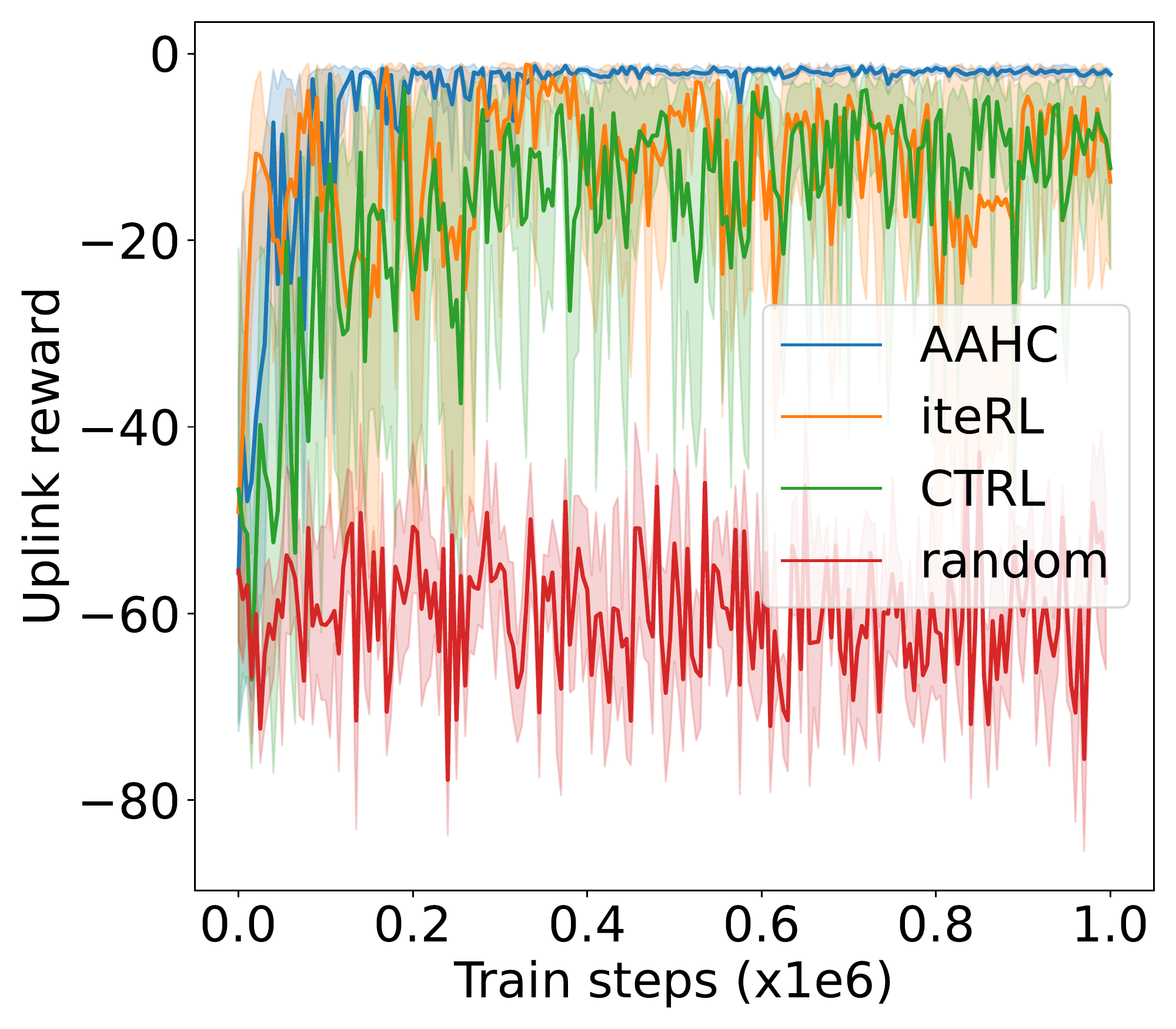}
\vspace{-10mm}
\end{minipage}%
}%
\subfigure[3-5 uplink reward.]{
\begin{minipage}[t]{0.2\linewidth}
\centering
\includegraphics[width=1\linewidth]{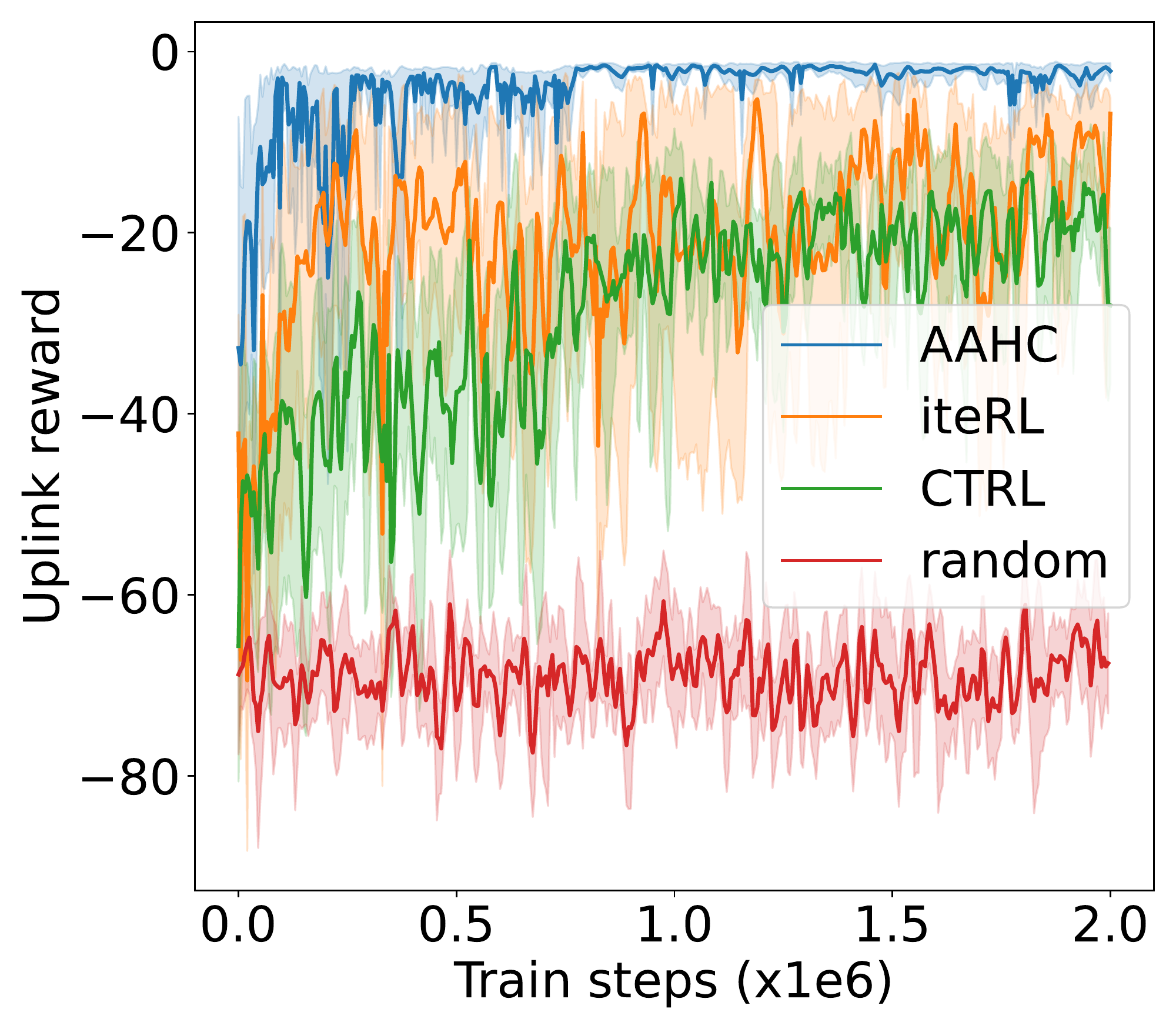}
\vspace{-10mm}
\end{minipage}%
}%
\subfigure[3-6 uplink reward.]{
\begin{minipage}[t]{0.2\linewidth}
\centering
\includegraphics[width=1\linewidth]{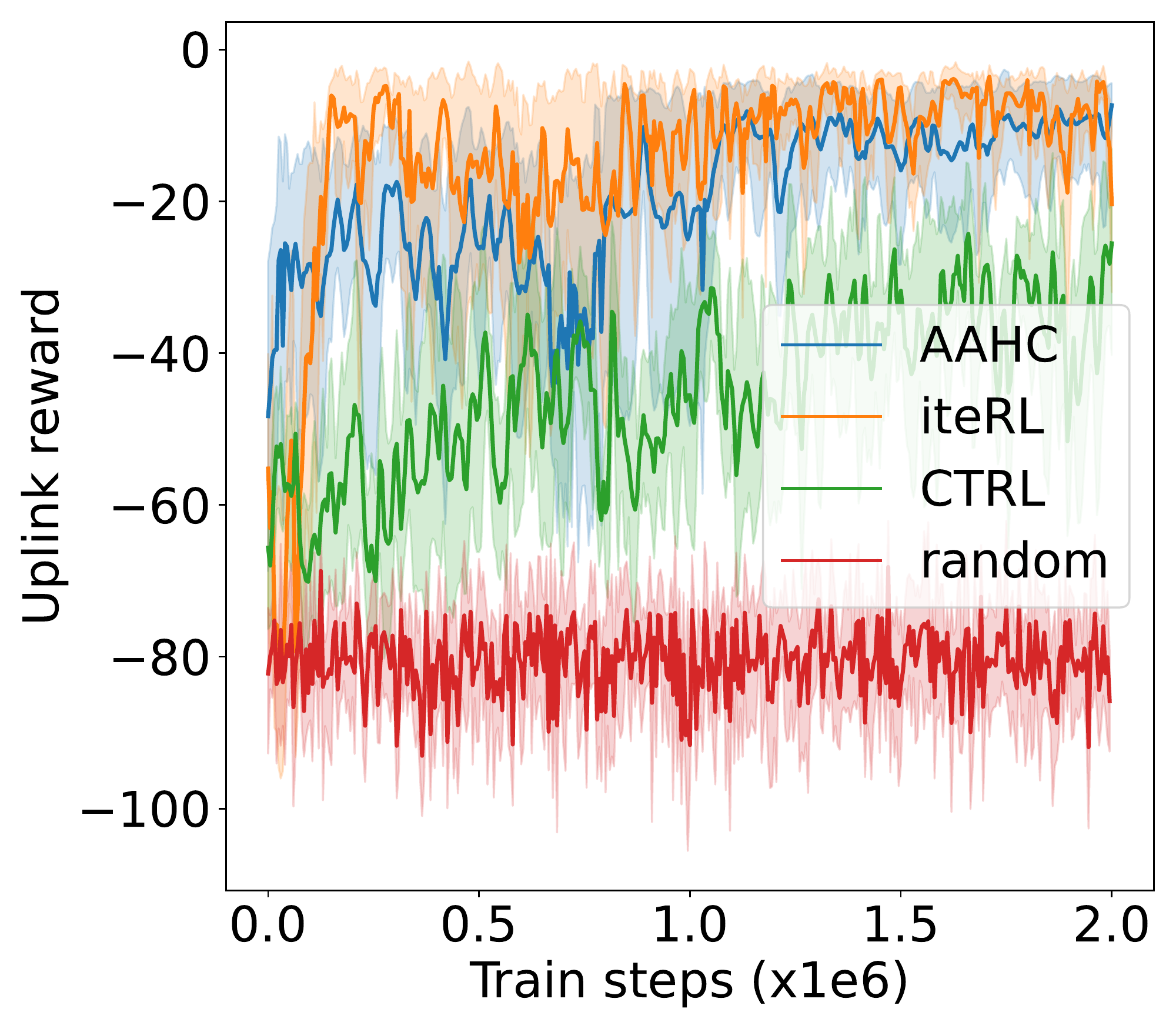}
\vspace{-10mm}
\end{minipage}
}%
\subfigure[3-7 uplink reward.]{
\begin{minipage}[t]{0.2\linewidth}
\centering
\includegraphics[width=1\linewidth]{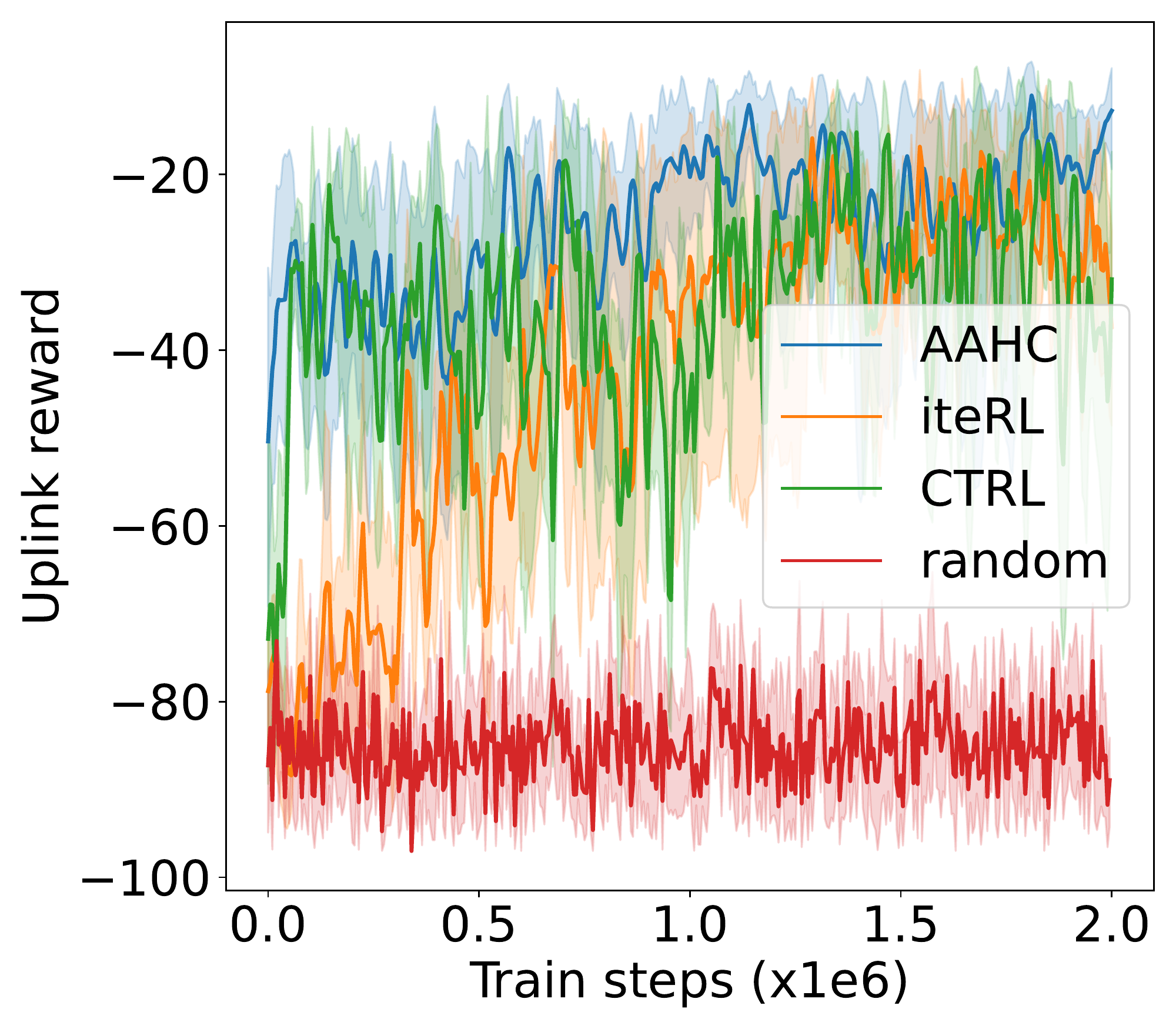}
\vspace{-10mm}
\end{minipage}
}%
\subfigure[3-8 uplink reward.]{
\begin{minipage}[t]{0.2\linewidth}
\centering
\includegraphics[width=1\linewidth]{experiment/38Ureward.pdf}
\vspace{-10mm}
\end{minipage}
}%

\subfigure[3-4 downlink reward.]{
\begin{minipage}[t]{0.2\linewidth}
\centering
\includegraphics[width=1\linewidth]{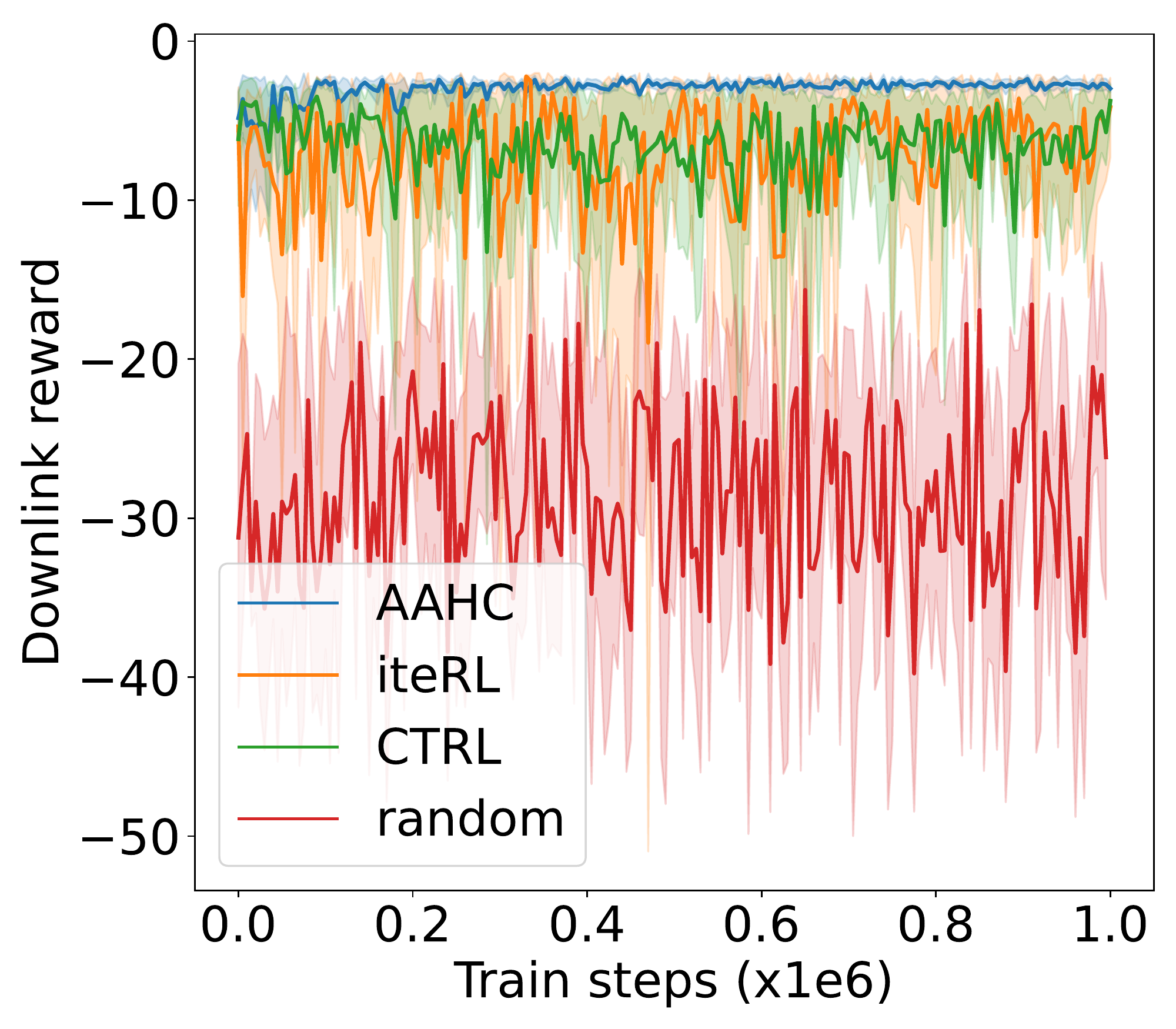}
\vspace{-10mm}
\end{minipage}%
}%
\subfigure[3-5 downlink reward.]{
\begin{minipage}[t]{0.2\linewidth}
\centering
\includegraphics[width=1\linewidth]{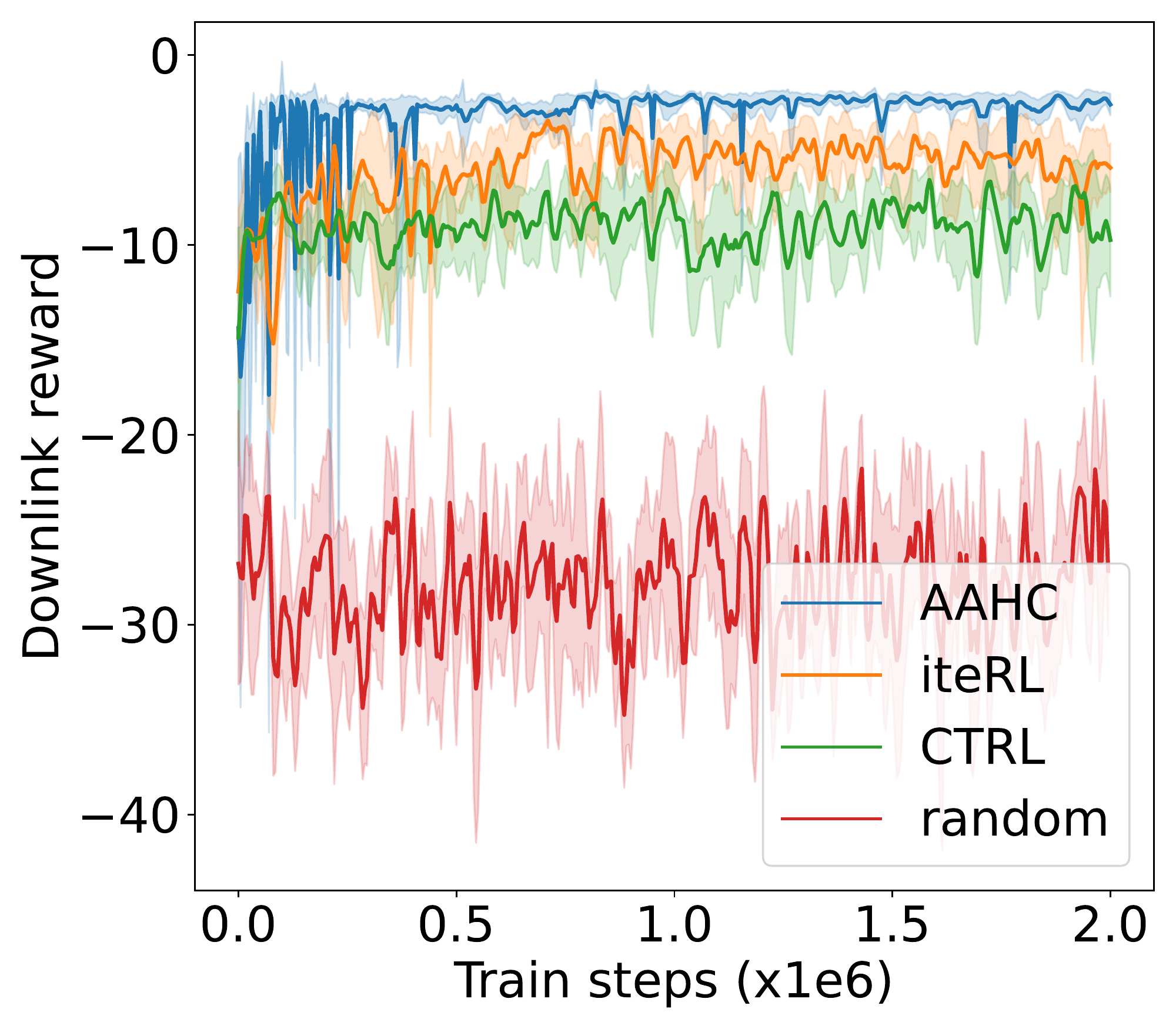}
\vspace{-10mm}
\end{minipage}%
}%
\subfigure[3-6 downlink reward.]{
\begin{minipage}[t]{0.2\linewidth}
\centering
\includegraphics[width=1\linewidth]{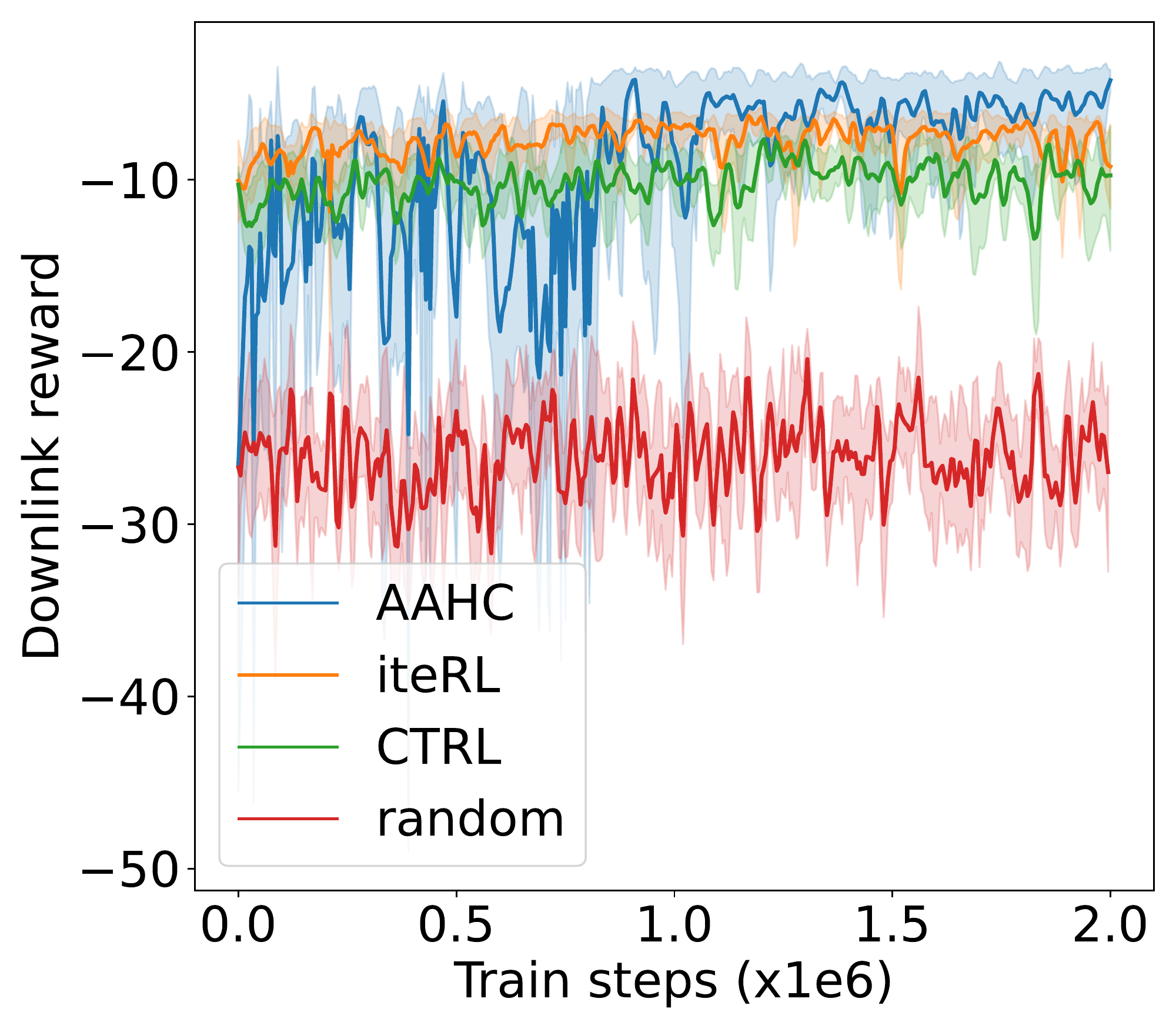}
\vspace{-10mm}
\end{minipage}
}%
\subfigure[3-7 downlink reward.]{
\begin{minipage}[t]{0.2\linewidth}
\centering
\includegraphics[width=1\linewidth]{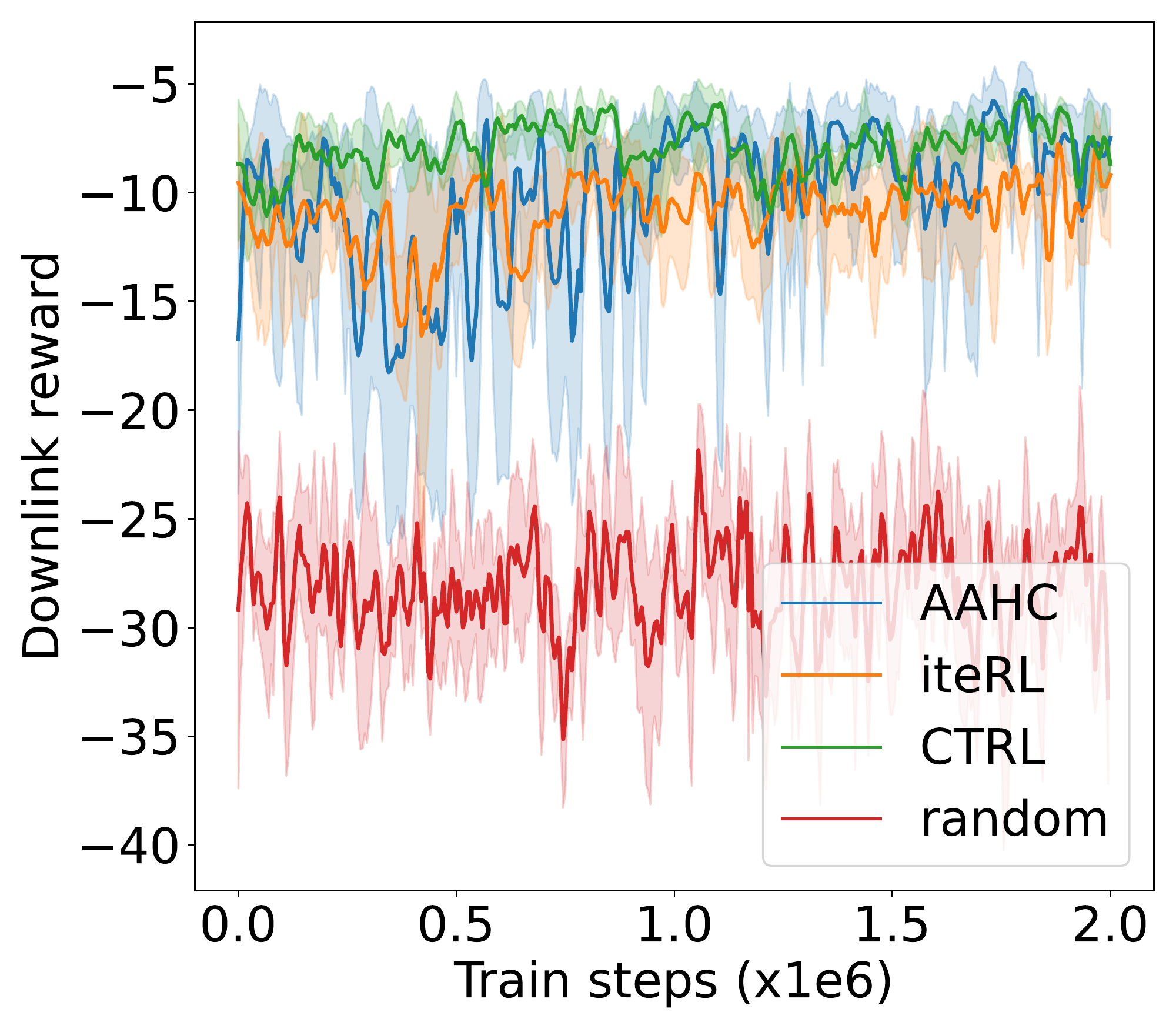}
\vspace{-10mm}
\end{minipage}
}%
\subfigure[3-8 downlink reward.]{
\begin{minipage}[t]{0.2\linewidth}
\centering
\includegraphics[width=1\linewidth]{experiment/38Dreward.pdf}
\vspace{-10mm}
\end{minipage}
}%
\caption{Total iterations, rewards, UL rate, energy consumption in 3-4 to 3-8 scenarios during training.} \label{fig8}
\label{fig:complete}
\vspace{-0.5cm}
\end{figure*}

We have provided selected experimental results (metrics under the 3-8 scenario during training) in Section~\ref{secResults}. Additional experimental results are presented in Figure~\ref{fig7} below and in Figure~\ref{fig8} on the next page.

\label{app:experiment}

\end{document}